\documentclass[10pt,twocolumn,letterpaper]{article}

\usepackage{iccv}
\usepackage{times}
\usepackage{epsfig}
\usepackage{graphicx}
\usepackage{amsmath}
\usepackage{amssymb}
\usepackage{booktabs}
\usepackage{bbding}
\usepackage{makecell}
\usepackage{multirow}
\usepackage{capt-of}
\usepackage{adjustbox}
\usepackage{tabularx}

\usepackage{etoolbox}
\usepackage[accsupp]{axessibility}

\usepackage[colorlinks,breaklinks=true,bookmarks=false]{hyperref}
\usepackage[capitalize]{cleveref}
\crefname{section}{Sec.}{Secs.}
\Crefname{section}{Section}{Sections}
\Crefname{table}{Table}{Tables}
\crefname{table}{Tab.}{Tabs.}

\iccvfinalcopy 

\def\iccvPaperID{7275} 

\ificcvfinal\fi

\begin{document}

\title{HairCLIPv2:~Unifying Hair Editing via Proxy Feature Blending}

\author{ Tianyi Wei\textsuperscript{\rm 1}, Dongdong Chen\textsuperscript{\rm 2}, Wenbo Zhou\textsuperscript{\rm 1,}, Jing Liao\textsuperscript{\rm 3},\\
	 Weiming Zhang\textsuperscript{\rm 1}, Gang Hua\textsuperscript{\rm 4}, Nenghai Yu\textsuperscript{\rm 1} \\
	\normalsize\textsuperscript{\rm 1}University of Science and Technology of China  \ \normalsize\textsuperscript{\rm 2}Microsoft Cloud AI  \  \\
 \normalsize\textsuperscript{\rm 3}City University of Hong Kong  \ 
	\normalsize\textsuperscript{\rm 4}Xi'an Jiaotong University  \    \\
	{\tt\small\{bestwty@mail., welbeckz@, zhangwm@, ynh@\}ustc.edu.cn } \\
	{\tt\small \{cddlyf@, ganghua@\}gmail.com}, {\tt\small jingliao@cityu.edu.hk}
}

\twocolumn[{
	\renewcommand\twocolumn[1][]{#1}
	\maketitle
	\setlength\tabcolsep{0.5pt}
	\centering
	\small
	\begin{tabular}{c}
		\includegraphics[width=0.99\textwidth]{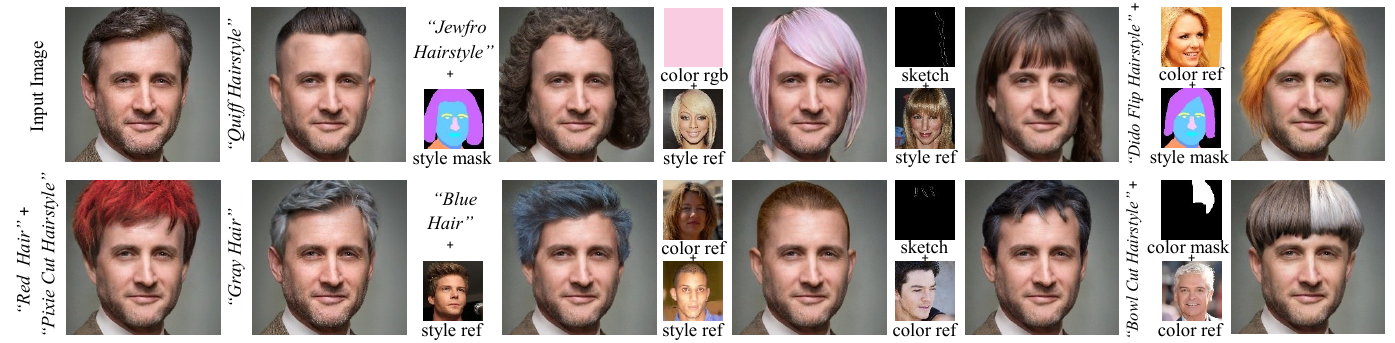}
	\end{tabular}
	\captionof{figure}{HairCLIPv2 supports hairstyle and color editing individually or jointly with unprecedented user interaction mode support, including text, mask, sketch, reference image, etc.}
	\vspace{1.5em}
	\label{fig:teaser}
}]

\maketitle

\ificcvfinal\thispagestyle{empty}\fi

\begin{abstract}
	Hair editing has made tremendous progress in recent years. Early hair editing methods use well-drawn sketches or masks to specify the editing conditions. Even though they can enable very fine-grained local control, such interaction modes are inefficient for the editing conditions that can be easily specified by language descriptions or reference images. Thanks to the recent breakthrough of cross-modal models (e.g., CLIP), HairCLIP is the first work that enables hair editing based on text descriptions or reference images. However, such text-driven and reference-driven interaction modes make HairCLIP unable to support fine-grained controls specified by sketch or mask. In this paper, we propose HairCLIPv2, aiming to support all the aforementioned interactions with one unified framework. Simultaneously, it improves upon HairCLIP with better irrelevant attributes (e.g., identity, background) preservation and unseen text descriptions support. The key idea is to convert all the hair editing tasks into hair transfer tasks, with editing conditions converted into different proxies accordingly. The editing effects are added upon the input image by blending the corresponding proxy features within the hairstyle or hair color feature spaces. Besides the unprecedented user interaction mode support, quantitative and qualitative experiments demonstrate the superiority of HairCLIPv2 in terms of editing effects, irrelevant attribute preservation and visual naturalness. Our code is available at \url{https://github.com/wty-ustc/HairCLIPv2}.
\end{abstract}

\section{Introduction}
\label{sec:intro}

Hair editing as an interesting and challenging problem has attracted a lot of research attention from both academia and industry. Over the past few decades, tremendous progress \cite{zhu2020sean, xiao2021sketchhairsalon,tan2020michigan,wei2022hairclip} has been made in this field, enabling high-fidelity hair editing based on various types of user interactions or controls. In earlier hair editing methods~\cite{zhu2020sean, xiao2021sketchhairsalon}, commonly supported user editing conditions are sketches and masks, which can enable fine-grained local controls. But in real scenarios, many hair editing conditions can be specified by simpler interactions, e.g., text descriptions (e.g., ``bowl cut hairstyle'') and reference images.  

\begin{table*}[t]
	\centering
	\footnotesize
	\scalebox{0.95}
	{
		\begin{tabular}{l|cccccccc}
			\hline
			& HairCLIP~\cite{wei2022hairclip} & LOHO~\cite{saha2021loho} & Barbershop~\cite{zhu2021barbershop} & HairNet~\cite{zhu2022hairnet} & SYH~\cite{kim2022style} & MichiGAN~\cite{tan2020michigan} & SketchSalon~\cite{xiao2021sketchhairsalon} & Ours \\
			\hline
			Aligned Hair Transfer & \CheckmarkBold & \CheckmarkBold & \CheckmarkBold & \CheckmarkBold & \CheckmarkBold & \CheckmarkBold & \XSolidBrush & \CheckmarkBold    \\
			Unaligned Hair Transfer & \CheckmarkBold & \XSolidBrush & \XSolidBrush & \CheckmarkBold & \CheckmarkBold & \XSolidBrush & \XSolidBrush & \CheckmarkBold    \\
			Text & \CheckmarkBold & \XSolidBrush & \XSolidBrush & \XSolidBrush & \XSolidBrush & \XSolidBrush & \XSolidBrush & \CheckmarkBold \\
			Mask & \XSolidBrush & \XSolidBrush & \CheckmarkBold & \XSolidBrush & \CheckmarkBold & \CheckmarkBold & \XSolidBrush & \CheckmarkBold    \\
			Sketch & \XSolidBrush & \XSolidBrush & \XSolidBrush & \XSolidBrush & \XSolidBrush & \CheckmarkBold & \CheckmarkBold & \CheckmarkBold \\
			Local Hairstyle Editing & \XSolidBrush & \XSolidBrush & \XSolidBrush & \XSolidBrush & \XSolidBrush & \CheckmarkBold & \CheckmarkBold & \CheckmarkBold \\
			Local Hair Color Editing & \XSolidBrush & \XSolidBrush & \XSolidBrush & \XSolidBrush & \XSolidBrush & \XSolidBrush & \CheckmarkBold & \CheckmarkBold \\
			\hline
		\end{tabular}
	}
    \vspace{0.3em}
	\caption{Comparisons between our approach and mainstream hair editing methods in terms of available interaction modes and functionality. Only our method supports all interaction modes and enables both global and local hair editing.}
	\label{tab:interaction_compare}
\end{table*}


Recently, cross-modal visual and language representation learning~\cite{su2019vl, radford2021learning,yuan2021florence,dong2023maskclip,wang2022omnivl,weng2023transforming,zhao2023x} has made remarkable breakthrough, which makes text-guided image manipulation possible. HairCLIP~\cite{wei2022hairclip} presents the first attempt that supports hair editing via text description and reference image within one unified framework. Despite such text-driven and reference-driven interaction being more efficient and user-friendly, HairCLIP cannot support fine-grained controls like sketches and masks. Moreover, HairCLIP has two other limitations: 1) Since hair editing in HairCLIP is accomplished by pure latent code manipulation, it will inevitably alter other irrelevant attributes (e.g., identity, background) because fully decoupling different attributes in latent codes is difficult; 2) It struggles in yielding satisfactory results for text descriptions that differ significantly from training texts.

In this paper, we take a step forward and propose HairCLIPv2, a unified hair editing system that unprecedentedly supports all the aforementioned interaction modes, including the natural text/reference-driven interaction and fine-grained local interaction. In Table \ref{tab:interaction_compare}, we list the interaction modes and editing functionality supported by existing hair editing methods. Moreover, with fundamentally different editing mechanism design, HairCLIPv2 makes great improvement upon HairCLIP with better irrelevant attributes preservation and unseen text description support. 

The key idea of HairCLIPv2 is \emph{converting all the hair editing tasks into hair transfer tasks, and the editing conditions are converted into different transfer proxies accordingly}. Conceptually, it can be understood as ``find proxy hair images that satisfy the editing conditions and transfer the corresponding attributes to the source image". Note that we use the StyleGAN latent code or feature corresponding to such proxies rather than use the proxy images explicitly.

More specifically, we first transform the input source image into the bald proxy, which inpaints the hair-covered regions (e.g., background, ears) with reasonable semantic attributes. This can help avoid the editing artifacts caused by occlusion when blending the source image with condition proxies. 
For different editing proxies, we define their generation as different tasks performed in StyleGAN according to their characteristics. Depending on the users' editing preferences, hair editing effects are then enforced upon the input image by blending the corresponding proxy features within the hairstyle feature space or hair color feature space. This is different from HairCLIP that achieves the editing effect by manipulating the 1-d latent codes. Such feature blending based editing naturally supports global and local hair editing by controlling the blending area to cover the entire hair area or part of it.

To show the superiority of HairCLIPv2, we conduct extensive comparisons. In addition to more complete user interaction modes support, HairCLIPv2 also shows obvious advantages in terms of manipulation accuracy, irrelevant attribute preservation, and visual naturalness. Some interactive editing examples are provided in Figure \ref{fig:teaser}. Our contributions can be summarized as below:
\begin{itemize}
	\item We present a fresh perspective for hair editing tasks and propose a novel hair editing paradigm that unifies various types of editing into the form of proxy hair transfers. We achieve all editing effects with the feature blending mechanism, which not only alleviates the editing pressure on each proxy but also enables excellent irrelevant attribute preservation.

	\item We dedicately design the proxy generation for different conditions based on their own special properties, e.g., for the text proxy, the decoupled proxy design and optimization starting point selection strategy help us achieve better editing effects and arbitrary text support; for the sketch proxy, we achieve local hairstyle editing support for the first time within the StyleGAN-based framework by formalizing     its generation as the image translation task and incorporating insights of semantic layering in StyleGAN.

	\item Our system pushes the interactions of hair editing to a new level, supporting arbitrary text, mask, reference image, sketch and their combinations, and enabling both global and local hair editing, which has never been realized before.
	
\end{itemize}

\section{Related Work}
\noindent\textbf{Generative Adversarial Networks.} Since being invented, GANs have made considerable progress in terms of training strategies~\cite{karras2020analyzing,Tran2021OnDA}, loss functions~\cite{Ansari2020ACF,Arjovsky2017WassersteinGA, feng22bprincipled}, and network structures~\cite{Schnfeld2020AUB,Gulrajani2017ImprovedTO, feng2021understanding,tan2021diverse,tan2021efficient}. In the field of image synthesis, a series of works called StyleGAN~\cite{karras2019style, karras2020analyzing, karras2020training, karras2021alias} represents the cutting edge of GANs. Given its promising semantically decoupled latent space~\cite{collins2020editing, shen2020interpreting} and high-quality image synthesis abilities, the pre-trained StyleGAN has become the preferred choice for performing image editing. In this paper, we choose StyleGAN2 to develop our framework, which is consistent with other hair editing methods~\cite{wei2022hairclip, zhu2021barbershop, kim2022style, patashnik2021styleclip, xia2021tedigan, saha2021loho} to be compared.

\vspace{0.7em}

\noindent\textbf{Latent Space Embedding and Editing.} As the bridge connecting the pre-trained StyleGAN and other downstream editing tasks, GAN inversion aims to yield the ideal embedding of the real image in the latent space. Based on the application purposes, we roughly classify the GAN inversion methods into two categories: methods~\cite{tov2021designing, zhu2020domain, zhu2020improved} suitable for editing and methods~\cite{abdal2020image2stylegan2, richardson2021encoding, wei2022e2style} for better reconstruction. The former methods project the real image into the embedding subspace more suitable for editing at the expense of reconstruction. Among them, e4e~\cite{tov2021designing} has become the most popular method for editing tasks~\cite{wei2022hairclip, patashnik2021styleclip, wu2022hairmapper} performed in the latent space. The latter approaches~\cite{abdal2020image2stylegan2, richardson2021encoding, wei2022e2style} aim to achieve the perfect reconstruction of the real image. However, limited by the representation capability of the latent space, all these methods cannot achieve the perfect reconstruction. To address this issue, Barbershop~\cite{zhu2021barbershop} proposes a novel inversion method, which additionally introduces a feature space $ \mathcal{F} $ of StyleGAN combined with the latent space $ \mathcal{S} $ to form a new embedding space $ \mathcal{FS} $. Inspired by this, we decouple the editing task from the reconstruction task by blending editing proxy features in the feature space to achieve a unified hair editing system that supports a wide range of interactions.

\noindent\textbf{Hair Editing Using GANs.} 
Existing hair editing methods can be roughly categorized as conditional GANs~\cite{xiao2021sketchhairsalon, tan2020michigan, jo2019sc} based and pre-trained StyleGAN based~\cite{wei2022hairclip, patashnik2021styleclip, xia2021tedigan, saha2021loho, zhu2021barbershop, kim2022style, zhu2022hairnet, song2023hairstyle}. As a pioneering work of hair transfer, MichiGAN~\cite{tan2020michigan} accomplishes hairstyle transfer by extracting the orientation map of the reference image. Barbershop~\cite{zhu2021barbershop} performs hair transfer within their proposed $\mathcal{FS}$ embedding space. But these methods often struggle when large pose differences exist between source and target image. Recently, some improvements~\cite{kim2022style, zhu2022hairnet} on Barbershop make pose unaligned hair transfers possible. Our framework is also compatible with pose unaligned hair transfers and additionally offers more interaction modes. SketchHairSalon~\cite{xiao2021sketchhairsalon} enables local editing of hairstyle and hair color by using colored sketches as the input to the conditional translation network. Unlike them, we show for the first time that the StyleGAN-based framework can also perform local hair editing with sketches as the condition. 

Benefiting from the development of cross-modal models~\cite{su2019vl, radford2021learning}, text-guided hair editing~\cite{wei2022hairclip, patashnik2021styleclip, xia2021tedigan} has become the new trend. StyleCLIP~\cite{patashnik2021styleclip} and TediGAN~\cite{xia2021tedigan} utilize CLIP loss to perform hair editing in an optimized manner. However, since the embedding of real image deviates from the original suitable editing latent space, these methods will fail for some cases. HairCLIP~\cite{wei2022hairclip} alleviates the problem by training a hair mapper on a large-scale dataset, but struggles in yielding good results for descriptions that differ significantly from the training text. Moreover, none of these methods can preserve the irrelevant attributes well. In this work, we present a new perspective to enable text-guided hair editing methods. By decoupling the editing task from the reconstruction, we can better preserve the irrelevant attributes  while enabling high-quality hair manipulation via arbitrary text descriptions. More importantly, there is no prior work that supports so many interaction modes and functionality as we offer.

\section{Proposed Method}

\begin{figure*}[t]
	\centering
	\includegraphics[width=\textwidth]{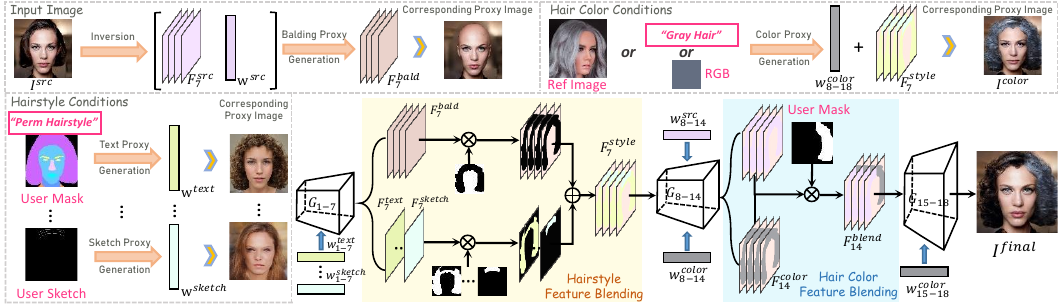} 
	\caption{Overview of HairCLIPv2: Example with hairstyle description text, sketch, mask and hair color RGB values as conditional inputs. Corresponding proxy images are just for better understanding. We complete the hair editing by converting different conditions into different proxies and achieve editing effects by blending them in StyleGAN feature spaces.} 
	\label{fig:modelfigure}
\end{figure*}

\subsection{Preliminaries}

\noindent\textbf{StyleGAN}~\cite{karras2020analyzing} can synthesize photorealistic images with a progressive upsampling network consisting of $ 18 $ layers. Its $ \mathcal{W} \subsetneq \mathbb{R}^{512} $ latent space exhibits good semantic decoupling properties~\cite{collins2020editing, shen2020interpreting}, thus enabling various editing tasks. To perform semantic editing while ensuring the reconstruction quality, some inversion methods~\cite{abdal2019image2stylegan, richardson2021encoding, wei2022e2style} extend the original $ \mathcal{W} $ space to $ \mathcal{W+} $ space, which is defined as a cascade of $ 18 $ different latent codes $ w_i $ from the $ \mathcal{W} $ space, i.e., $ \mathcal{W+} \subsetneq \mathbb{R}^{18\times512} $.

\noindent\textbf{$ \mathbf{\mathcal{FS}} $ Embedding Space} is proposed by Barbershop~\cite{zhu2021barbershop}, which is designed to increase the representational capability of embedding space for details and enable the spatial control of image features. The new $ \{F_7,S\} $ latent code replaces the first 7 layers of $ \mathcal{W+} $ latent code with the $ 32\times32\times512 $ features $ F_7 $ of style-block $ 7 $ of StyleGAN $ G $, i.e., $ F_7 \in \mathcal{F} \subsetneq \mathbb{R}^{32\times32\times512}, S=[w_8, \cdotp\cdotp\cdotp w_i, \cdotp\cdotp\cdotp w_{18}], w_i \in \mathcal{W} $.

\noindent\textbf{CLIP}~\cite{radford2021learning} is a multi-modality model pretrained on web-scale image-text pairs. It can well measure the semantic similarity between given image and text.

\subsection{Overview}
Since the $ \mathcal{FS} $ embedding space~\cite{zhu2021barbershop} is proposed, performing seamless feature blending in $ \mathcal{F} $-space has become the de facto standard for many hair transfer works~\cite{zhu2021barbershop, kim2022style, zhu2022hairnet}, because it can encode the spatial information and preserve local details. On the other hand, many hair editing efforts~\cite{wei2022hairclip, patashnik2021styleclip, xia2021tedigan, wu2022hairmapper} choose to perform editing in the $ \mathcal{W+} $ space, despite unsatisfactory reconstruction, because it can encode rich disentangled semantics~\cite{collins2020editing, shen2020interpreting}. Considering that $ \mathcal{F} $-space is expressive and enables realistic feature integration results while $ \mathcal{W+} $ space is editable, we therefore wonder ``\textit{Can we enjoy the best features of both spaces to facilitate the hair editing task?}". To achieve this goal, we formulate the hair editing tasks as the hair transfer tasks. Specifically, we convert all editing conditions (e.g., text, reference image, sketch) into different proxies in the $ \mathcal{W+} $ space, and accomplish hair editing by seamless proxy feature blending in the feature spaces of StyleGAN. The proxy features of different conditions are obtained with tailored methods based on the condition characteristics.

Following the design in HairCLIP~\cite{wei2022hairclip}, we edit hairstyle and hair color sequentially by blending editing proxy features in the early and later StyleGAN feature space respectively. In detail, as shown in Figure \ref{fig:modelfigure}, we choose to perform proxy feature blending of hairstyle and hair color on feature spaces $ \mathcal{F}^{style} \subsetneq \mathbb{R}^{32\times32\times512} $ and $ \mathcal{F}^{color} \subsetneq \mathbb{R}^{256\times256\times128} $, which correspond to the features of $7$-th and $14$-th style-block in StyleGAN respectively. And users can select various interactions (global or local, single or combined) to edit hairstyles and hair color individually or jointly.

\subsection{Converting Input Image to Bald Proxy}
Given a source image $ I^{src} $ to be edited, we obtain its latent code $ w^{src} $ in $ \mathcal{W+} $ space and feature $ F^{src}_7 $ in $ \mathcal{FS} $ space by II2S~\cite{zhu2020improved} and FS embedding algorithms~\cite{zhu2021barbershop}, respectively. Balding proxy is then generated to inpaint the hair-covered region with reasonable semantic attributes (e.g., background, ears, face, etc.), which can avoid editing artifacts due to occlusion when blending the original image with different editing proxies. 

\noindent\textbf{Balding Proxy.} To remove the occlusion from the hair area of the source image, we bald it using HairMapper~\cite{wu2022hairmapper}: $ w^{bald}=\mathbb{B}(w^{src})$.
$ \mathbb{B} $ denotes HairMapper, which completes the balding editing operation in $ \mathcal{W+} $ space on the latent code $ w^{src} $  to yield the latent code $ w^{bald} $ corresponding to the balded source image. Since editing in $ \mathcal{W+} $ space inevitably gets other irrelevant attributes modified, we circumvent this issue by blending bald feature with source image feature in $  \mathcal{F}^{style} $ space:
\begin{equation}
	F^{bald}_7=G(w^{bald}_{1-7})\times M^{bald}+F^{src}_7\times (1-M^{bald}),
\label{eq:bald_blending}
\end{equation}
where $ G $ stands for StyleGAN, $ G(w^{bald}_{1-7}) $ represents the bald feature in $  \mathcal{F}^{style} $ space. $ M^{bald} $ is the binary mask indicating the hair and ear regions of the source image, which is obtained by the facial parsing network BiseNET~\cite{yu2018bisenet} and downsampled to $ 32\times32 $. With the guidance of $ M^{bald} $, the irrelevant attributes region in $F^{bald}_7$ can continue to be preserved in following proxy feature blending. 

\subsection{Hairstyle Editing}
Below we elaborate on how to generate the proxy for different hairstyle conditions.

\noindent\textbf{Text Proxy.} We formalize text proxy generation as the editing task done in $ \mathcal{W+} $ space based on the CLIP loss guidance. Unlike prior works~\cite{wei2022hairclip, patashnik2021styleclip, xia2021tedigan}, our text proxy generation process is free from the pressure of irrelevant attribute preservation, and allows us to select a more suitable starting point for the optimization process, which leads to better editing effects. In order to ensure both the optimal editing effect and the diversity of editing results, we choose to sample a random point around the mean face latent code as the optimization starting point for our text proxy latent code $ w^{text} $. In detail, we adopt the truncation trick of StyleGAN as $ w^{init}=w^{mean}+\psi(w^{random}-w^{mean}) $, where $ w^{mean} $ is the mean face latent code and $ w^{random} $ is sampled randomly. By setting a small value of $ \psi$, we ensure that the initial optimization starting point $ w^{init} $ is around the average face latent code $ w^{mean} $. We will show the benefits of this initialization strategy in the ablation analysis. We adopt the CLIP loss with transformation augmentations~\cite{avrahami2022blended} to perform the text-guided hairstyle editing while reducing the disturbance caused by adversarial examples:
\begin{equation}
	L^{clip}=\frac{1}{N}\sum^{N}_{i=1}(1-cos(E_{i}(A_{i}(G(w^{text}))), E_{t}(st))),
\end{equation}
where $ A_i$ represents $i$-th transformation augmentation, $ N $ denotes the number of augmentations ($ N=4 $ by default), $ cos(\cdot) $ means cosine similarity, $ G(w^{text}) $ means the editing result for each pass in the optimization process, $ E_{i} $ and $ E_{t} $ stands for the CLIP image encoder and text encoder respectively, and $ st $ refers to the user-supplied text description. Besides, pose alignment loss $ L^{pose} $ is utilized to ensure that the face shape and pose of text proxy are consistent with the source image to ease the subsequent feature blending:
\begin{equation}
	L^{pose}=\frac{1}{N_{k}}||E_{p}(I^{src})-E_{p}(G(w^{text}))||^{2}_{2},
\end{equation}
where $ E_{p} $ represents the 3D keypoint extractor~\cite{bulat2017far} and $ N_{k} $ denotes the number of keypoints. Optionally, the shape loss $ L^{shape} $ is added to constrain the shape of the generated hair according to whether the user provides the hair region mask. We then obtain text proxy feature $ F^{text}_7 $ using the optimized $ w^{text} $: $ F^{text}_7=G(w^{text}_{1-7}). $

\noindent\textbf{Reference Proxy.} Given a hairstyle reference image $ I^{sr} $, we generate the reference proxy by performing the unaligned hairstyle transfer task. $ I^{sr} $ is first inverted by the II2S~\cite{zhu2020improved} embedding algorithm into the $ \mathcal{W+} $ space to get $ w ^{ref}$, which is served as the starting point for hairstyle transfer. During the transfer process, we expect to keep the original hair structure of the reference image while ensuring its pose and facial shape to be consistent with the source image. Thus, $ L^{pose} $ is imposed between $ G(w ^{ref}) $ and $ I^{src} $. In addition, a style loss $ L^{style} $ based on the gram matrix~\cite{gatys2016image} is used to ensure that the hairstyle structure of $ I^{sr} $ remains unchanged during the alignment process:
\begin{multline}
	L^{style}=\frac{1}{4}\sum_{i=1}^{4}||\mathcal{G}_{i}(VGG_{i}(I^{sr}\times M^{rh}))\\
	-\mathcal{G}_{i}(VGG_{i}(G(w ^{ref})\times M^{gh}))||^{2}_{2},
\end{multline}
where $ \mathcal{G}_{i}(\gamma_{i}) $ represents the gram matrix calculated on the $ i $-th layer features and a total of $ 4 $ layers of features are extracted, i.e.,$ \{relu1\_2, relu2\_2, relu3\_3, relu4\_3\} $ of VGG~\cite{simonyan2014very}.  $ M^{rh} $ and $ M^{gh} $ are hair masks for the reference image and the generated image of each round during optimization process predicted by BiseNET~\cite{yu2018bisenet}. For the same purpose, the $ L_{2} $ norm of the manipulation magnitude in the latent space is utilized during optimization:
\begin{equation}
	L^{reg}=||w ^{ref}_{t}-w ^{ref}_{t-1}||^{2}_{2},
\end{equation}
where $ w ^{ref}_{t} $ and $ w ^{ref}_{t-1} $ represent the latent code of the current step and the previous step, respectively. Similar to the text proxy, $ L^{shape} $ is optionally added to allow the user to customize the shape of the hair. We then obtain reference proxy feature $ F^{ref}_7 $ using the optimized $ w^{ref} $: $ F^{ref}_7=G(w^{ref}_{1-7}). $

\noindent\textbf{Sketch Proxy.} Enabling sketch-based local hairstyle editing within our framework is nontrivial. It is hard to find suitable losses to constrain the local hairstyle structure to conform to the sketch given by the user.
To circumvent this problem, we innovatively formalize the synthesis of sketch proxy as an image translation task based on StyleGAN. Utilizing the sketch-hair dataset created by SketchHairSalon~\cite{xiao2021sketchhairsalon}, we train a sketch2hair inverter $ T $, which is based on E2Style~\cite{wei2022e2style} and aims to find the most appropriate latent code in $ \mathcal{W+} $ space to accurately translate a given sketch to the corresponding hair structure. The training loss consists of regular pixel-level $ L_{2} $ loss, feature-level LPIPS~\cite{zhang2018unreasonable} loss and multi-layer face parsing loss~\cite{wei2022e2style} which is introduced to provide more local supervision. During the training process, we randomly remove a portion of the strokes to make our sketch2hair inverter adapt to a variety of sketch inputs from fine to coarse, e.g., even just one stroke. Given a local hairstyle sketch $S$, our sketch proxy features $F^{sketch}_7$ are synthesized by pre-trained sketch2hair $ T $ and StyleGAN $ G $:
\begin{equation}
	w^{sketch}=T(S),\quad  F^{sketch}_7=G(w^{sketch}_{1-7}).
\end{equation}

\noindent\textbf{Proxy Feature Blending.} For text and reference image based hairstyle condition, we perform global blending in $ \mathcal{F}^{style} $ space:
\begin{equation}
	F^{global}_7=F^{tr}_7\times M^{global}+F^{bald}_7\times (1-M^{global}),
\end{equation}
where $ F^{tr}_7\in\{F^{text}_7, F^{ref}_7\} $ and $ M^{global} $ is the binary mask corresponding to the hair region of $ F^{tr}_7 $. Optionally, the sketch-based local hairstyle editing is applied:
\begin{equation}
	F^{style}_7=F^{sketch}_7\times M^{local}+F^{global}_7\times (1-M^{local}),
\end{equation}
where $ M^{local} $ is obtained by downsampling the user input sketch $ S $ to $ 32\times32 $ after dilation. A natural concern is the artifacts brought by the mismatch between hair features within $ M^{local} $ and other hair features. But thanks to the semantic layering characteristics of StyleGAN, the resulting image shows consistent tones as these hair features will be modulated by the later layers. Our framework allows users to only edit the hairstyle: $ I^{style}=G(F^{style}_7, w^{src}_{8-18}), $ by skipping the following hair color editing.

\subsection{Hair Color Editing}
We achieve hair color editing by performing proxy feature blending in $ \mathcal{F}^{color} $ space. By choosing to use the feature $ F^{src}_7 $ or $ F^{style}_7 $, we allow to edit only hair color or both hair style\&hair color. Below, we use $ F^{style}_7 $ as the example.

\noindent\textbf{Color Proxy.} We initialize the color proxy with $ F^{style}_7 $ and $ w^{color}_{8-18}=w^{src}_{8-18} $, and set $ w^{color}_{10-13} $ to be optimizable. The loss $ L^{color} $ in the optimization process consists of $ L^{modal} $ and $ L^{bg} $, where $ L^{modal} $ can be defined as $ L^{clip} $ or the average color $ L_{2} $ loss of the hair region depending on the hair color condition types (text, reference image, RGB values), and $ L^{bg} $ is defined as follows:
\begin{equation}
	L^{bg}=||(I^{style}-I^{color})\times(M^{n-hair})||^{2}_{2},
\end{equation}
where $ I^{color}=G(F^{style}_7, w^{color}_{8-18}) $, $ M^{n-hair} $ is the mask of the non-hair region intersection between $ I^{style} $ and $ I^{color} $.

\begin{figure*}[t]
	\begin{center}
		\setlength{\tabcolsep}{0.5pt}
		\begin{tabular}{m{0.3cm}<{\centering}m{1.38cm}<{\centering}m{1.38cm}<{\centering}m{1.38cm}<{\centering}m{1.38cm}<{\centering}m{1.38cm}<{\centering}m{1.38cm}<{\centering}m{0.3cm}<{\centering}m{1.38cm}<{\centering}m{1.38cm}<{\centering}m{1.38cm}<{\centering}m{1.38cm}<{\centering}m{1.38cm}<{\centering}m{1.38cm}<{\centering}}
			& \scriptsize{Input Image} & \scriptsize{Ours} & \scriptsize{HairCLIP} & \scriptsize{StyleCLIP} & \scriptsize{TediGAN} & \scriptsize{DiffCLIP} &  & \scriptsize{Input Image} & \scriptsize{Ours} & \scriptsize{HairCLIP} & \scriptsize{StyleCLIP} & \scriptsize{TediGAN} & \scriptsize{DiffCLIP}
			\\
			
			\raisebox{0.15cm}{\rotatebox[origin=c]{90}{\footnotesize{{afro}}}}
			&\includegraphics[width=1.35cm]{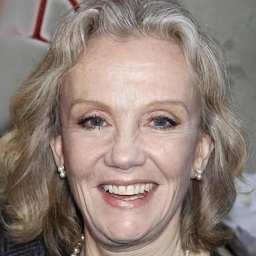}
			&\includegraphics[width=1.35cm]{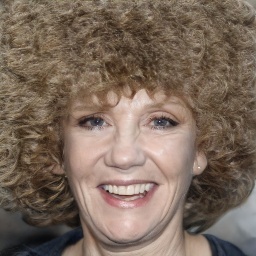}
			&\includegraphics[width=1.35cm]{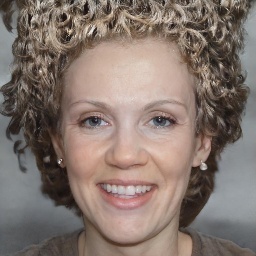}
			&\includegraphics[width=1.35cm]{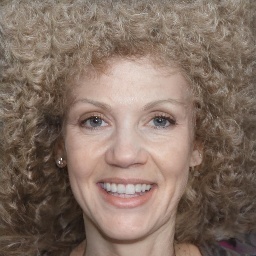}
			&\includegraphics[width=1.35cm]{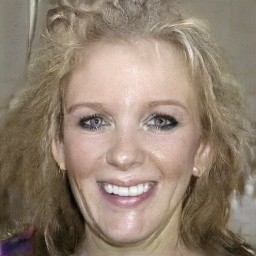}
			&\includegraphics[width=1.35cm]{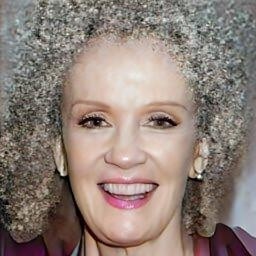}
			&\raisebox{0.15cm}{\rotatebox[origin=c]{90}{\footnotesize{{blond}}}}
			&\includegraphics[width=1.35cm]{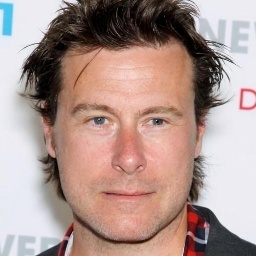}
			&\includegraphics[width=1.35cm]{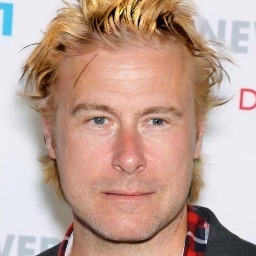}
			&\includegraphics[width=1.35cm]{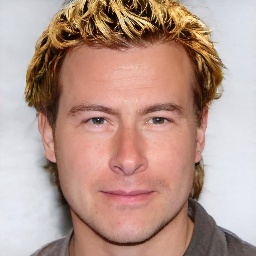}
			&\includegraphics[width=1.35cm]{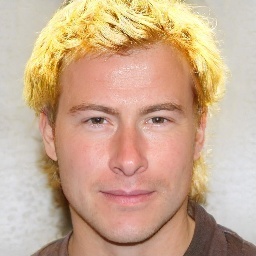}
			&\includegraphics[width=1.35cm]{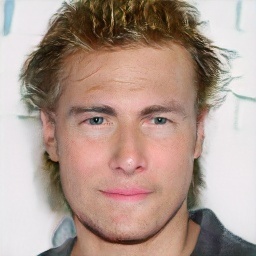}
			&\includegraphics[width=1.35cm]{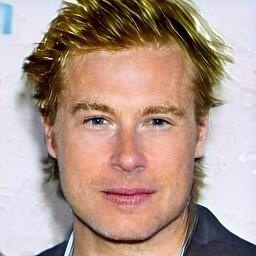}				
			\\

			\raisebox{0.31cm}{\rotatebox[origin=c]{90}{\footnotesize{{bowlcut}}}}
			&\includegraphics[width=1.35cm]{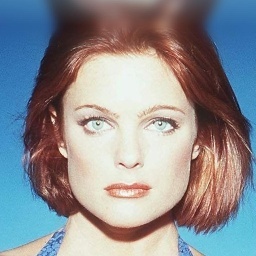}
			&\includegraphics[width=1.35cm]{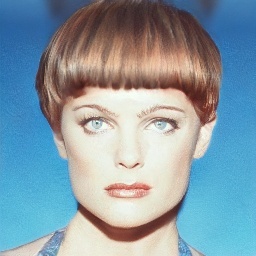}
			&\includegraphics[width=1.35cm]{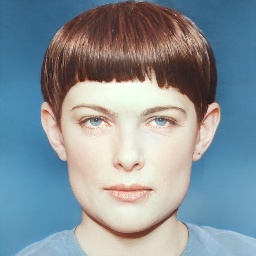}
			&\includegraphics[width=1.35cm]{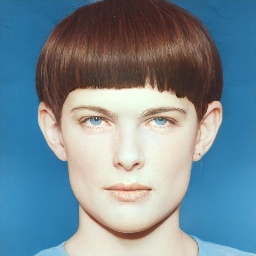}
			&\includegraphics[width=1.35cm]{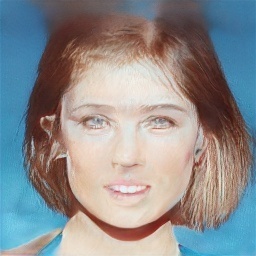}
			&\includegraphics[width=1.35cm]{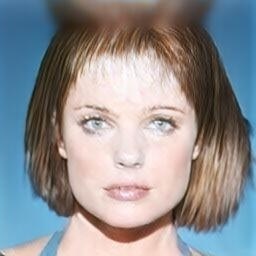}
            &\raisebox{0.59cm}{\rotatebox[origin=c]{90}{\footnotesize{{braid brown}}}}
            &\includegraphics[width=1.35cm]{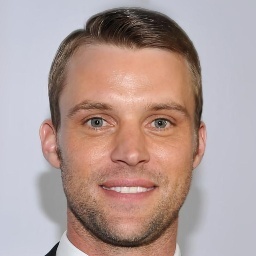}
            &\includegraphics[width=1.35cm]{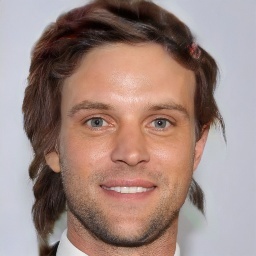}
            &\includegraphics[width=1.35cm]{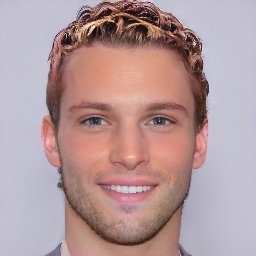}
            &\includegraphics[width=1.35cm]{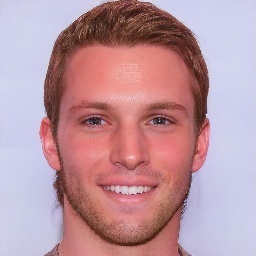}
            &\includegraphics[width=1.35cm]{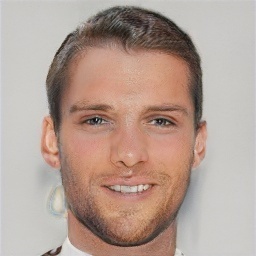}
            &\includegraphics[width=1.35cm]{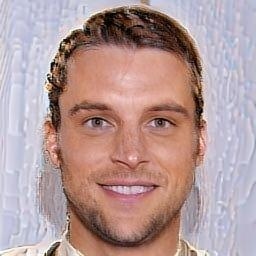}
			\\

		\end{tabular}
	\end{center}
	\caption{Visual comparison with HairCLIP~\cite{wei2022hairclip},  StyleCLIP-Mapper~\cite{patashnik2021styleclip}, TediGAN~\cite{xia2021tedigan} and DiffusionCLIP~\cite{kim2022diffusionclip}. The simplified text descriptions (editing hairstyle, hair color, or both of them) are listed on the leftmost side. Our approach demonstrates better editing effects and irrelevant attribute preservation (e.g., identity, background, etc.).} 
	\label{fig:textcomparefig}
\end{figure*}

\noindent\textbf{Proxy Feature Blending.} Even with the $ L^{bg} $ constraint, we find non-hair regions are often inevitably modified because of imperfect semantic decoupling of the $ \mathcal{W+} $ space. We solve this by performing proxy feature blending in $ \mathcal{F}^{color} $ space, which also naturally supports local hair color editing:
\begin{multline}
	F^{blend}_{14}=G(F^{style}_7, w^{color}_{8-14})\times M^{color}\\
	+G(F^{style}_7, w^{src}_{8-14})\times (1-M^{color}),
\end{multline}
where $ M^{color} $ is the hair area mask or a local editing area mask drawn by the user. We set $ F^{blend}_{14} $ and $ w^{color}_{15-18} $ to be optimizable to further perform the optimization. In the optimization process, we use the $ L_{blend} $ loss, which consists of $ L_{2} $ loss and LPIPS loss to constrain $ I^{final} $ to be similar to $ I^{color} $ inside $ M^{color} $ and similar to $ I^{style} $ outside $ M^{color} $ simultaneously. The final edited image is synthesized as follows: $ I^{final}=G(F^{blend}_{14}, w^{color}_{15-18}). $

\section{Experiments}

Implementation details of our approach are provided in the supplementary material. For all compared methods, we use their official codes or pre-trained models.
\subsection{Quantitative and Qualitative Comparison}

\begin{table}[t]
	\centering
	\setlength{\tabcolsep}{1em}{
		\begin{tabular}{lccc}
			\hline
			Methods & IDS$ \uparrow $ & PSNR$ \uparrow $ & SSIM$ \uparrow $ \\
			\hline
			Ours & \textbf{0.84} & \textbf{29.5} & \textbf{0.91}  \\
			HairCLIP~\cite{wei2022hairclip} & 0.45 & 21.6 & 0.74  \\
			StyleCLIP~\cite{patashnik2021styleclip} & 0.43 & 19.6 & 0.72  \\
			TediGAN~\cite{xia2021tedigan} & 0.16 & 22.5 & 0.74  \\
			DiffCLIP~\cite{kim2022diffusionclip} & 0.71 & 26.8 & 0.86  \\
			\hline
		\end{tabular}
	}
    \vspace{1em}
	\caption{Quantitative comparison for irrelevant attributes preservation. IDS denotes identity similarity, PSNR and SSIM are calculated at the intersected non-hair regions before and after editing.}
	\label{tab:textcomparetable}
\end{table}

\noindent\textbf{Comparison with Text-Driven Hair Editing Methods.} We compare HairCLIPv2 with leading text-driven hair editing methods on the CelebA-HQ~\cite{karras2018progressive} testset ($ 2,000 $ images) and follow the evaluation settings of HairCLIP. For HairCLIP~\cite{wei2022hairclip} and StyleCLIP~\cite{patashnik2021styleclip} (``Mapper" version), we first invert using e4e~\cite{tov2021designing} to obtain the latent code for a given real image before performing the editing. For DiffusionCLIP~\cite{kim2022diffusionclip}, we finetune a model for each text description. For both TediGAN~\cite{xia2021tedigan} and our method, the number of optimization iterations is set to $ 200 $. As shown in Figure \ref{fig:textcomparefig} and Table \ref{tab:textcomparetable}, our method accomplishes satisfactory hair editing effects with better naturalness while maximizing the preservation of irrelevant attributes. It is worth noting that, even though HairCLIP and StyleCLIP also have pretty good hair editing capabilities, they cannot preserve the irrelevant attributes very well such as background, identity and clothes. Our method also demonstrates better preservation of the original hair structure when editing only the hair color.

\begin{figure}[tb]
	\begin{center}
		\setlength{\tabcolsep}{0.5pt}
		\begin{tabular}{m{0.3cm}<{\centering}m{1.3cm}<{\centering}m{1.3cm}<{\centering}m{1.3cm}<{\centering}m{1.3cm}<{\centering}m{1.3cm}<{\centering}m{1.3cm}<{\centering}}
			& \scriptsize{Input Image} & \scriptsize{Example} & \scriptsize{Ours} & \scriptsize{HairCLIP} & \scriptsize{StyleCLIP} & \scriptsize{TediGAN}
			\\
			
			\raisebox{0.5cm}{\rotatebox[origin=c]{90}{\footnotesize{{comb over}}}}
			&\includegraphics[width=1.25cm]{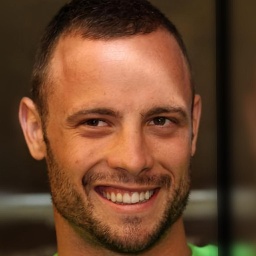}
			&\includegraphics[width=1.25cm]{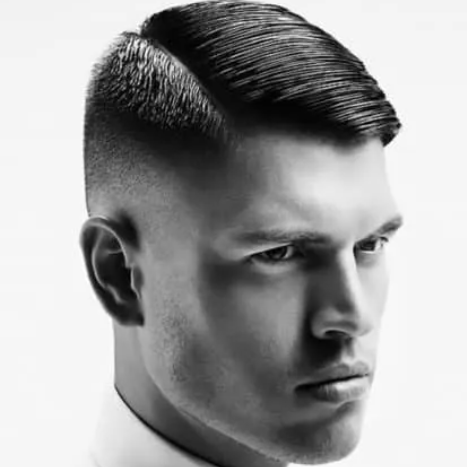}
			&\includegraphics[width=1.25cm]{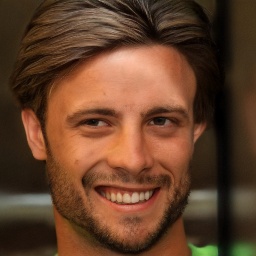}
			&\includegraphics[width=1.25cm]{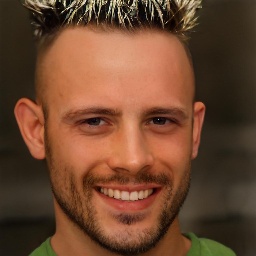}
			&\includegraphics[width=1.25cm]{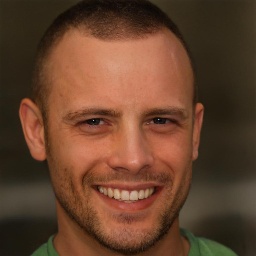}
			&\includegraphics[width=1.25cm]{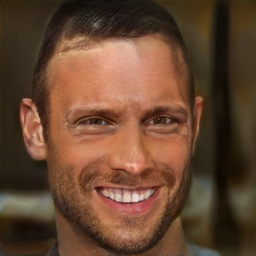}	
			\\

			\raisebox{0.5cm}{\rotatebox[origin=c]{90}{\footnotesize{{curly short}}}}
			&\includegraphics[width=1.25cm]{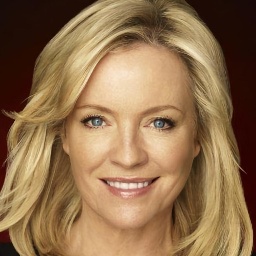}
			&\includegraphics[width=1.25cm]{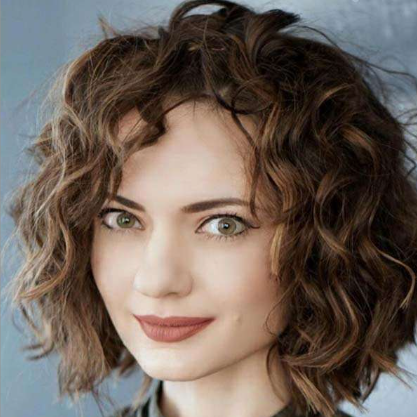}
			&\includegraphics[width=1.25cm]{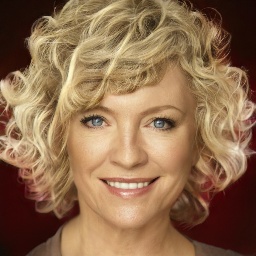}
			&\includegraphics[width=1.25cm]{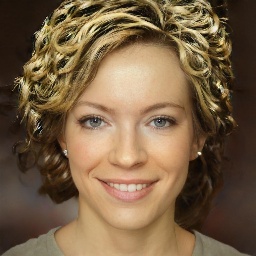}
			&\includegraphics[width=1.25cm]{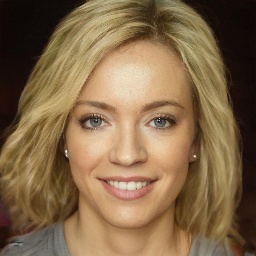}
			&\includegraphics[width=1.25cm]{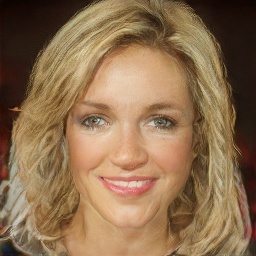}	
			\\
			
			\raisebox{0.35cm}{\rotatebox[origin=c]{90}{\footnotesize{{devilock}}}}
			&\includegraphics[width=1.25cm]{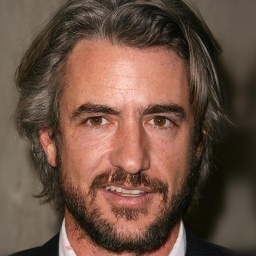}
			&\includegraphics[width=1.25cm]{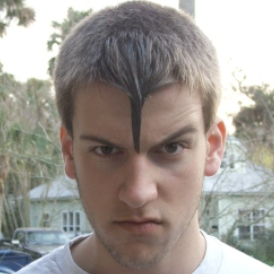}
			&\includegraphics[width=1.25cm]{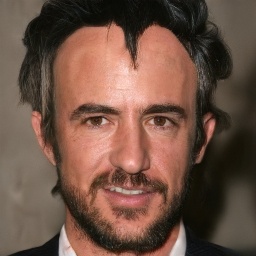}
			&\includegraphics[width=1.25cm]{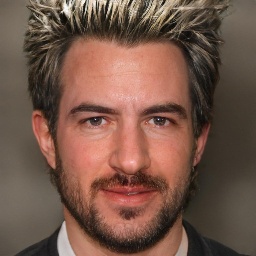}
			&\includegraphics[width=1.25cm]{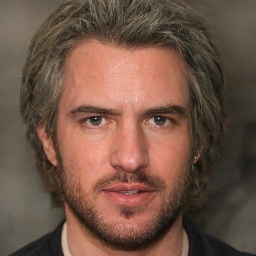}
			&\includegraphics[width=1.25cm]{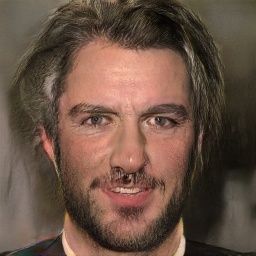}
			\\

		\end{tabular}
	\end{center}
	\caption{Visual comparison with HairCLIP~\cite{wei2022hairclip},  StyleCLIP-Optimization~\cite{patashnik2021styleclip} and TediGAN~\cite{xia2021tedigan} under any description setting. We additionally provide an example image for each description for better comparison.}
	\label{fig:anytextfig}
\end{figure}

For arbitrary hair editing word scenarios, the only methods that are instantly feasible without retraining the model are HairCLIP, StyleCLIP (``Optimization" version), and TediGAN. As shown in Figure \ref{fig:anytextfig}, our method perform much better at such cases. In contrast, HairCLIP can only produce plausible results for text (``\textit{Curly Short Hairstyle}'') similar to the training texts, while all other methods struggle to produce reasonable editing effects.

\begin{figure*}[tb]
	\begin{center}
		\setlength{\tabcolsep}{0.5pt}
		\begin{tabular}{m{1.85cm}<{\centering}m{1.85cm}<{\centering}m{1.85cm}<{\centering}|m{1.85cm}<{\centering}m{1.85cm}<{\centering}m{1.85cm}<{\centering}m{1.85cm}<{\centering}m{1.85cm}<{\centering}m{1.85cm}<{\centering}}
			\scriptsize{Input Image} & \scriptsize{Hairstyle Ref} & \scriptsize{Color Ref} & \scriptsize{Ours} & \scriptsize{HairCLIP~\cite{wei2022hairclip}} & \scriptsize{LOHO~\cite{saha2021loho}} & \scriptsize{Barbershop~\cite{zhu2021barbershop}} & \scriptsize{SYH~\cite{kim2022style}} & \scriptsize{MichiGAN~\cite{tan2020michigan}}
			\\
			
			\includegraphics[width=1.8cm]{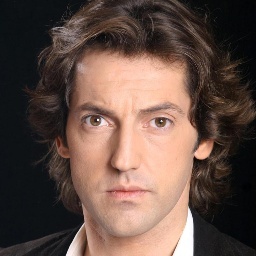}
			&\includegraphics[width=1.8cm]{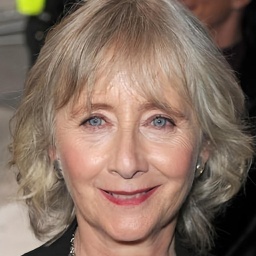}
			&\includegraphics[width=1.8cm]{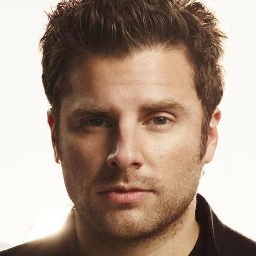}
			&\includegraphics[width=1.8cm]{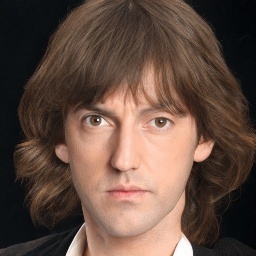}
			&\includegraphics[width=1.8cm]{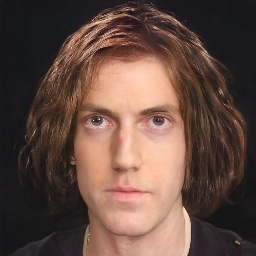}
			&\includegraphics[width=1.8cm]{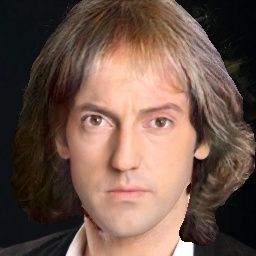}
			&\includegraphics[width=1.8cm]{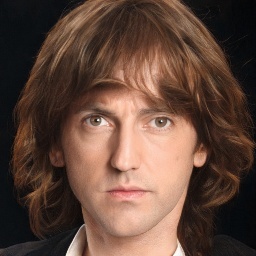}
			&\includegraphics[width=1.8cm]{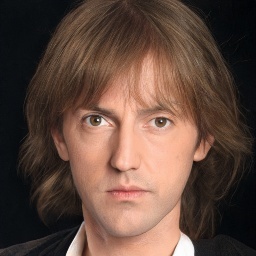}
			&\includegraphics[width=1.8cm]{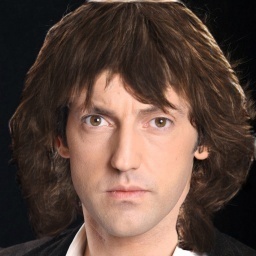}			
			\\
			
			\includegraphics[width=1.8cm]{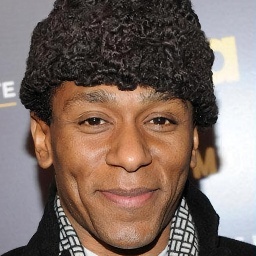}
			&\includegraphics[width=1.8cm]{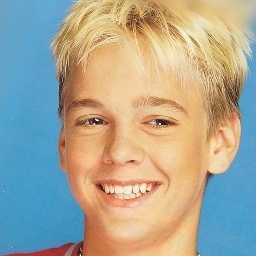}
			&\includegraphics[width=1.8cm]{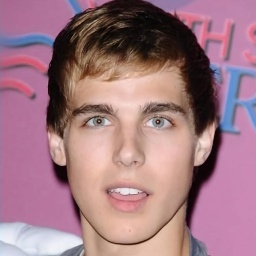}
			&\includegraphics[width=1.8cm]{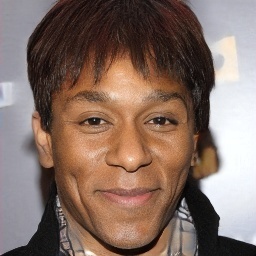}
			&\includegraphics[width=1.8cm]{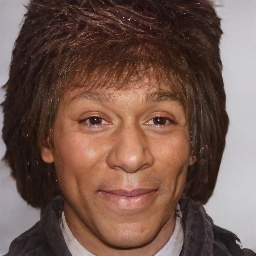}
			&\includegraphics[width=1.8cm]{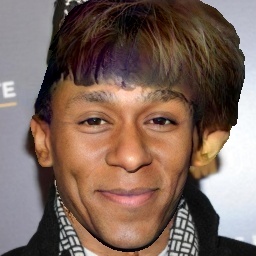}
			&\includegraphics[width=1.8cm]{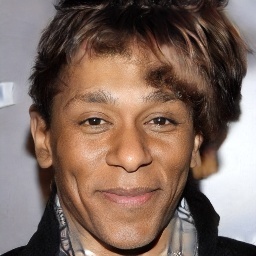}
			&\includegraphics[width=1.8cm]{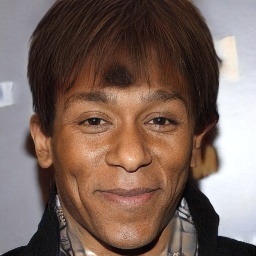}
			&\includegraphics[width=1.8cm]{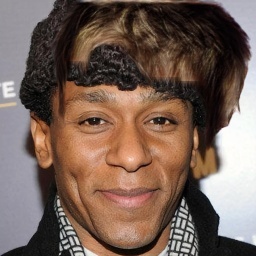}			
			\\

		\end{tabular}
	\end{center}
	\caption{Visual comparison with HairCLIP~\cite{wei2022hairclip},  LOHO~\cite{saha2021loho}, Barbershop~\cite{zhu2021barbershop}, SYH~\cite{kim2022style} and MichiGAN~\cite{tan2020michigan} on hair transfer. Only our method and SYH can accomplish unaligned hair transfer while keeping irrelevant attributes unmodified.} 
	\label{fig:transfercomparefig}
\end{figure*}

\noindent\textbf{Comparison with Hair Transfer Methods.} We compare with state-of-the-art methods on hair transfer tasks. Among the $ 2,000 $ images of the CelebA-HQ testset, the first $ 666 $ are set as the input images, the middle $ 666 $ are set as the hairstyle reference images, and the last $ 666 $ are set as the hair color reference images. As shown in Figure \ref{fig:transfercomparefig}, when the hairstyle reference image is broadly aligned with the input image (first row), most methods yield plausible results. However, when not aligned (second row), only our method and SYH~\cite{kim2022style} are able to perform a more consistent hair transfer, which is achieved by introducing the pose alignment loss during the transfer to ensure that the facial shape and pose of the reference image are consistent with the source image. Compared to SYH, we achieve comparable hair transfer results, but support text, sketch, and other interactions beyond hair transfer.

\noindent\textbf{Comparison with Local Hair Editing Methods.} In terms of sketch-based local editing, we compare with the SOTA methods MichiGAN~\cite{tan2020michigan} and SketchSalon~\cite{xiao2021sketchhairsalon}. MichiGAN~\cite{tan2020michigan} uses user-drawn sketches to modify the orientation map to accomplish local hair editing. SketchSalon~\cite{xiao2021sketchhairsalon} trains a sketch-to-hair conditional translation network, with an additional soft alpha matte used to facilitate more natural blending. To generalize SketchSalon to local editing, we utilize the same mask as our method instead of a soft alpha matte, and the input sketch is colored as the average color within the mask area. As shown in Figure \ref{fig:sketchcomparefig}, MichiGAN struggles to perform satisfactory local editing and the reconstruction of other non-editing hair areas is slightly worse. Even ignoring the obvious blending artifacts, the local hair texture generated by SketchSalon is not in harmony with the surrounding hair. Compared to these two methods, our approach not only achieves satisfactory local editing but also better maintains the non-editing regions.

\begin{figure}[tb]
	\begin{center}
		\setlength{\tabcolsep}{0.5pt}
		\begin{tabular}{m{1.55cm}<{\centering}m{1.55cm}<{\centering}|m{1.55cm}<{\centering}m{1.55cm}<{\centering}m{1.55cm}<{\centering}}
			\scriptsize{Input Image} & \scriptsize{Input Sketch} & \scriptsize{Ours} & \scriptsize{MichiGAN} & \scriptsize{SketchSalon}
			\\
			
			\includegraphics[width=1.5cm]{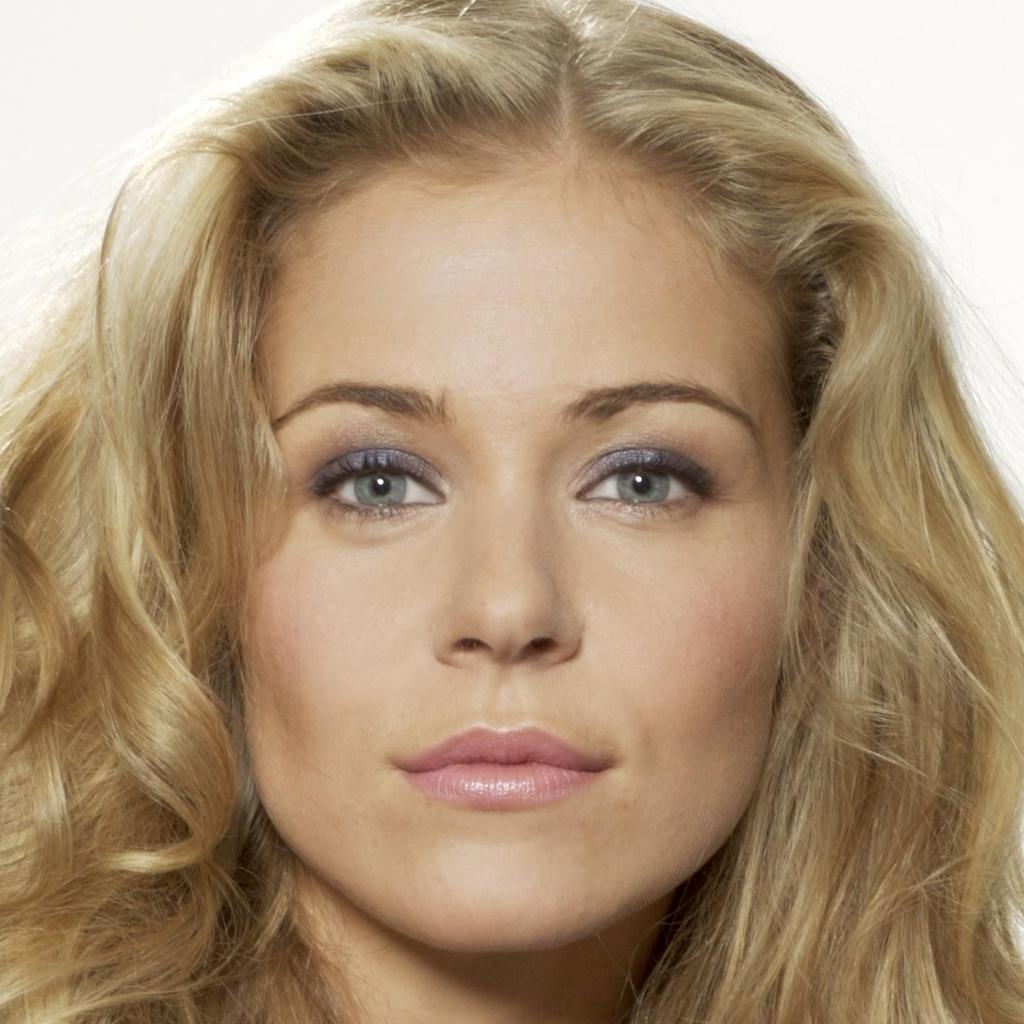}
			&\includegraphics[width=1.5cm]{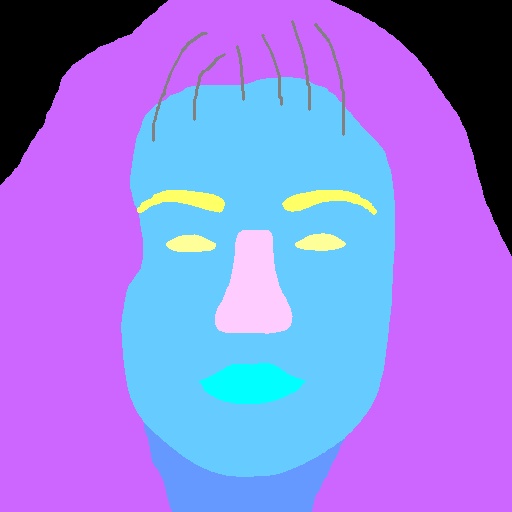}
			&\includegraphics[width=1.5cm]{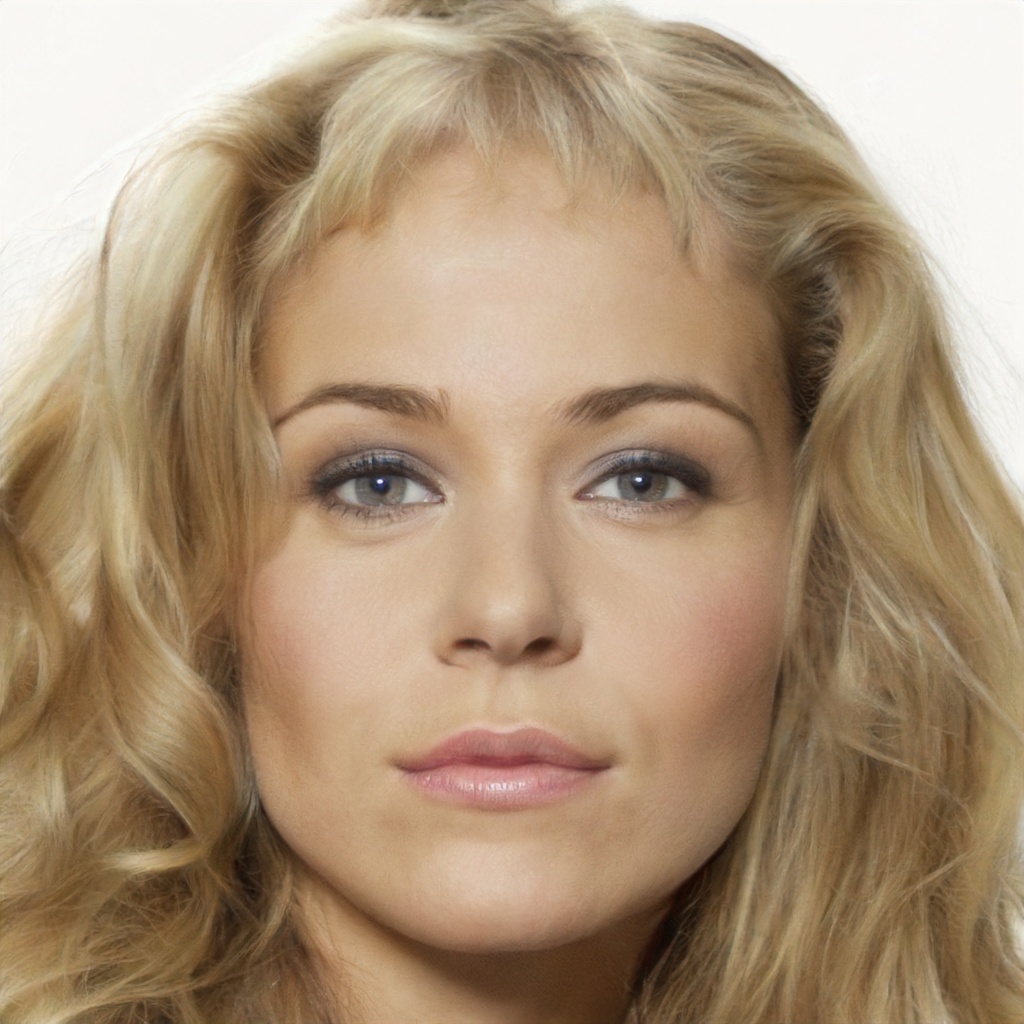}
			&\includegraphics[width=1.5cm]{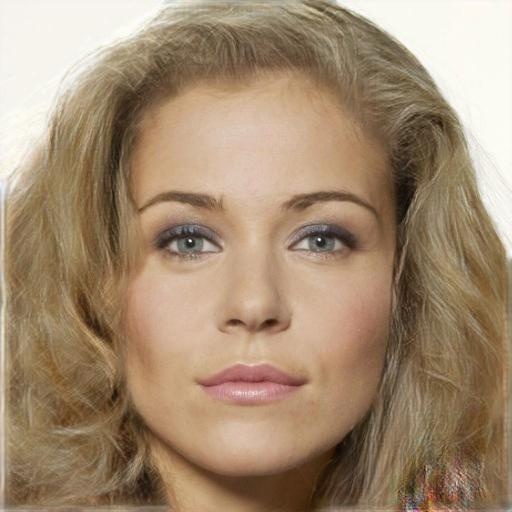}
			&\includegraphics[width=1.5cm]{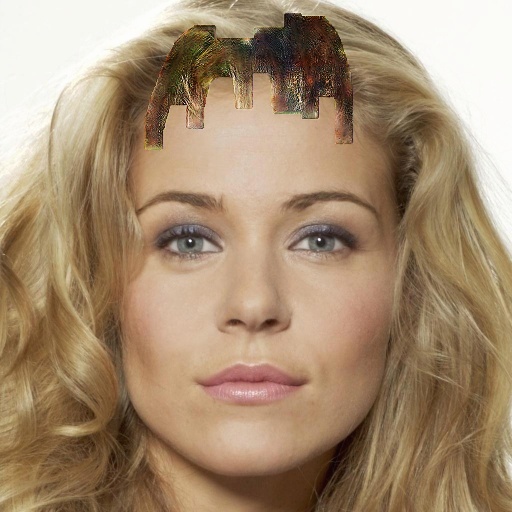}	
			\\
			
			\includegraphics[width=1.5cm]{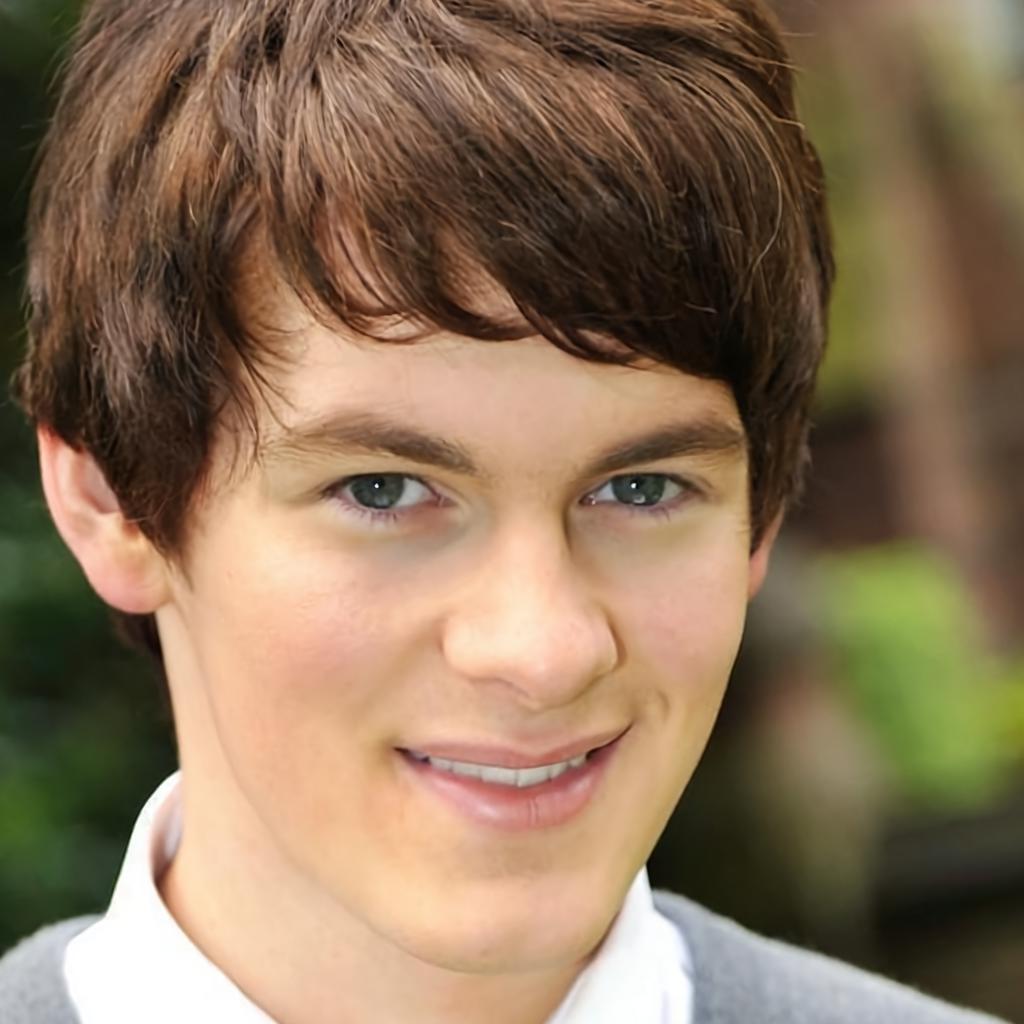}
			&\includegraphics[width=1.5cm]{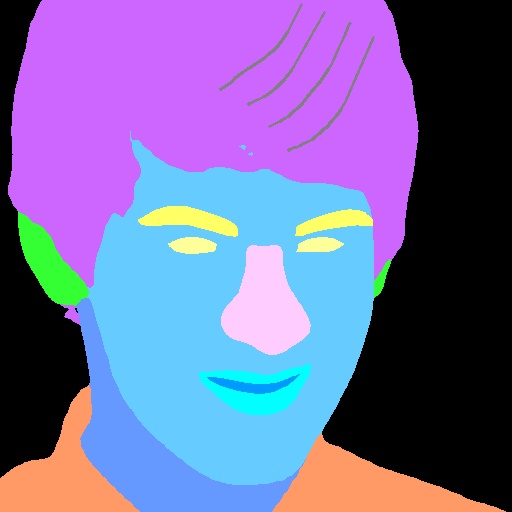}
			&\includegraphics[width=1.5cm]{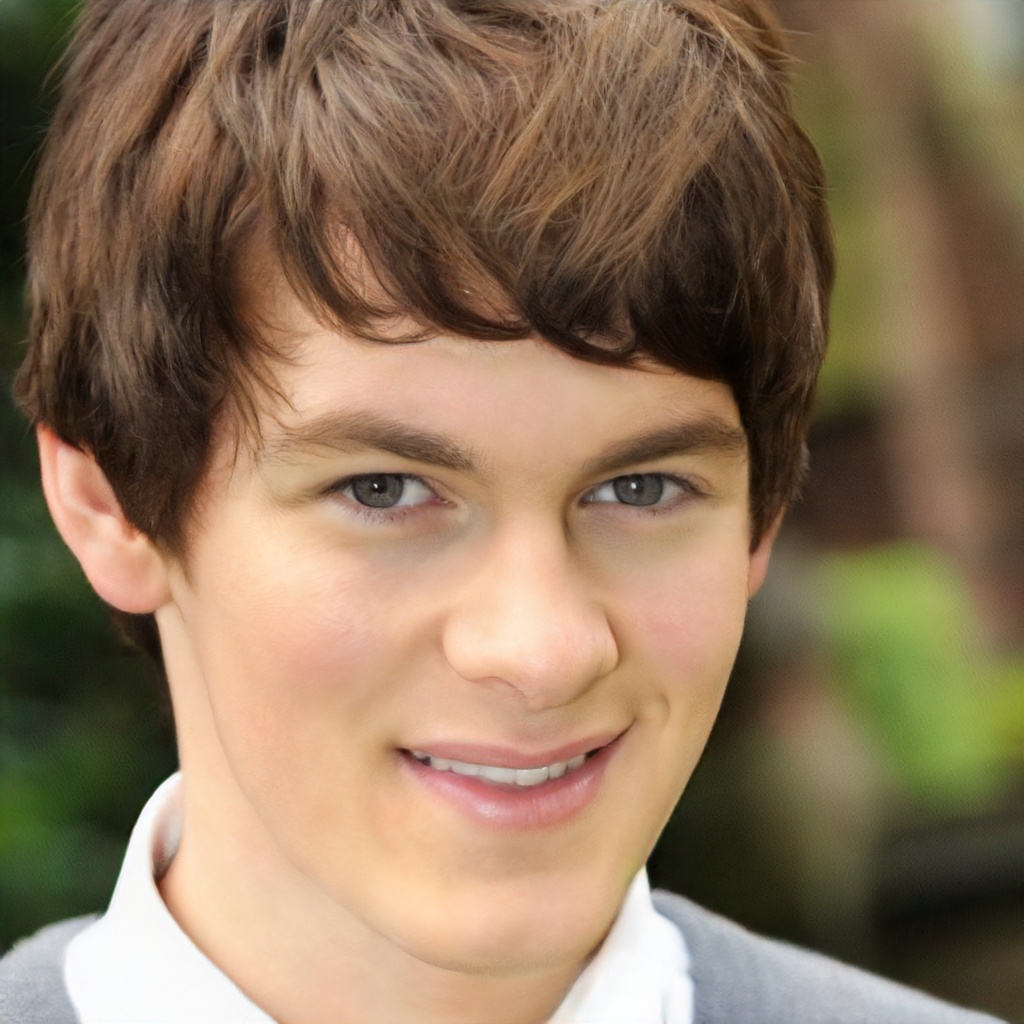}
			&\includegraphics[width=1.5cm]{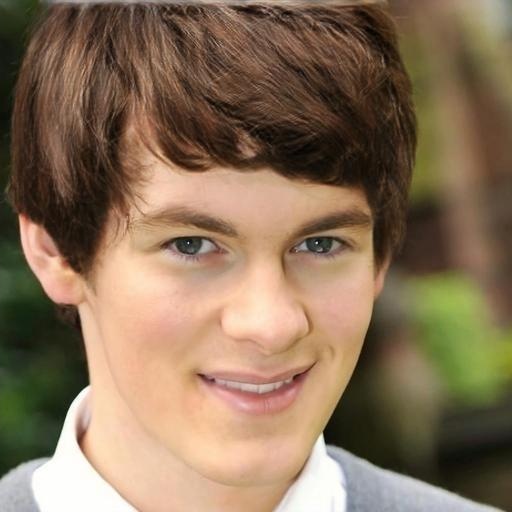}
			&\includegraphics[width=1.5cm]{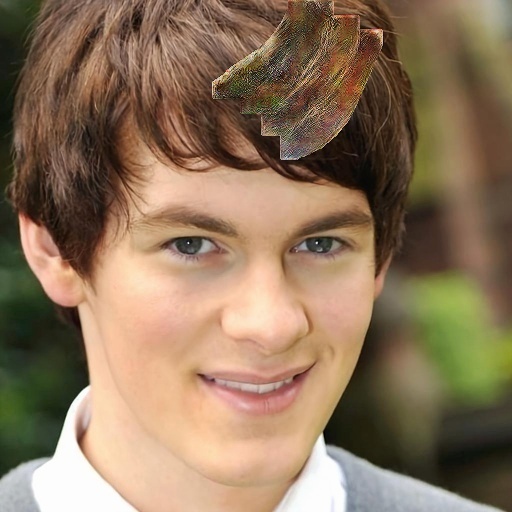}	
			\\
			
		\end{tabular}
	\end{center}
	\caption{Comparison with MichiGAN~\cite{tan2020michigan} \& SketchSalon~\cite{xiao2021sketchhairsalon} on sketch-based local hair editing. We provide sketches in the facial parsing map for better visualization.} 
	\label{fig:sketchcomparefig}
\end{figure}

\noindent\textbf{Comparison with Cross-Modal Hair Editing Methods.} To the best of our knowledge, the only method that supports multimodal conditions to complete hairstyle and hair color editing is HairCLIP~\cite{wei2022hairclip}. As the comparison shown in Figure \ref{fig:crossmodalcomparefig}, our approach not only perfectly prevents irrelevant attributes (identity, background, etc.) from being modified, but also achieves higher-quality editing effects. Moreover, HairCLIP allows only text and reference image while our method additionally supports sketch, mask, and RGB values. More diverse and comprehensive interactive editing results are shown in Figure \ref{fig:teaser} and supplementary materials.

\begin{figure}[t]
	\centering
	\includegraphics[width=\columnwidth]{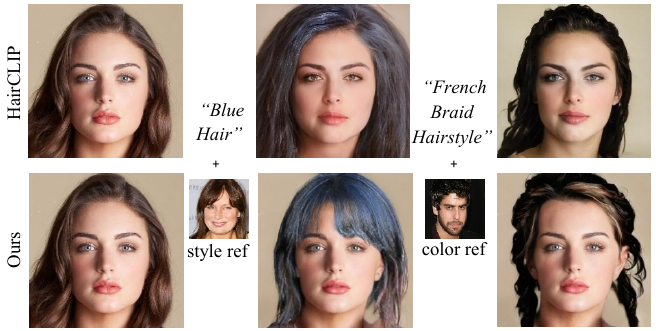}
    \vspace{0.05em}
	\caption{Qualitative comparison with HairCLIP on cross-modal conditional input. Our approach shows better editing effects \& excellent preservation of irrelevant attributes.}
	\label{fig:crossmodalcomparefig}
\end{figure}

\begin{table*}[t]
	\centering
	\small
	\setlength{\tabcolsep}{2.5pt}{
		\begin{tabular}{l|ccccc|cccccc|ccc|cc}
			\hline
			\multicolumn{1}{ c }{{}}	& \multicolumn{5}{ c }{{\small{Text-Driven}}} & \multicolumn{6}{ c }{{\small{Hair Transfer}}} & \multicolumn{3}{ c }{{\small{Sketch-Based}}} & \multicolumn{2}{ c }{{\small{Cross-Modal}}}\\
			\hline
			\small{Metrics} & \small{Ours} & \small{\cite{wei2022hairclip}} & \small{\cite{patashnik2021styleclip}} & \small{\cite{xia2021tedigan}} & \small{\cite{kim2022diffusionclip}} & \small{Ours} & \small{\cite{wei2022hairclip}} & \small{\cite{saha2021loho}} & \small{\cite{zhu2021barbershop}} & \small{\cite{kim2022style}}& \small{\cite{tan2020michigan}} & \small{Ours} & \small{\cite{tan2020michigan}} & \small{\cite{xiao2021sketchhairsalon}} & \small{Ours} & \small{\cite{wei2022hairclip}}\\
			\hline
			Accuracy  & \textbf{\small{41.5\%}} & \small{32.3\%} & \small{22.5\%} & \small{1.0\%} & \small{2.8\%} & \small{28.0\%} & \small{2.3\%} & \small{4.8\%} & \textbf{\small{29.3\%}} & \small{28.5\%} & \small{7.3\%} & \textbf{\small{76.8\%}} & \small{22.0\%} & \small{1.3\%} & \textbf{\small{82.8\%}} & \small{17.3\%} \\
			Preservation & \textbf{\small{81.0\%}} & \small{5.3\%} & \small{3.3\%} & \small{0.3\%} & \small{10.3\%} & \textbf{\small{32.8\%}} & \small{2.8\%} & \small{8.3\%} & \small{15.3\%} & \small{26.3\%} & \small{14.8\%} & \textbf{\small{62.0\%}} & \small{33.3\%} & \small{4.8\%} & \textbf{\small{94.0\%}} & \small{6.0\%} \\
			Naturalness & \textbf{\small{46.8\%}} & \small{25.5\%} & \small{22.0\%} & \small{1.8\%} & \small{4.0\%} & \small{26.5\%} & \small{9.5\%} & \small{2.5\%} & \small{22.3\%} & \textbf{\small{35.0\%}} & \small{4.3\%} & \textbf{\small{60.5\%}} & \small{38.0\%} & \small{1.5\%} & \textbf{\small{65.3\%}} & \small{34.8\%} \\
			\hline
		\end{tabular}
	}
    \vspace{0.2em}
	\caption{User study on text-driven image manipulation, hair transfer, sketch-based local hair editing and cross-modal hair editing methods. Accuracy denotes the manipulation accuracy for given conditional inputs, Preservation indicates the ability to preserve irrelevant regions and Naturalness denotes the visual realism of the manipulated image.}	
	\label{tab:userstudy}
\end{table*}

\noindent\textbf{User Study.} For the above four types of comparisons, we recruit $ 20 $ volunteers with computer vision-related research backgrounds to execute a comprehensive user study. We randomly selected 20 groups of results from each experiment to form 80 test samples in total. The order of the different methods in each test sample is randomly shuffled. For each test sample, volunteers are asked to select the best option in terms of manipulation accuracy, irrelevant attribute preservation, and visual naturalness, respectively. As shown in Table \ref{tab:userstudy}, our method outperforms the baseline methods for most cases, except comparable results to Barbershop~\cite{zhu2021barbershop} and SYH~\cite{kim2022style} in the hair transfer setting. But our irrelevant attribute preservation performs best because of our hair color feature space blending mechanism, as demonstrated in Figure \ref{fig:transfercomparefig}. It is worth mentioning that our goal is not to improve the performance of hair transfer, but to design a unified system that supports various hair editing and hair transfer tasks. Therefore, performing comparably with the state-of-the-art methods on the hair transfer task is acceptable.

\subsection{Ablation Analysis}

\begin{figure}[tb]
	\begin{center}
		\setlength{\tabcolsep}{0.5pt}
		\begin{tabular}{m{2cm}<{\centering}m{2cm}<{\centering}m{2cm}<{\centering}m{2cm}<{\centering}}
			\scriptsize{Input Image} & \scriptsize{Mean ($ \mathcal{W} $)} & \scriptsize{Random ($ \mathcal{W} $)} & \scriptsize{Inverted ($ \mathcal{W+} $)}
			\\
			
			\includegraphics[width=1.9cm]{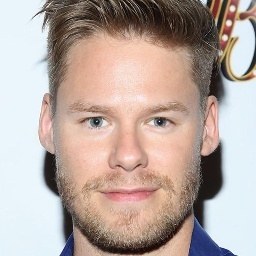}
			&\includegraphics[width=1.9cm]{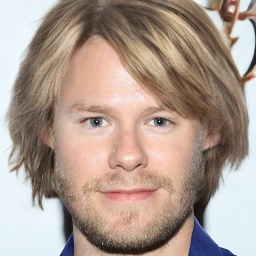}
			&\includegraphics[width=1.9cm]{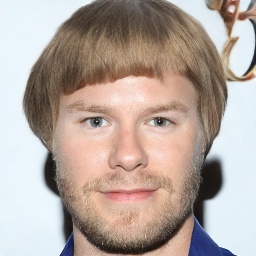}
			&\includegraphics[width=1.9cm]{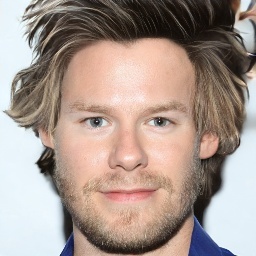}	
			\\
			
			\includegraphics[width=1.9cm]{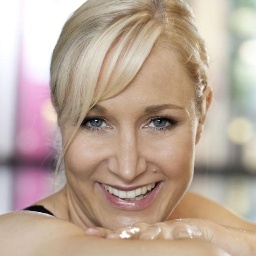}
			&\includegraphics[width=1.9cm]{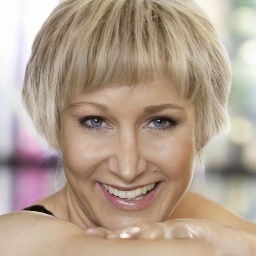}
			&\includegraphics[width=1.9cm]{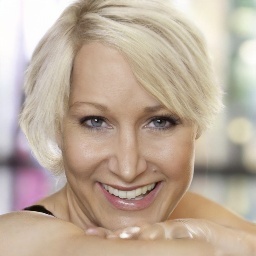}
			&\includegraphics[width=1.9cm]{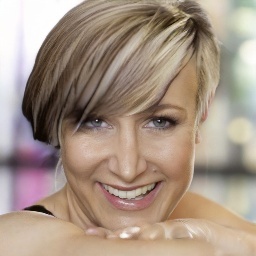}	
			\\			
			
		\end{tabular}
	\end{center}
	\caption{Ablation of different starting point settings for text proxy. The text description is ``\textit{Bob Cut Hairstyle}''.} 
	\label{fig:start_ablation}
\end{figure}

\noindent\textbf{Importance of Initialization Strategy for Text Proxy Optimization.} To justify our optimization starting point strategy for text proxy, we ablate three different optimization starting points: mean latent code, random sampling from $ \mathcal{W} $ space, and inverted latent code of the input image within $ \mathcal{W+} $ space. All other settings remain the same. As illustrated in Figure \ref{fig:start_ablation}, the first two settings are more likely to complete high-quality editing, while the last one fails because the starting point deviates from the more suitable editing region. This may explain why  StyleCLIP (``Optimization" version) and TediGAN struggle in performing hairstyle editing. Our strategy of randomly initializing the latent code around the mean within $ \mathcal{W} $ space enjoys both good editability and diversity of the generated results.

\begin{figure}[tb]
	\begin{center}
		\setlength{\tabcolsep}{0.5pt}
		\begin{tabular}{m{1.55cm}<{\centering}m{1.55cm}<{\centering}|m{1.55cm}<{\centering}m{1.55cm}<{\centering}m{1.55cm}<{\centering}}
			\scriptsize{Input Image} & \scriptsize{Input Sketch} & \scriptsize{Ours} & \scriptsize{Orientation} & \scriptsize{LPIPS}
			\\
			
			\includegraphics[width=1.5cm]{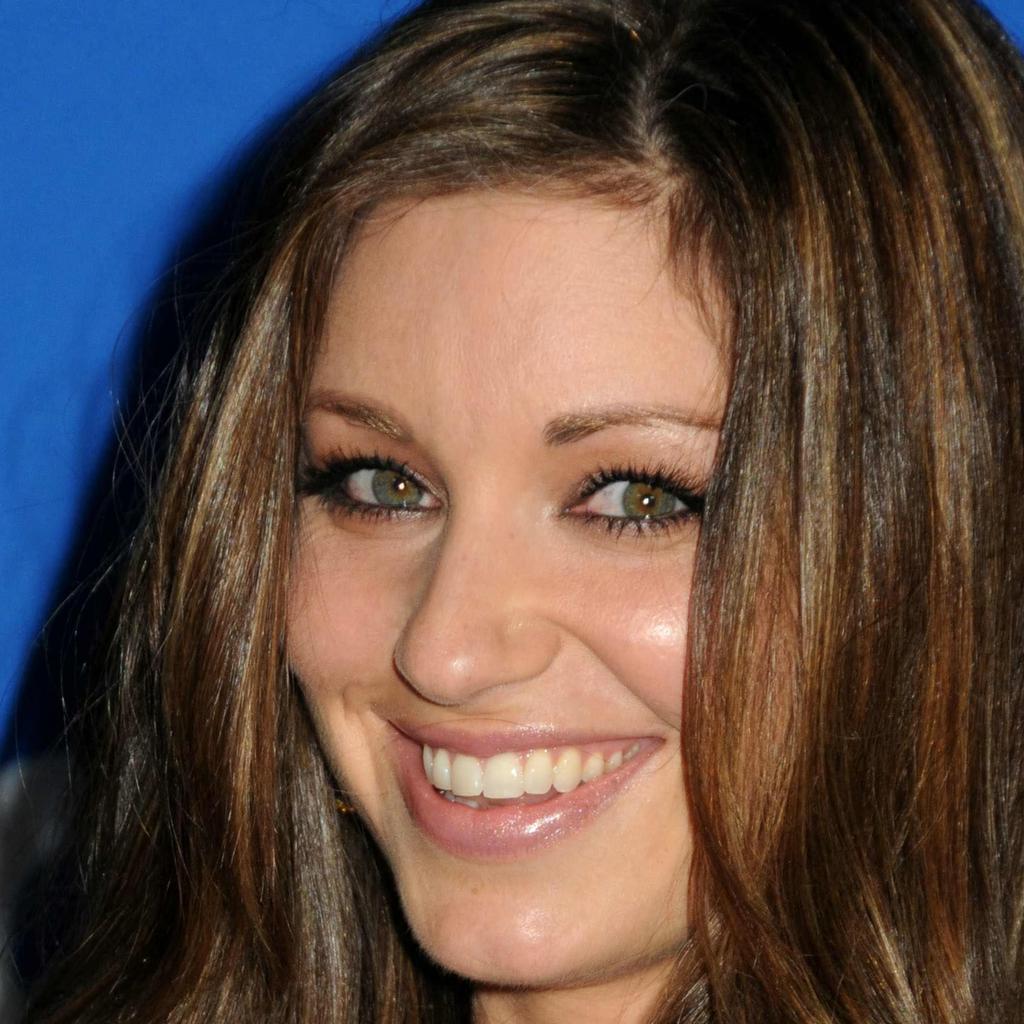}
			&\includegraphics[width=1.5cm]{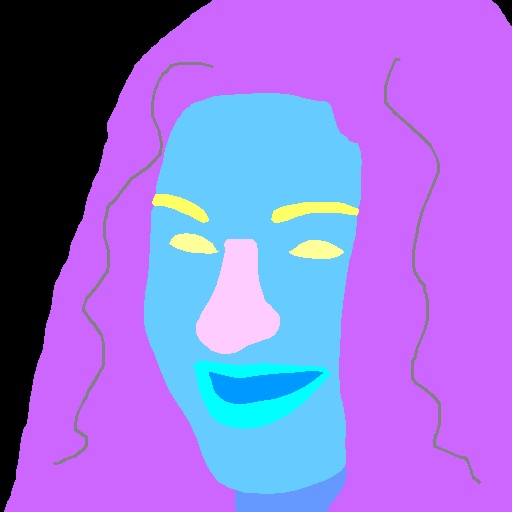}
			&\includegraphics[width=1.5cm]{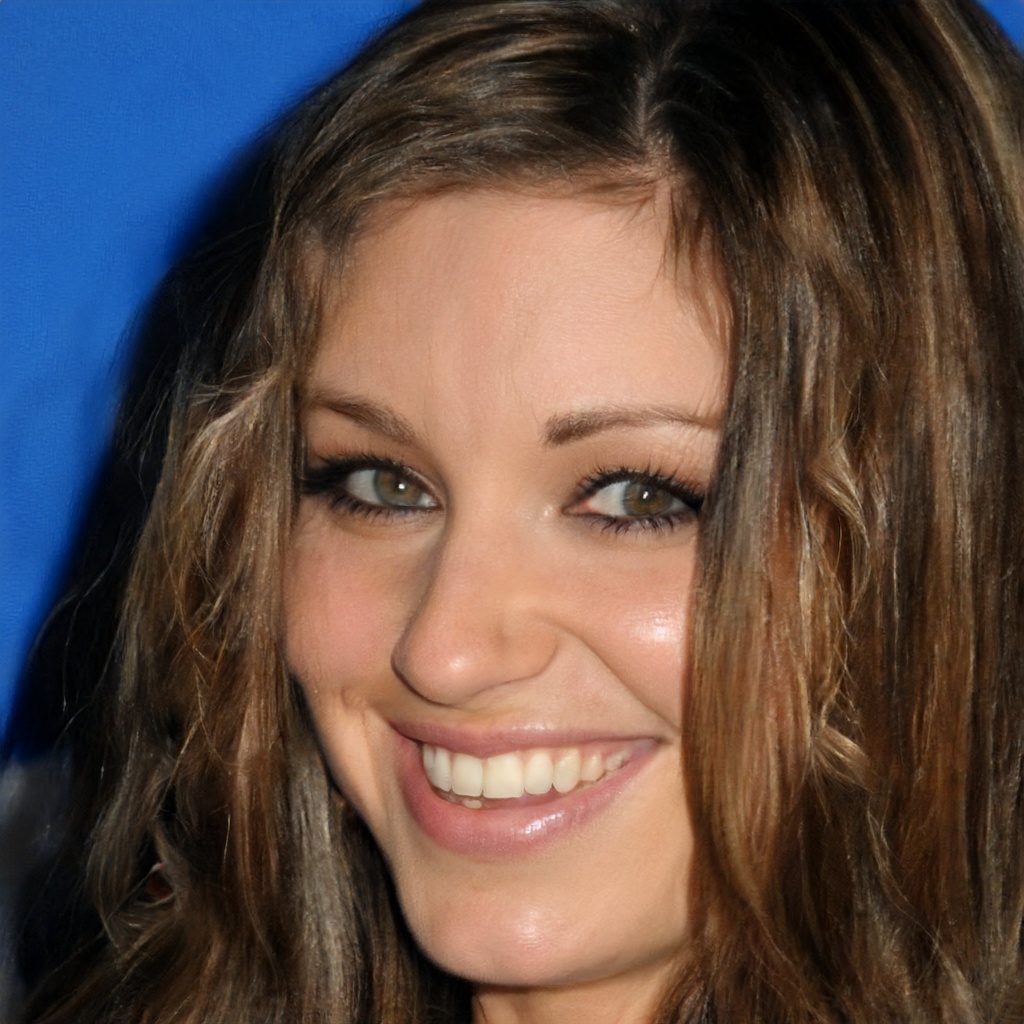}
			&\includegraphics[width=1.5cm]{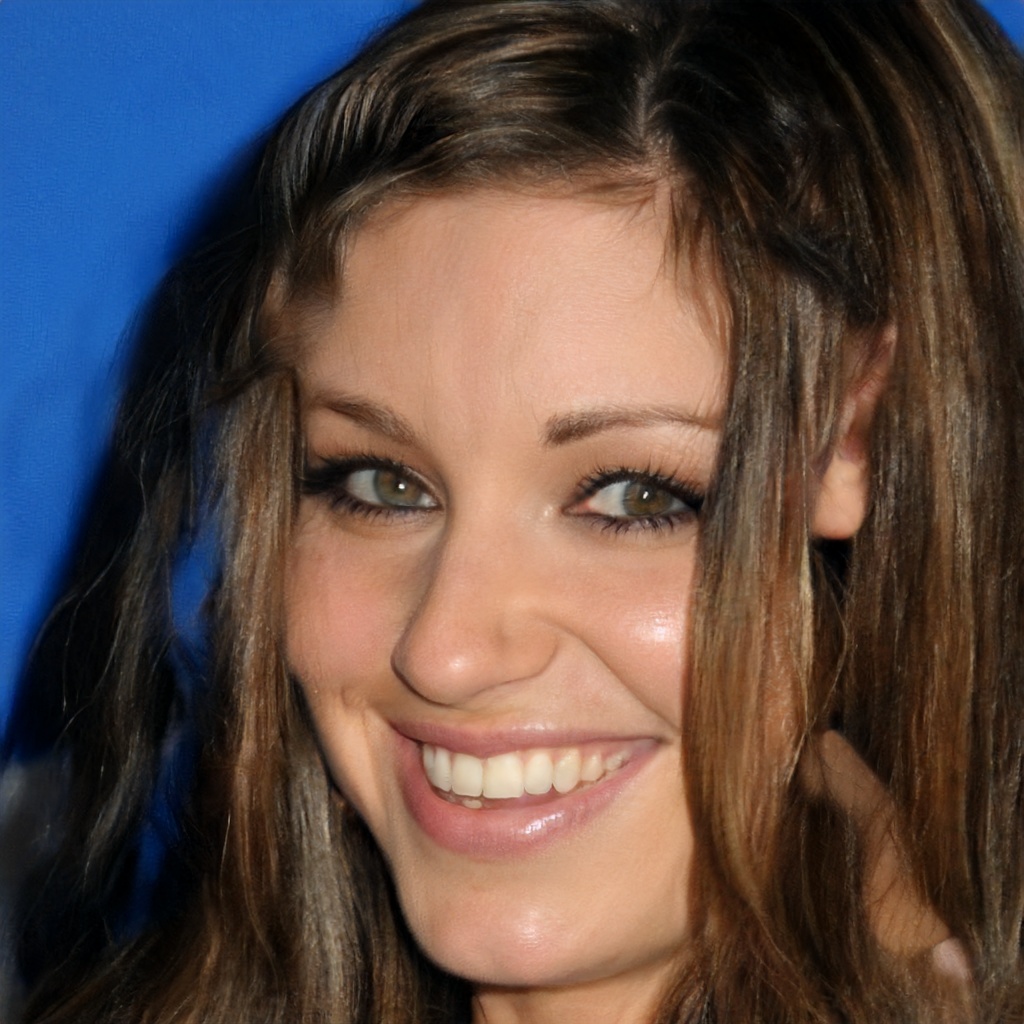}
			&\includegraphics[width=1.5cm]{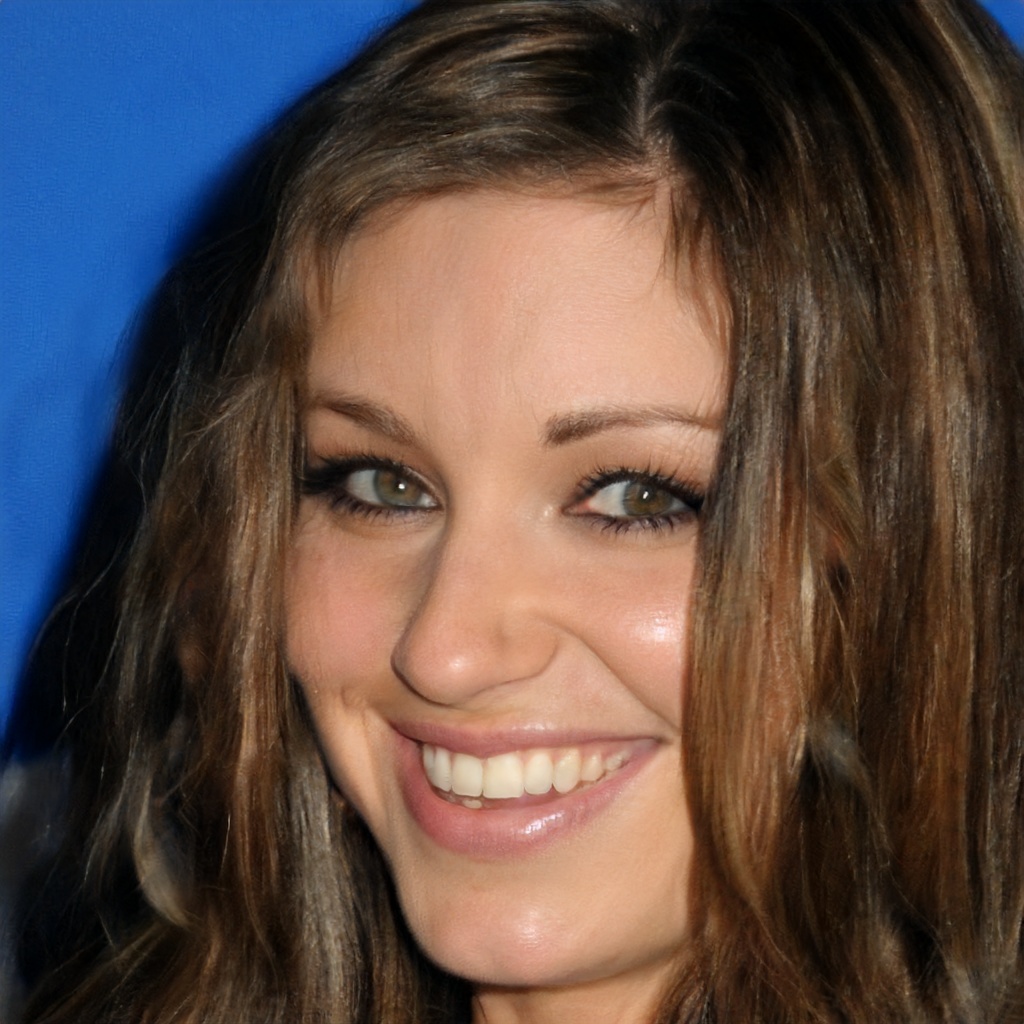}	
			\\
			
			\includegraphics[width=1.5cm]{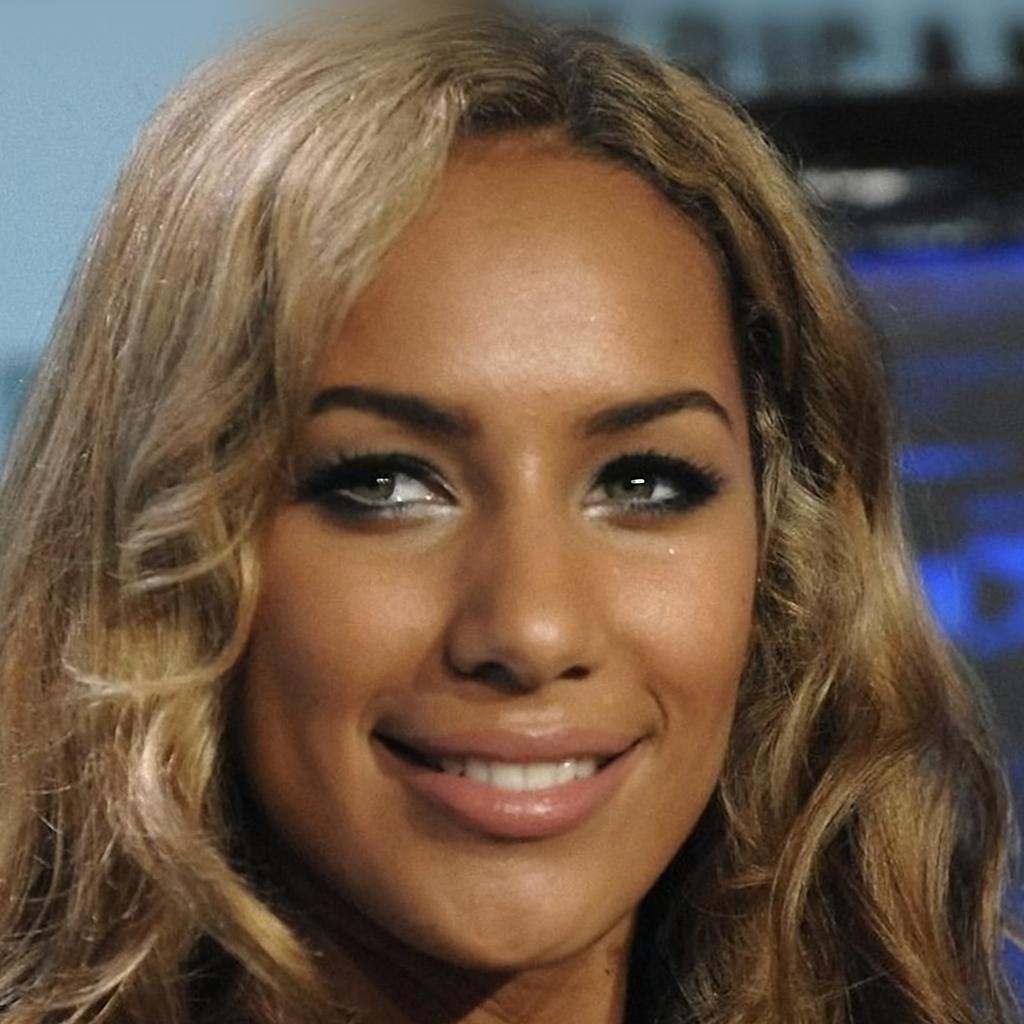}
			&\includegraphics[width=1.5cm]{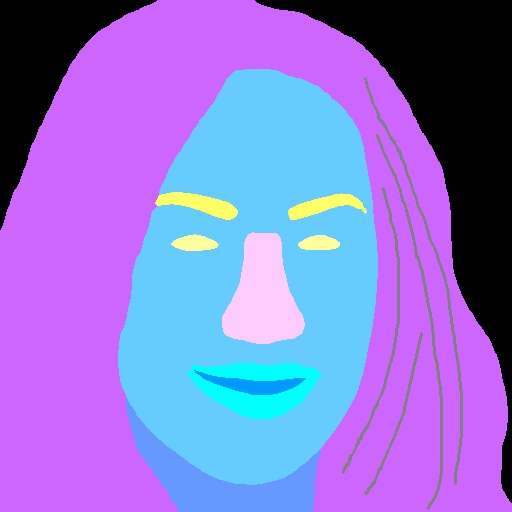}
			&\includegraphics[width=1.5cm]{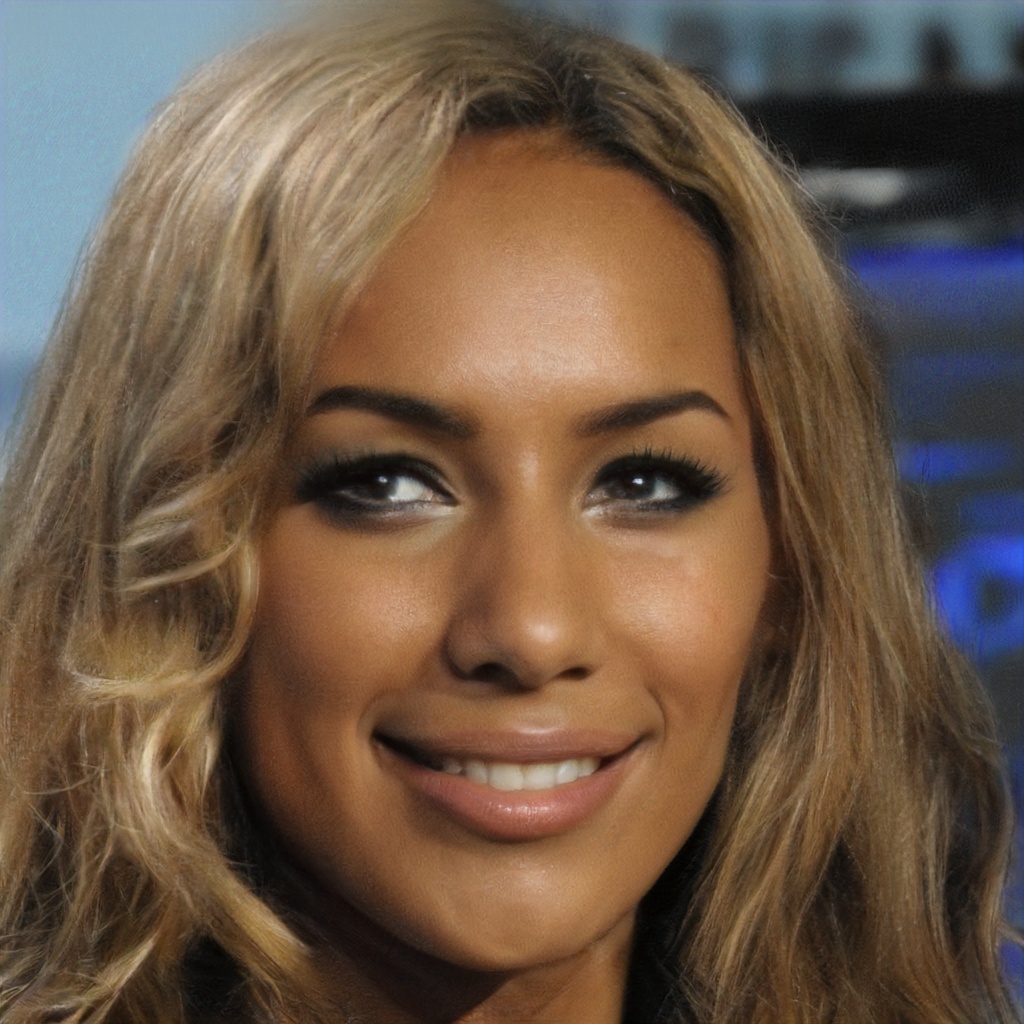}
			&\includegraphics[width=1.5cm]{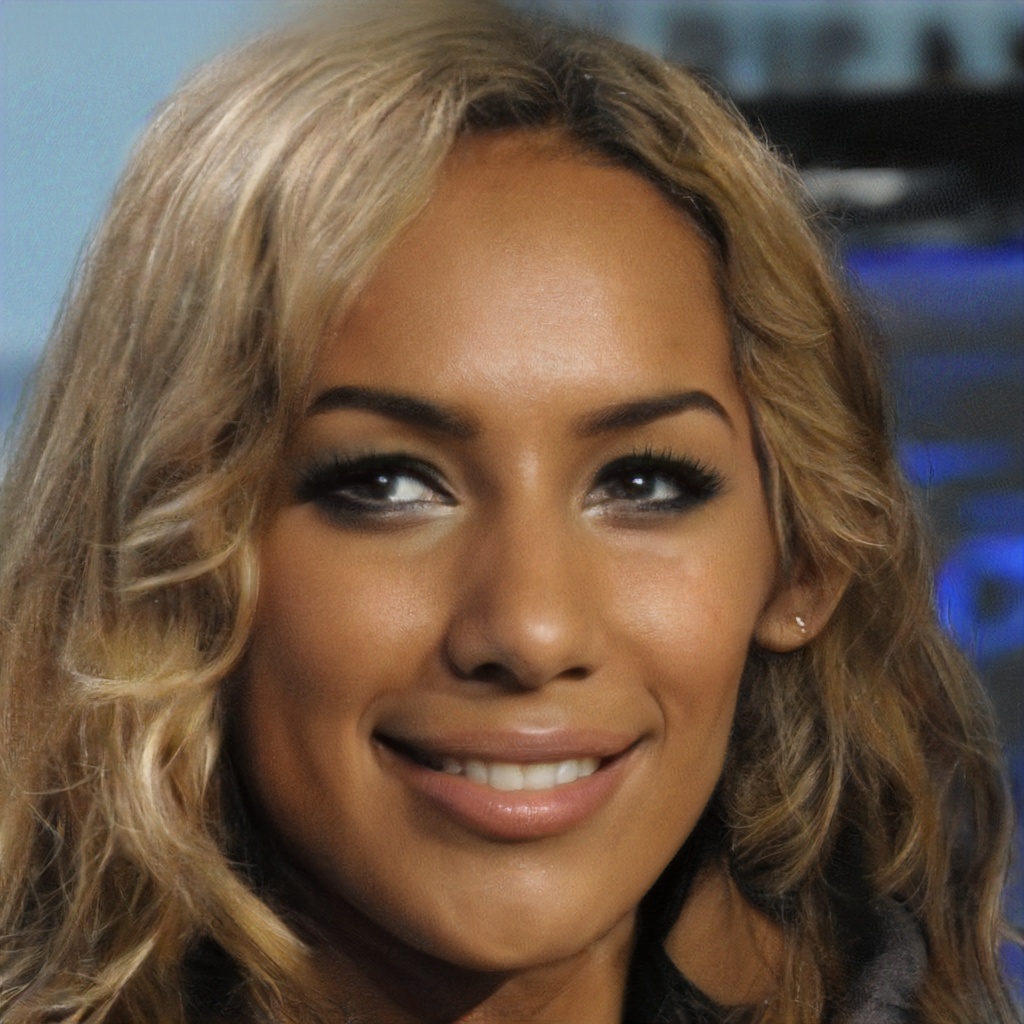}
			&\includegraphics[width=1.5cm]{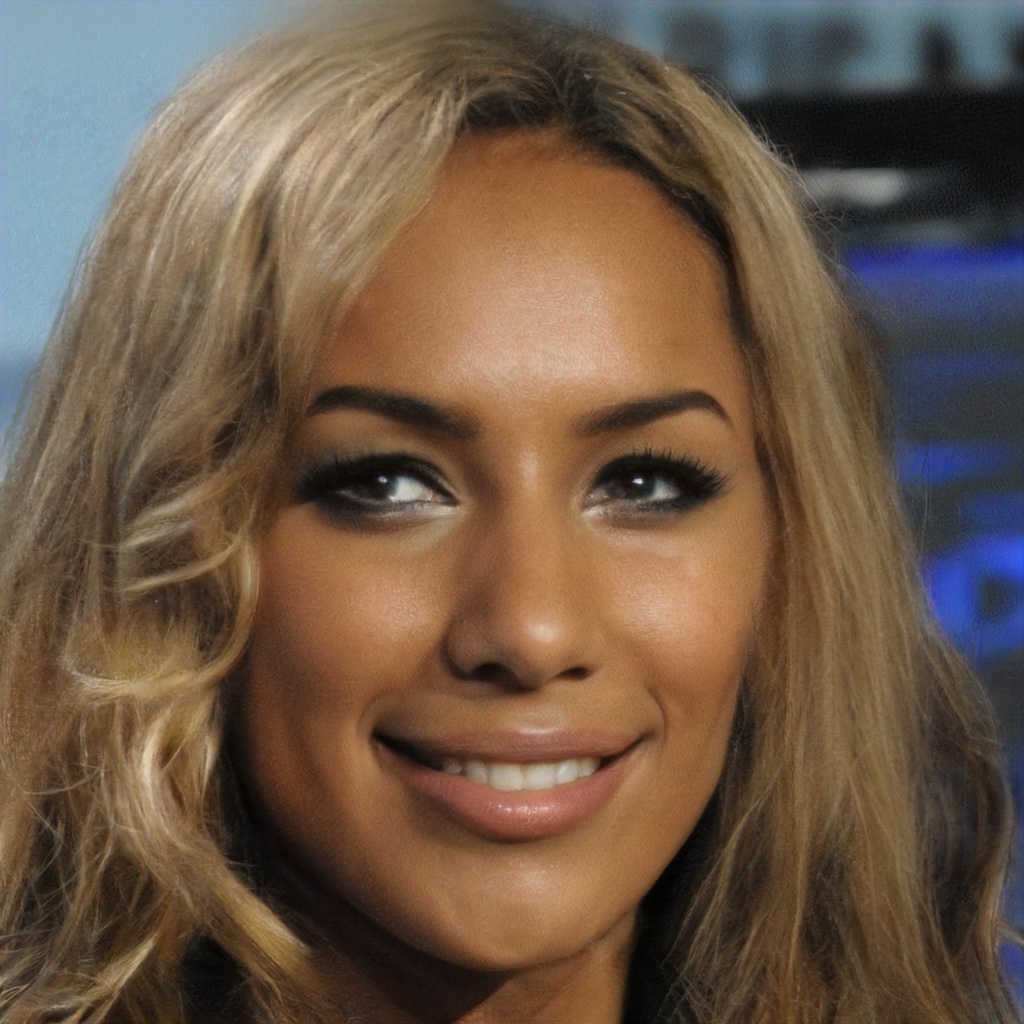}	
			\\
			
		\end{tabular}
	\end{center}
	\caption{Ablation on sketch proxy generation design. We provide sketches drawn in the facial parsing map for better visualization.} 
	\label{fig:sketchabltionfig}
\end{figure}

\noindent\textbf{Superiority of Sketch Proxy Generation Design.} To generate sketch proxy, we ablate another two possible ways. The first way is to constrain the orientation field of the sketch proxy to be similar to that of user's sketch in the drawing region with the orientation loss used in MichiGAN\cite{tan2020michigan}. The second way is to first obtain the image corresponding to the sketch through the conditional translation network (for simplicity, we choose sketch2hair here), and then constrain the sketch proxy to be similar to it in the drawing area by LPIPS~\cite{zhang2018unreasonable} loss during the optimization process. The comparison is shown in Figure \ref{fig:sketchabltionfig}. Only our method and the LPIPS optimization-based version are feasible. However, our method requires only a single feed forward, which is more efficient.

\begin{figure}[tb]
	\begin{center}
		\setlength{\tabcolsep}{0.5pt}
		\begin{tabular}{m{1.55cm}<{\centering}m{1.55cm}<{\centering}m{1.55cm}<{\centering}|m{1.55cm}<{\centering}m{1.55cm}<{\centering}}
			\scriptsize{Input Image} & \scriptsize{Hairstyle Ref} & \scriptsize{Input Sketch} & \scriptsize{Feature} & \scriptsize{Latent Code}
			\\
			
			\includegraphics[width=1.5cm]{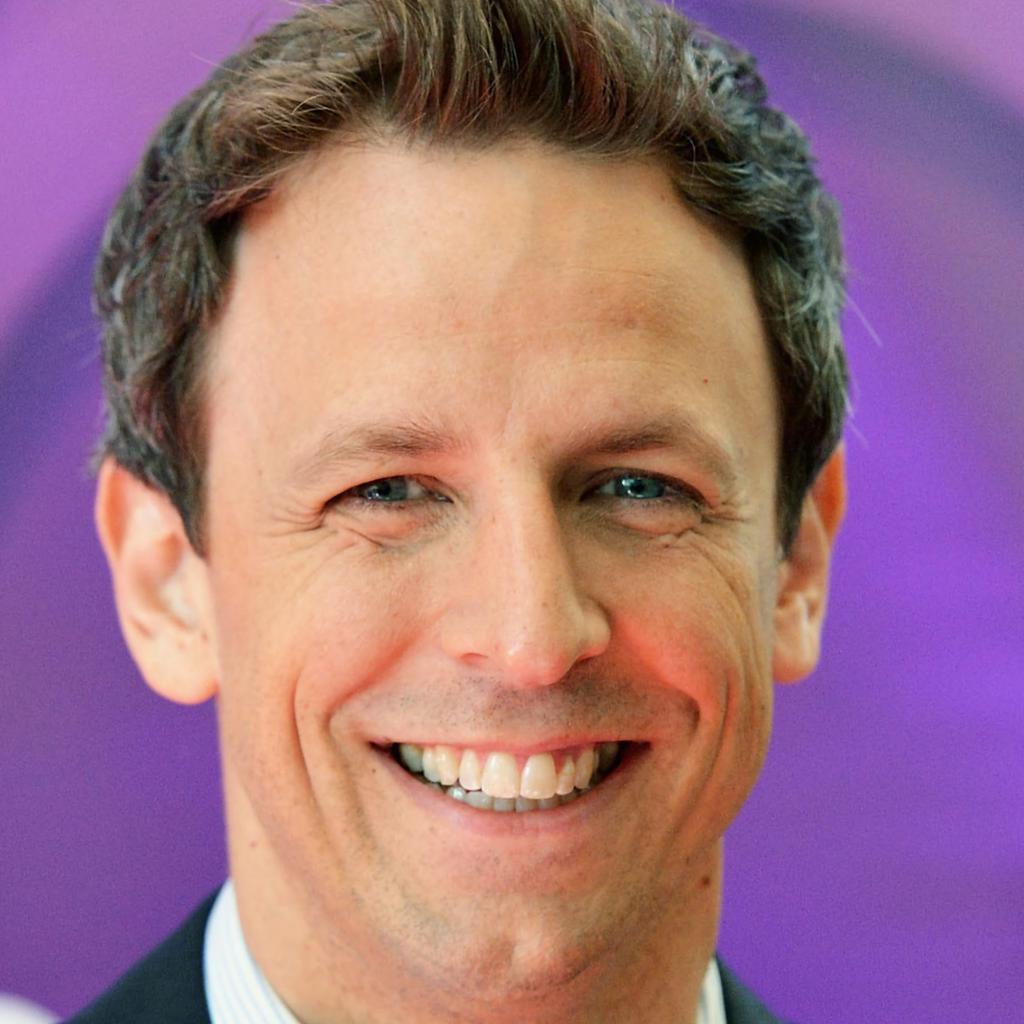}
			&\includegraphics[width=1.5cm]{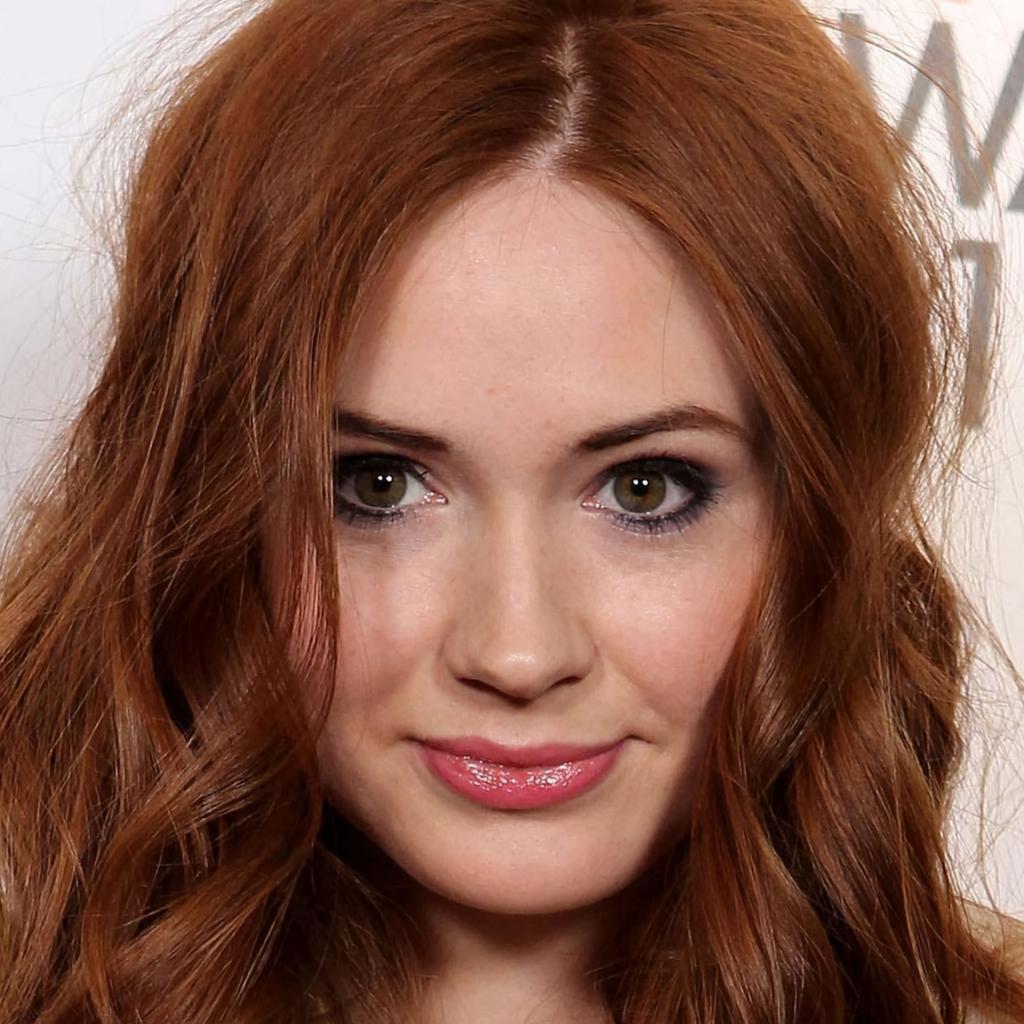}
			&\includegraphics[width=1.5cm]{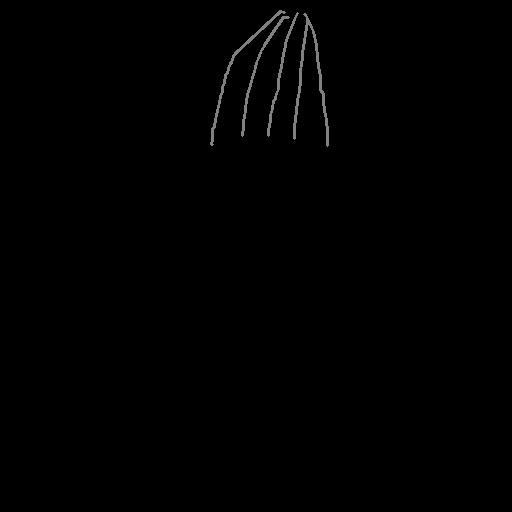}
			&\includegraphics[width=1.5cm]{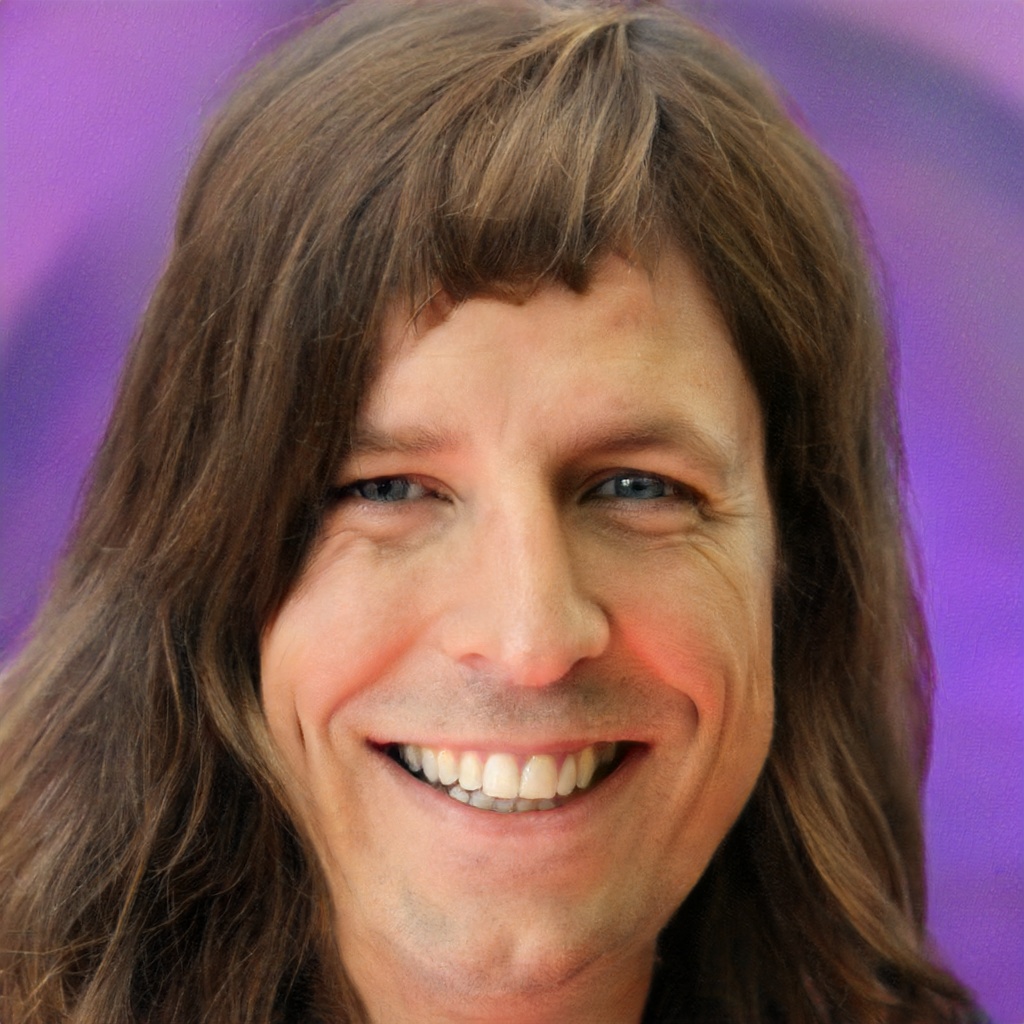}
			&\includegraphics[width=1.5cm]{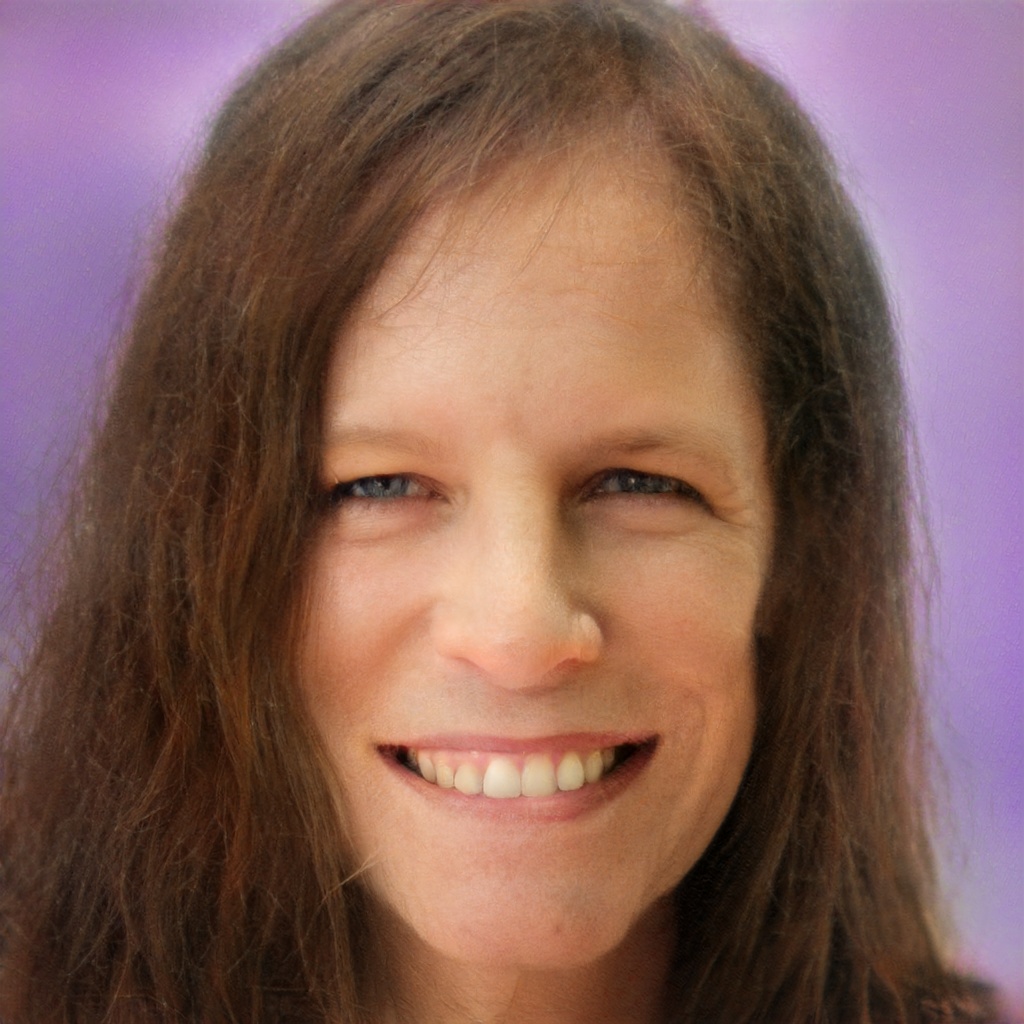}	
			\\
			
			\includegraphics[width=1.5cm]{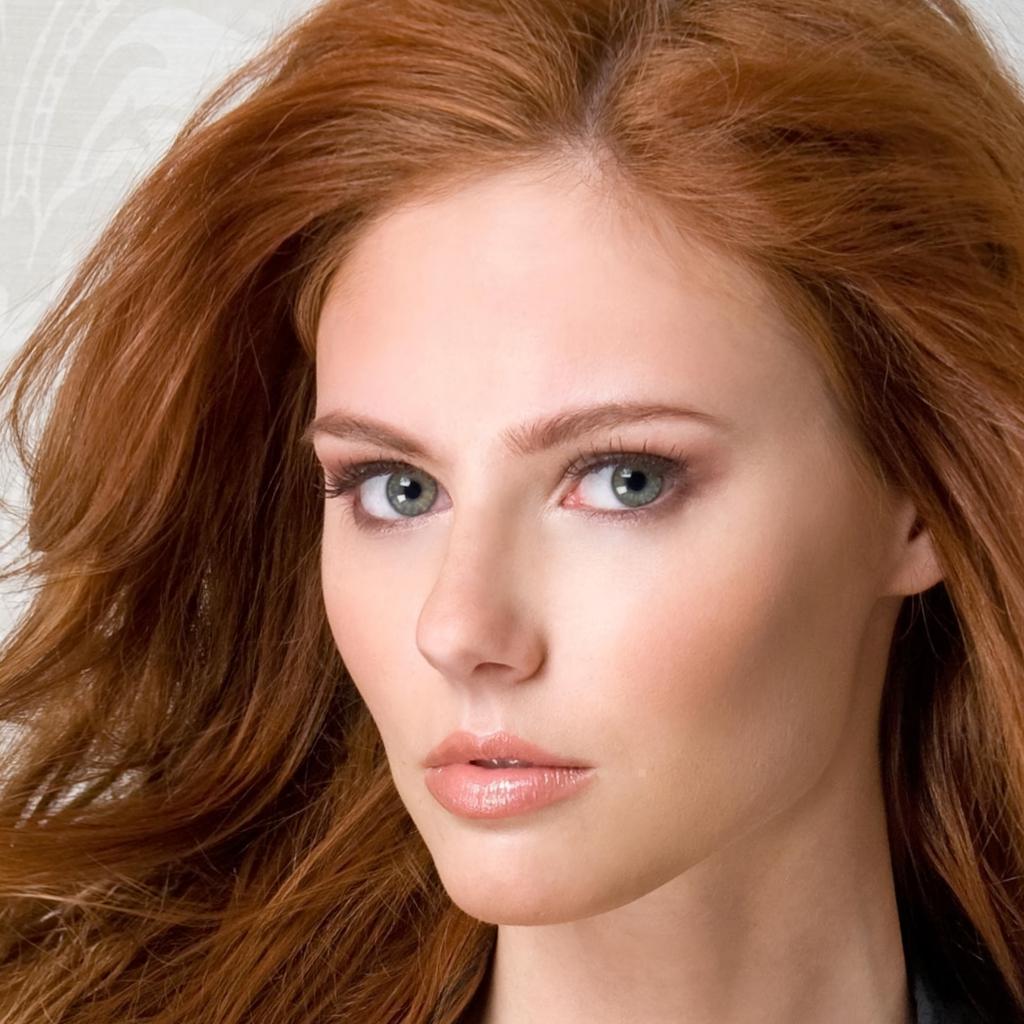}
			&\includegraphics[width=1.5cm]{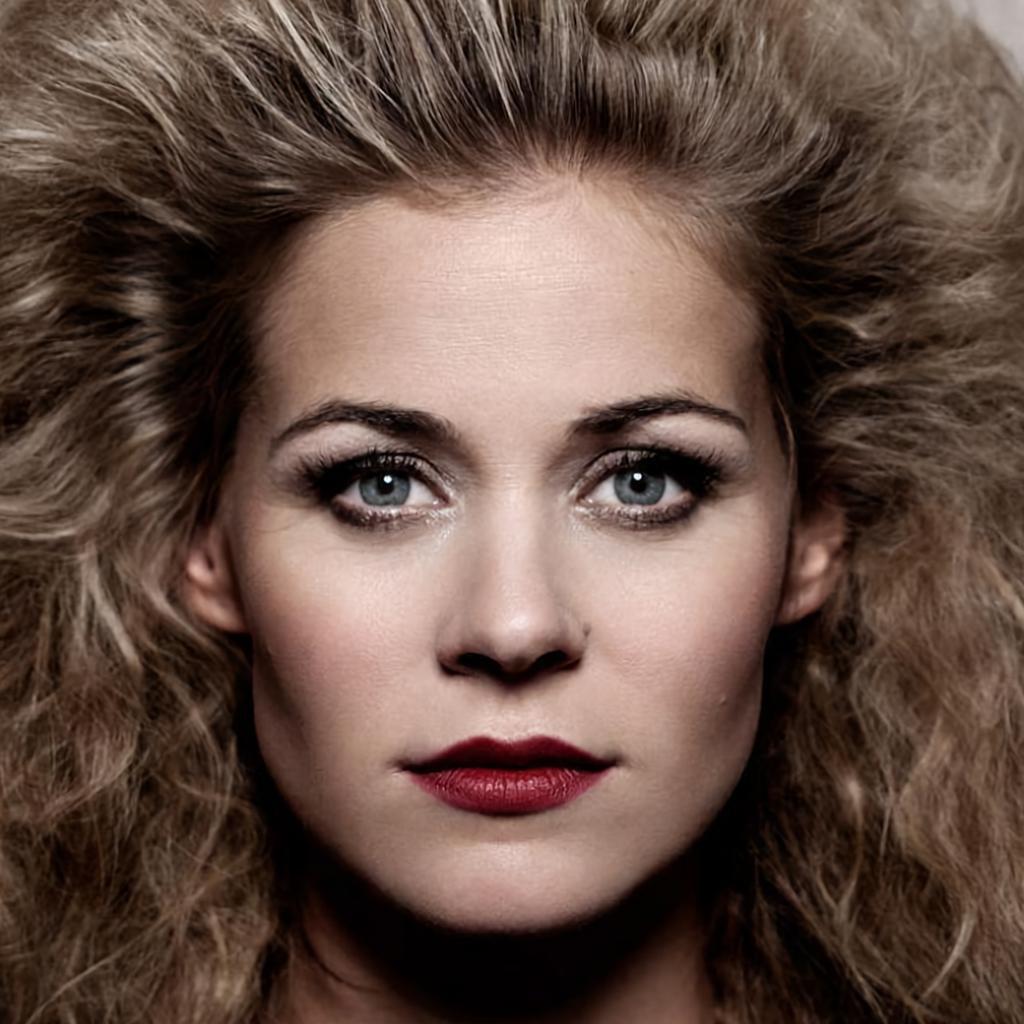}
			&\includegraphics[width=1.5cm]{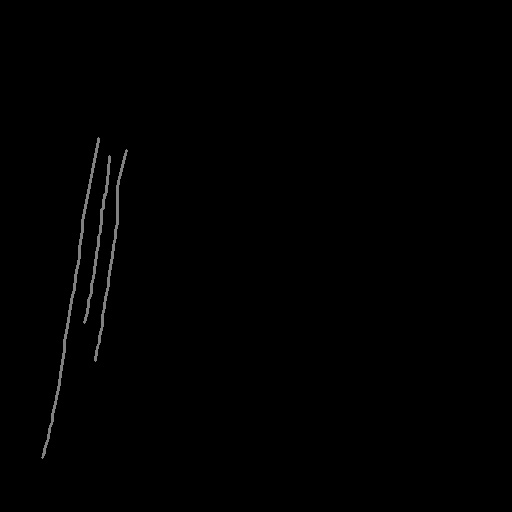}
			&\includegraphics[width=1.5cm]{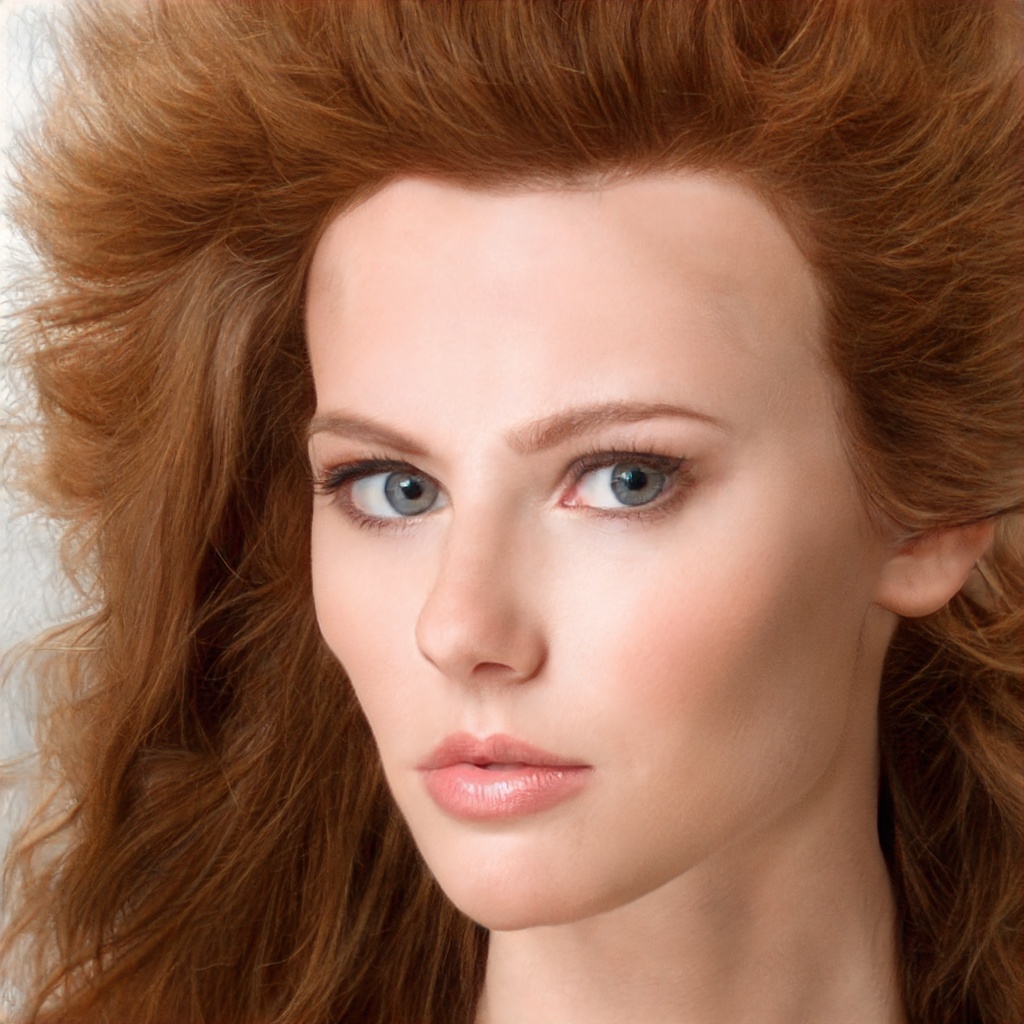}
			&\includegraphics[width=1.5cm]{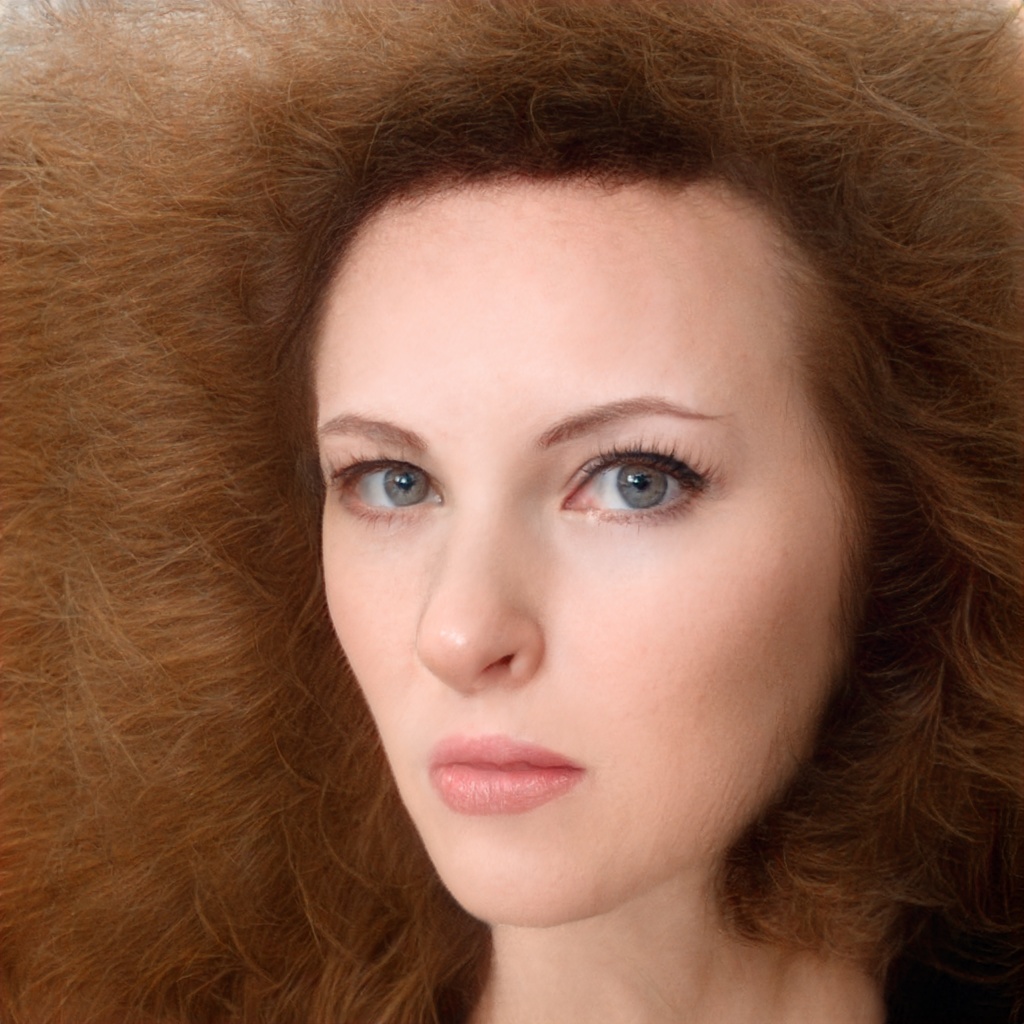}	
			\\
			
		\end{tabular}
	\end{center}
	\caption{Ablation of feature blending vs. latent code blending. Feature blending can enable global editing, local editing, and irrelevant attributes preservation simultaneously.} 
	\label{fig:blendablationfig}
\end{figure}

\noindent\textbf{Feature Blending vs. Latent Code Blending.} To demonstrate the superiority of proxy feature blending, we compare it with the alternative scheme based on the linear combination of latent codes. In detail, we initialize an interpolation factor for the latent code of the global editing proxy, the local sketch proxy, and the input image, respectively. In the optimization process, we optimize these three interpolation factors so that the generated image corresponding to the interpolated latent code is similar to the proxies or input image within the corresponding region. In Figure \ref{fig:blendablationfig}, we use the hairstyle reference image as an example to generate the global proxy. It is obvious that our scheme accomplishes global editing and local editing while perfectly keeping the irrelevant properties unmodified, while latent code blending does not perform either one well.

\section{Conclusion and Discussion}
In this paper, we propose a unified hair editing system HairCLIPv2, which presents the first attempt that supports both simple text/reference image interaction and fine-grained local interactions. It innovatively converts all hair editing tasks into hair transfer tasks, with the corresponding editing conditions converted into transfer proxies. It can not only achieve high-quality hair editing results, but also well preserve the irrelevant attributes from being modified. 
In the future, we will study how to use feed-forward networks to generate all the proxies. Also, it is worthy to generalize the proposed framework to support generic natural images.

\section{Acknowledgement} 
This work was supported in part by the Natural Science Foundation of China under Grant  U20B2047, 62072421, 62002334, 62102386 and 62121002, the Fundamental Research Funds for the Central Universities under Grant WK5290000003, Key Research and Development program of Anhui Province under Grant 2022k07020008. This work was also partly supported by Shenzhen Key Laboratory of Media Security, and the Opening Fund of Key Laboratory of Cyberculture Content Cognition and Detection, Ministry of Culture and Tourism. This work was also partially supported by a GRF grant (Project No. CityU 11216122) from the Research Grants Council (RGC) of Hong Kong. Thank Yi Yin for her help in this work.

{\small
	\bibliographystyle{ieee_fullname}
	\bibliography{egbib}
}
\clearpage
\appendix
\section{Implementation Details} 
For the generation of all optimization-based proxies, we set the learning rate to $ 0.01 $ and use the Adam~\cite{kingma2015adam} optimizer. For text proxy, the overall loss in the optimization process is defined as follows: $ {L}^{text}=\lambda^{clip}{L}^{clip} + \lambda^{pose}{L}^{pose}+ \lambda^{shape}{L}^{shape}, $ where $ \lambda^{clip} $, $ \lambda^{pose} $, and $ \lambda^{shape} $ are set to $ 1 $, $ 200 $, and $ 1 $ respectively to make each loss balance. For the start point strategy for optimization of text proxy, we set $ \psi=0.3 $ to ensure that the initial optimization starting point $ w^{init} $ is around the average face latent code $ w^{mean} $. For reference proxy, The overall loss of hairstyle transfer is defined as follows: $ {L}^{ref}=\lambda^{style}{L}^{style} + \lambda^{pose}{L}^{pose}+ \lambda^{reg}{L}^{reg}+ \lambda^{shape}{L}^{shape}, $ where $ \lambda^{style} $, $ \lambda^{pose} $, $ \lambda^{reg} $, and $ \lambda^{shape} $ are set to $ 2000 $, $ 200 $, $ 1 $, and $ 1 $ respectively to make each loss balanced. 

For sketch proxy, the number of training iterations for the sketch2hair translation inverter $ T $ for local hairstyle editing is $ 500,000 $. The training loss includes regular pixel-level $ L_{2} $ loss $ {L}^{mse}=||I^{sketch}-G(T(S))||^{2}_{2} $, feature-level LPIPS~\cite{zhang2018unreasonable} loss $ {L}^{LPIPS}=||F(I^{sketch})-F(G(T(S)))||^{2}_{2} $, where $ S $ represents the hairstyle sketch input, $ I^{sketch} $ stands for the hair image corresponding to the hairstyle sketch $ S $, $ T $ means the sketch2hair invertor to be trained, and $ F $ denotes the AlexNet~\cite{krizhevsky2017imagenet} feature extractor. To provide more local supervision, we additionally use multi-layer face parsing loss $ {L}^{m\_par} $ which provides more detailed knowledge by introducing multi-layer features from the pre-trained face parsing network:
\begin{equation}
{L}^{m\_par}=\sum^{5}_{i=1}(1-cos(P_{i}(I^{sketch}), P_{i}(G(T(S))))),
\end{equation}
where $ P_{i}(I^{sketch}) $ represents the feature corresponding to the $ i $-th semantic level from the face parsing network $ P $~\cite{FaceParsing} of the hair image $ I^{sketch} $. The overall training losses are as follows:
\begin{equation}
	{L}^{sketch}=\lambda^{mse}{L}^{mse} + \lambda^{LPIPS}{L}^{LPIPS}+ \lambda^{m\_par}{L}^{m\_par},
\end{equation}
where $ \lambda^{mse} $, $ \lambda^{LPIPS} $, and $ \lambda^{m\_par} $ are set to $ 0.5 $, $ 0.8 $, and $ 1 $, respectively.

\section{Quantitative Results}



\subsection{Editing Speed} We compare the editing runtime with competitive methods in Table \ref{ref:speed}. We are faster than baseline methods in hair transfer and sketch-based editing. For text-based editing, we are slower but with better editing quality and irrelevant attributes preservation. Moreover, only our method excels at the task of hair editing with arbitrary text.

\begin{table}[h]
	\setlength{\tabcolsep}{3pt}
	\small
	\centering
	\begin{tabular}{l|ccc}
		\hline
		Text  & Ours(35.2) & TediGAN(28.0) & HairCLIP(\textbf{0.10})\\
		Transfer & Ours(\textbf{58.9}) & Barbershop(117.8) & SYH(136.8) \\
		Sketch & Ours(\textbf{0.04}) & MichiGAN(0.42) & SketchSalon(0.14)\\
		\hline
		
	\end{tabular}
    \vspace{1em}
	\caption{Editing Runtime on 2080 Ti (seconds).}
	\label{ref:speed}
\end{table}

\section{Ablation Analysis}
\subsection{Necessity of Balding Steps}

\begin{figure}[h]
	\begin{center}
		\setlength{\tabcolsep}{0.5pt}
		\begin{tabular}{m{2cm}<{\centering}m{2cm}<{\centering}m{2cm}<{\centering}m{2cm}<{\centering}}
			\scriptsize{Input Image} & \scriptsize{w/o Balding (A)} & \scriptsize{w/o Blending (B)} & \scriptsize{Ours}
			\\
			
			\includegraphics[width=1.9cm]{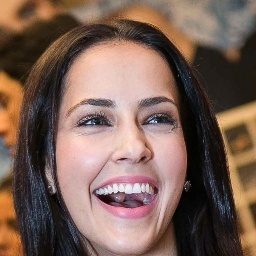}
			&\includegraphics[width=1.9cm]{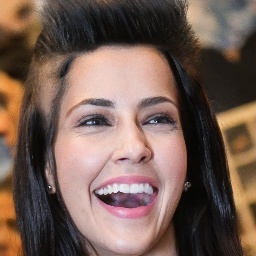}
			&\includegraphics[width=1.9cm]{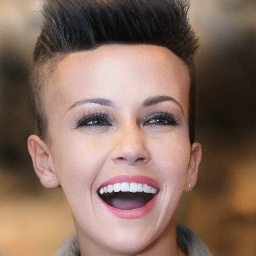}
			&\includegraphics[width=1.9cm]{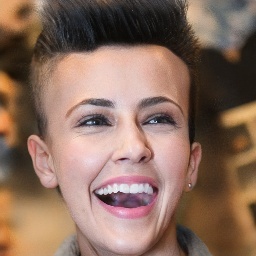}	
			\\
			
			\includegraphics[width=1.9cm]{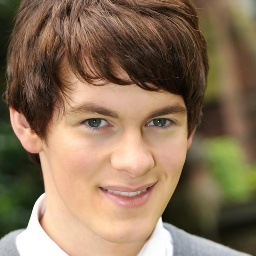}
			&\includegraphics[width=1.9cm]{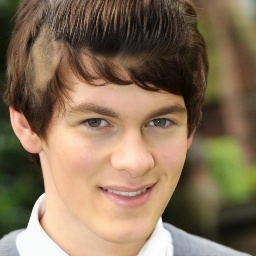}
			&\includegraphics[width=1.9cm]{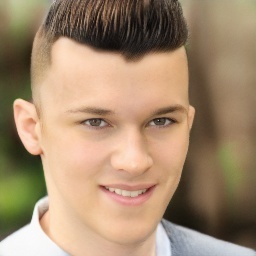}
			&\includegraphics[width=1.9cm]{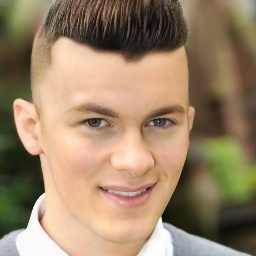}	
			\\			
			
		\end{tabular}
	\end{center}
	\caption{Ablation analysis on the necessity of balding steps. The text description is ``\textit{Mohawk Hairstyle}''.}
	\label{fig:bald_ablation}
\end{figure}

We employ two key steps during the process of converting the input image into bald proxy: first, editing the latent code of the input image to obtain its balded latent code; second, performing feature blending between the balded features and the original features of the input image to preserve the irrelevant attributes from being modified as shown in Eq. 1 of the main text. To verify the necessity of these two steps, we perform experiments on the following two variants: (A). without balding, i.e., step $ 1 $ is skipped and Eq. 1 of the main text becomes $ F^{bald}_7=F^{src}_7$; (B). without feature blending with original image after balding, i.e., Eq. 1 of the main text becomes $ F^{bald}_7=G(w^{bald}_{1-7}) $. The visual comparison results are shown in Figure \ref{fig:bald_ablation}. Since variant A does not employ the balding method to de-obscure, there are obvious artificial artifacts caused by blending the bald proxy features with the editing proxy features. Although the result of variant B looks relatively natural overall, the editing of the 1-dimensional latent code inevitably modifies other irrelevant attributes (background, identity, etc.). Combining the advantages of these two steps, our default setting achieves both the natural editing effect resulting from the balding operation to de-obscure while inpainting the hair area sensibly and the excellent irrelevant attribute preservation caused by feature blending.

\subsection{Robustness of Balding Technique}

\begin{figure}[tb]
	\begin{center}
		\setlength{\tabcolsep}{0.5pt}
		\begin{tabular}{m{0.3cm}<{\centering}m{1.3cm}<{\centering}m{1.3cm}<{\centering}m{1.3cm}<{\centering}m{1.3cm}<{\centering}m{1.3cm}<{\centering}m{1.3cm}<{\centering}}
			& \scriptsize{Input Image} & \scriptsize{Bald} & \scriptsize{Result} & \scriptsize{Input Image} & \scriptsize{Bald} & \scriptsize{Result}
			\\
			
			\raisebox{0.4cm}{\rotatebox[origin=c]{90}{\footnotesize{{lighting}}}}
			&\includegraphics[width=1.25cm]{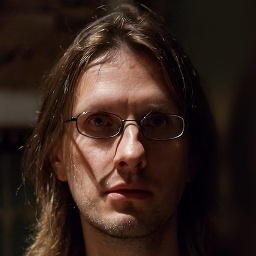}
			&\includegraphics[width=1.25cm]{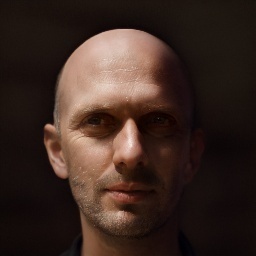}
			&\includegraphics[width=1.25cm]{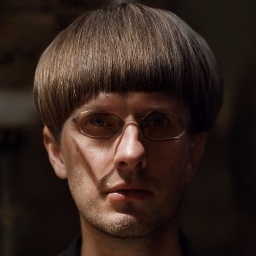}
			&\includegraphics[width=1.25cm]{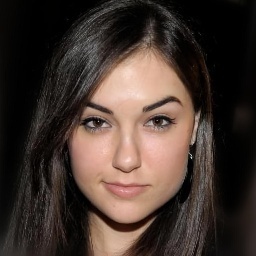}
			&\includegraphics[width=1.25cm]{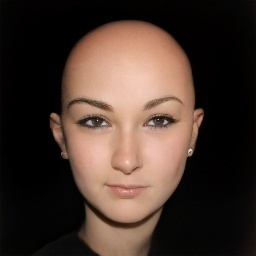}
			&\includegraphics[width=1.25cm]{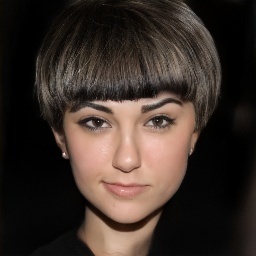}	
			\\

			\raisebox{0.3cm}{\rotatebox[origin=c]{90}{\footnotesize{{pose}}}}
			&\includegraphics[width=1.25cm]{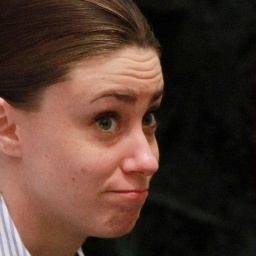}
			&\includegraphics[width=1.25cm]{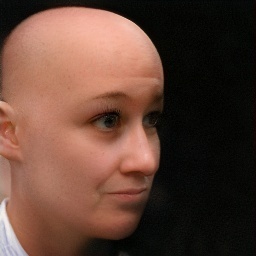}
			&\includegraphics[width=1.25cm]{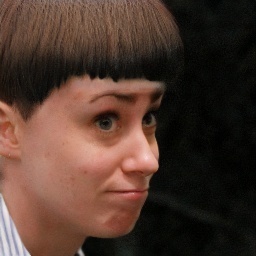}
			&\includegraphics[width=1.25cm]{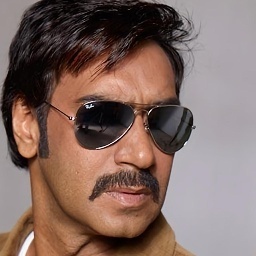}
			&\includegraphics[width=1.25cm]{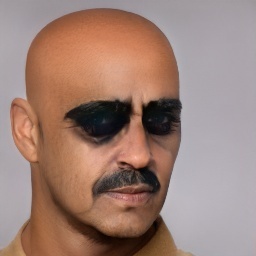}
			&\includegraphics[width=1.25cm]{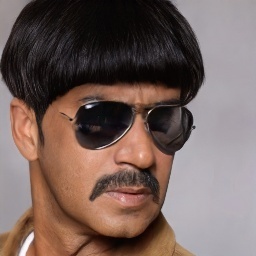}	
			\\
			
			\raisebox{0.43cm}{\rotatebox[origin=c]{90}{\footnotesize{{occlusion}}}}
			&\includegraphics[width=1.25cm]{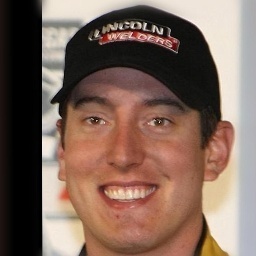}
			&\includegraphics[width=1.25cm]{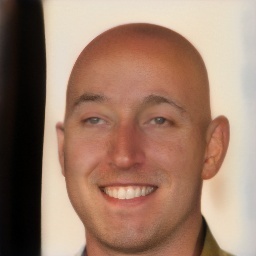}
			&\includegraphics[width=1.25cm]{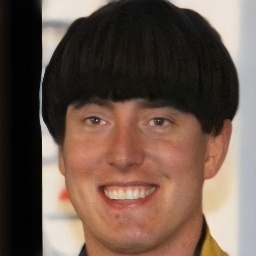}
			&\includegraphics[width=1.25cm]{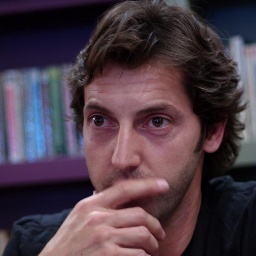}
			&\includegraphics[width=1.25cm]{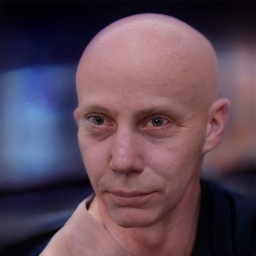}
			&\includegraphics[width=1.25cm]{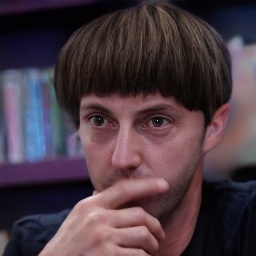}	
			\\

		\end{tabular}
	\end{center}
	\caption{Ablation analysis on the robustness of balding technique. The text description is ``\textit{Bowl Cut Hairstyle}''.}
	\label{fig:bald_env}
\end{figure}

Our system uses HairMapper~\cite{wu2022hairmapper} in the first step of generating bald proxy to baldify the input image and thus remove the occlusion and facilitate feature blending with the editing proxy later. In Figure \ref{fig:bald_env}, we illustrate the results of the balding technique~\cite{wu2022hairmapper} and our method under extreme lighting, pose, and self-occlusion conditions. Obviously, the balding technique performs relatively robustly in most extreme conditions. In the case of the self-occlusion condition, the balding technique shows significant artifacts at the hand position, while our method is not affected because of the feature blending mechanism adopted in the second step of generating the bald proxy.

\section{Limitations}
\begin{figure}[h]
	\begin{center}
		\setlength{\tabcolsep}{0.5pt}
		\begin{tabular}{m{1.3cm}<{\centering}m{1.3cm}<{\centering}m{1.3cm}<{\centering}m{1.3cm}<{\centering}m{1.3cm}<{\centering}m{1.3cm}<{\centering}}
			\scriptsize{Input Image} & \scriptsize{Color Ref} & \scriptsize{Result} & \scriptsize{Input Image} & \scriptsize{Color Ref} & \scriptsize{Result}
			\\
			
			\includegraphics[width=1.25cm]{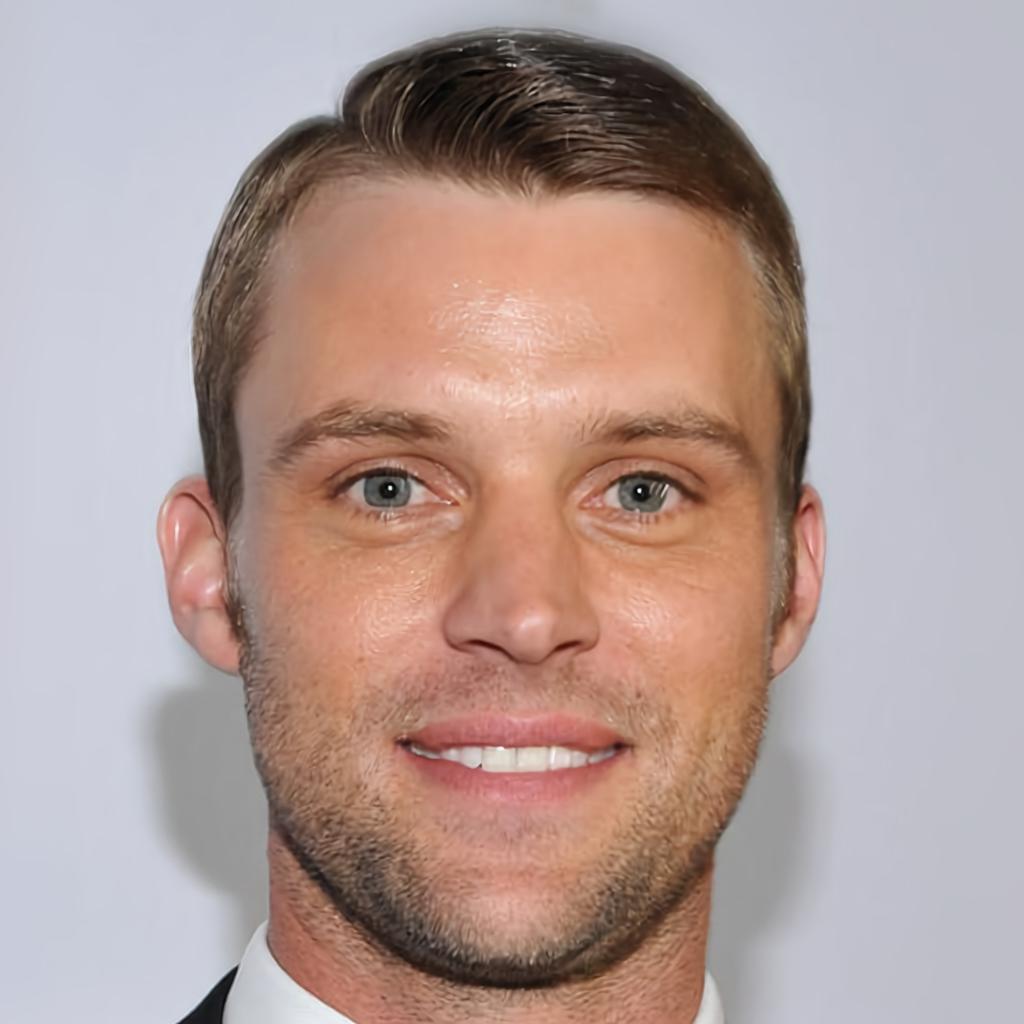}
			&\includegraphics[width=1.25cm]{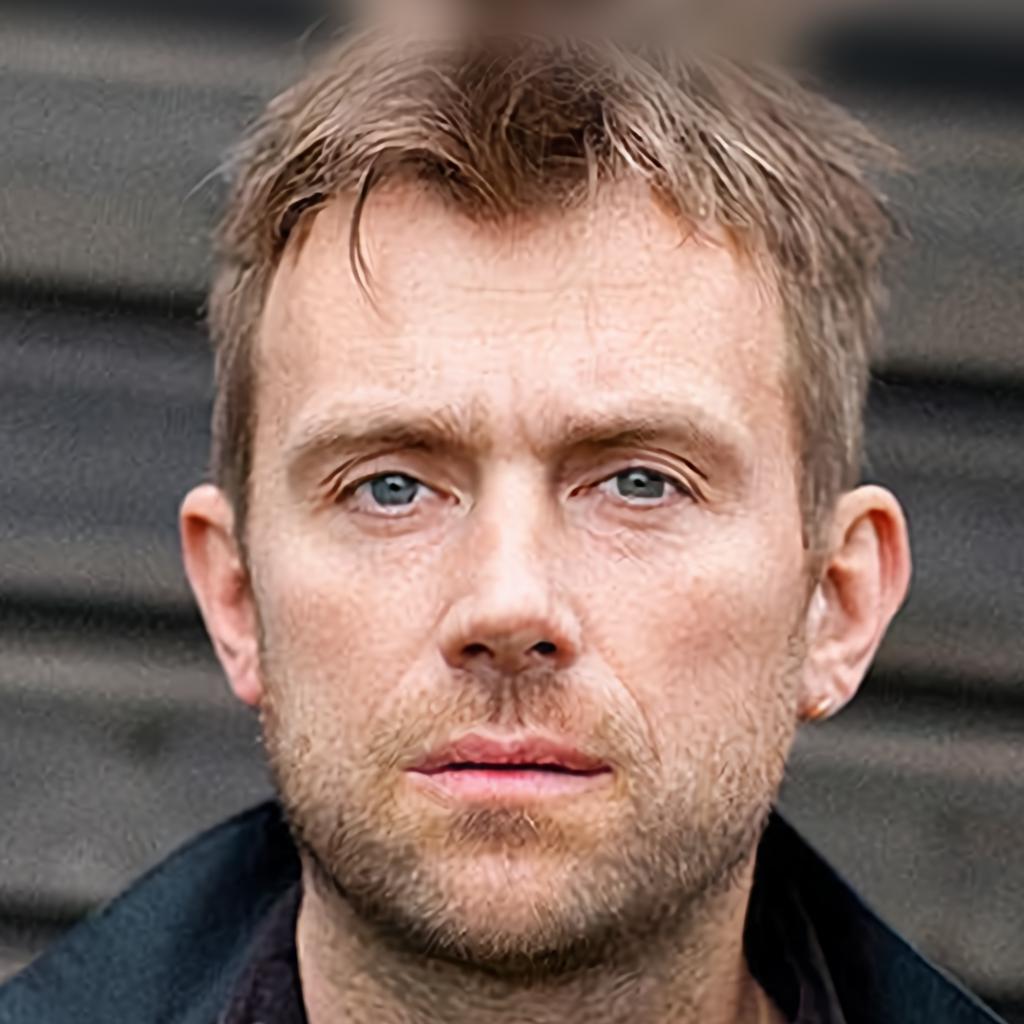}
			&\includegraphics[width=1.25cm]{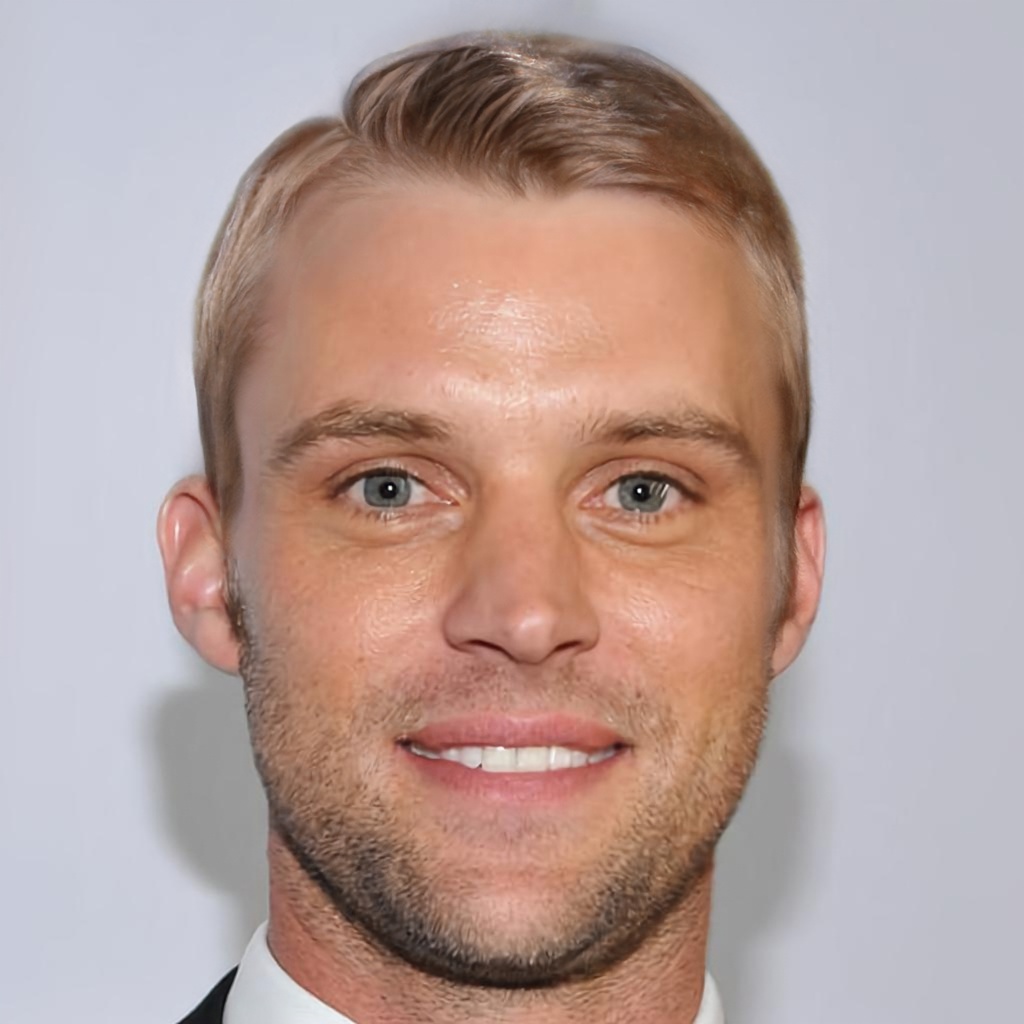}
			&\includegraphics[width=1.25cm]{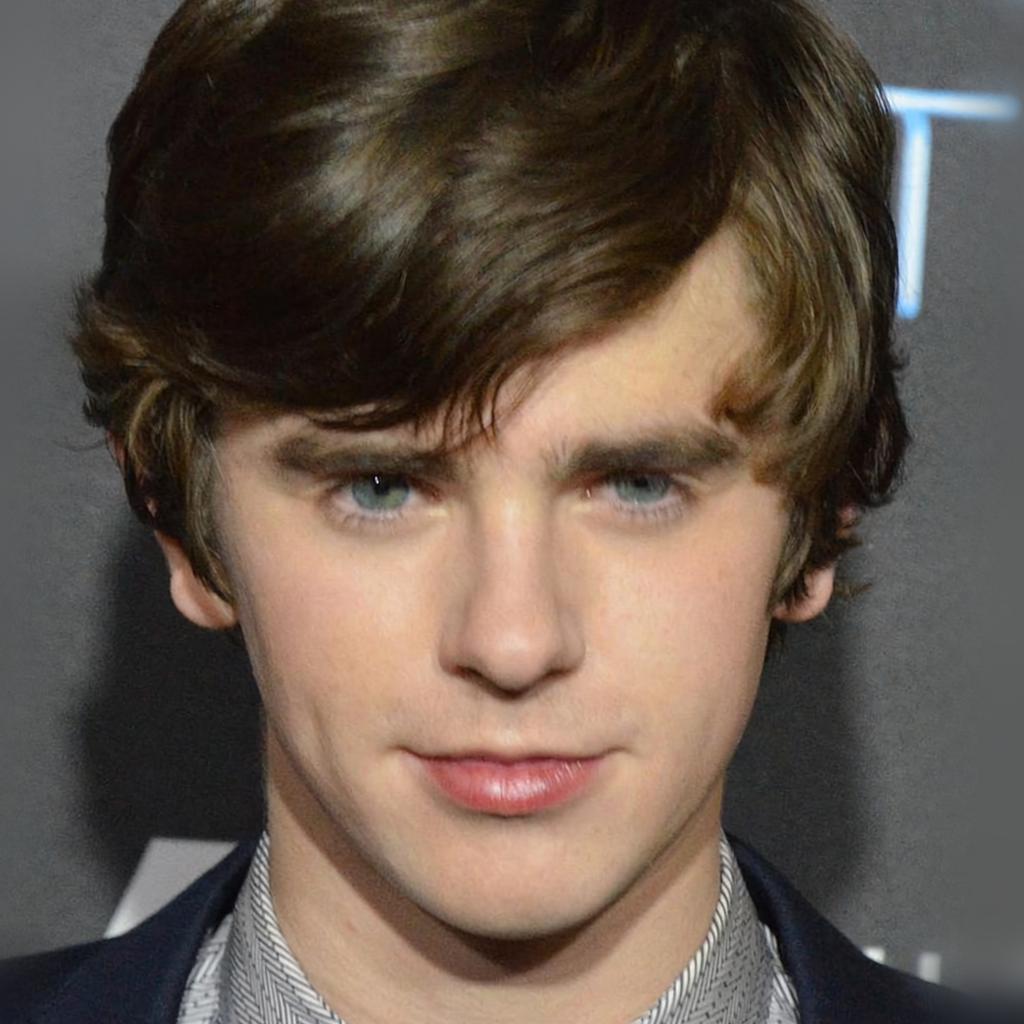}
			&\includegraphics[width=1.25cm]{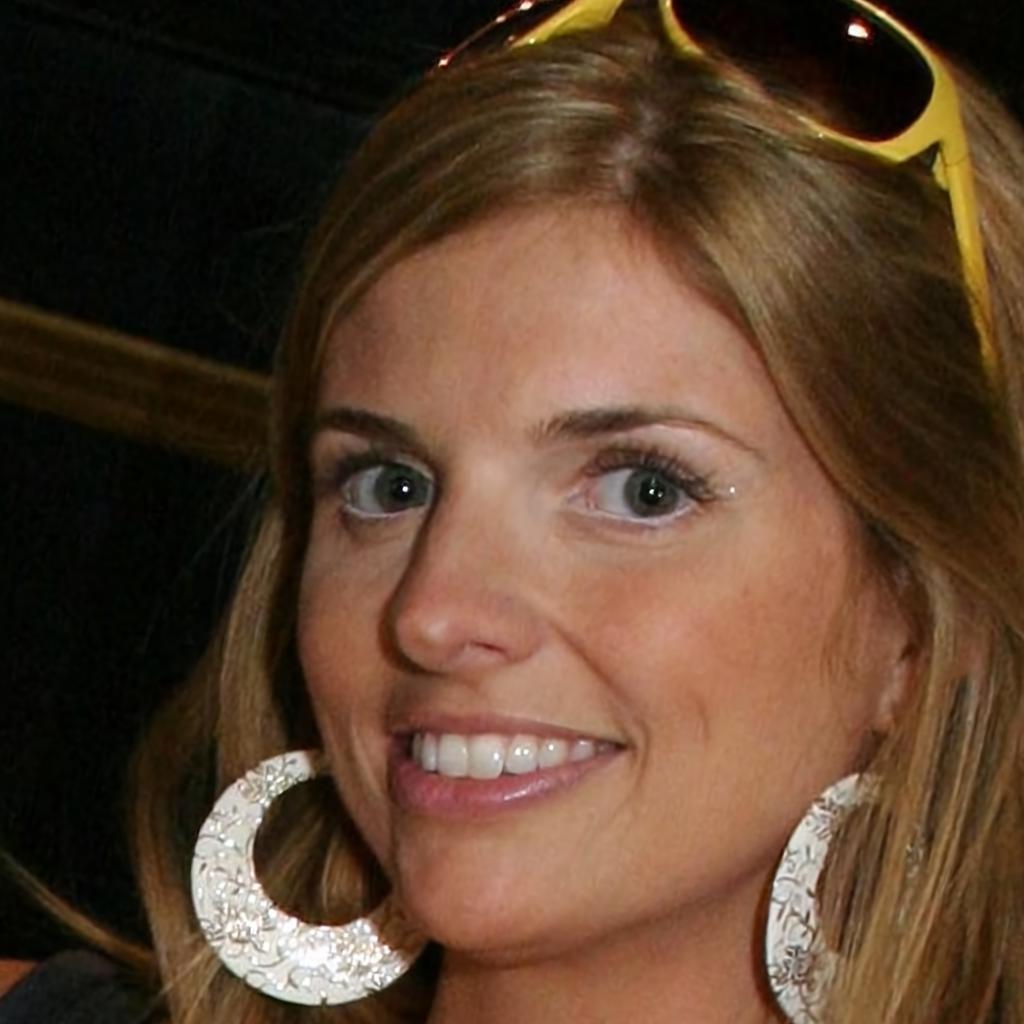}
			&\includegraphics[width=1.25cm]{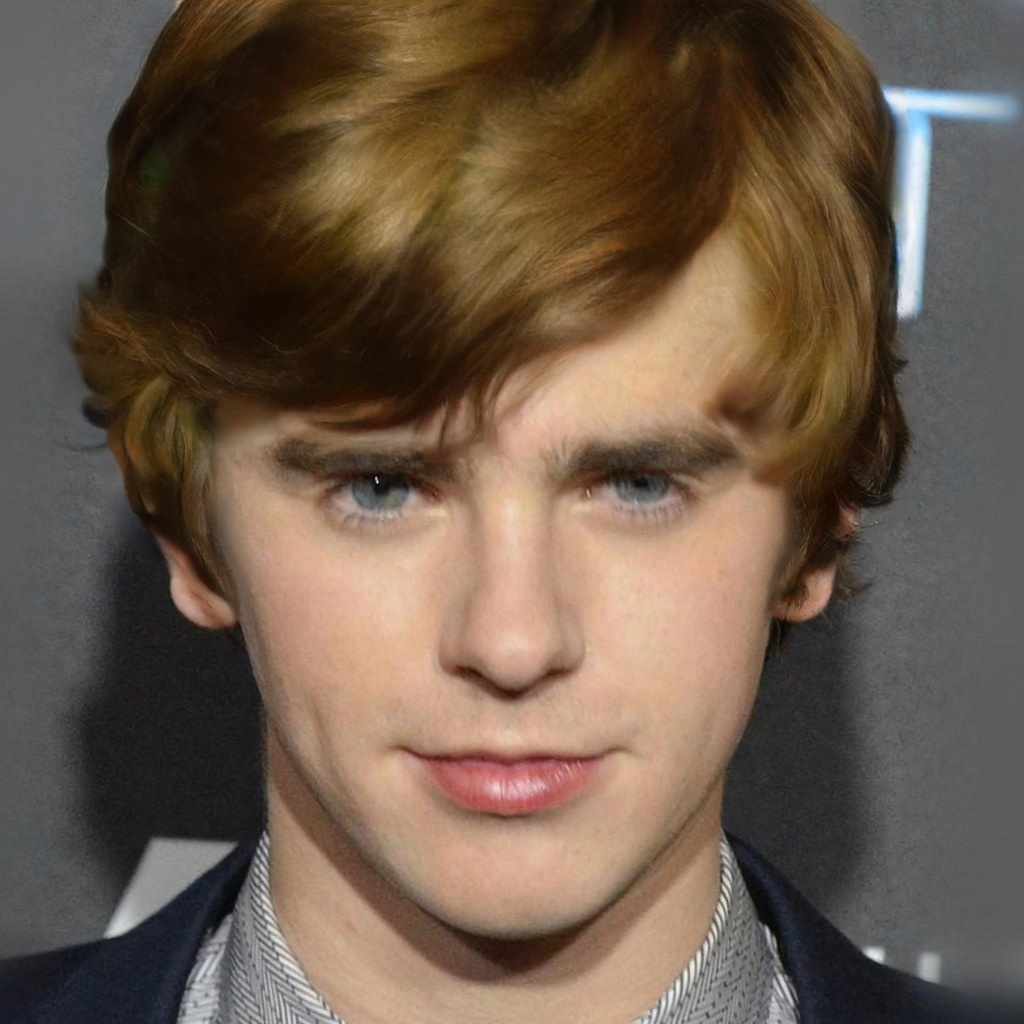}	
			\\
			
		\end{tabular}
	\end{center}
	\caption{Failure cases.}
	\label{fig:failcase}
\end{figure}

Despite the unprecedented unification, our method has some limitations. For example, our method only focuses on image hair editing, and cannot handle facial hair or coherent video hair editing. Moreover, our method cannot perfectly transfer the reference color for some cases (e.g., slight color bias in Fig. \ref{fig:failcase}), especially when the lighting is very different. Lastly, for some conditions our method still gets the proxy by optimization, thus real-time generation of all proxies is the future research direction.

\section{More Qualitative Results}
\begin{figure*}[tb]
	\begin{center}
		\setlength{\tabcolsep}{0.5pt}
		\begin{tabular}{m{0.3cm}<{\centering}m{2.15cm}<{\centering}m{2.15cm}<{\centering}m{2.15cm}<{\centering}m{2.15cm}<{\centering}m{2.15cm}<{\centering}m{2.15cm}<{\centering}m{2.15cm}<{\centering}}
			& \small{Input Image} & \small{Example} & \small{Ours} & \small{HairCLIP} & \small{StyleCLIP} & \small{TediGAN} & \small{DiffCLIP}
			\\
			
			\raisebox{0.15cm}{\rotatebox[origin=c]{90}{\footnotesize{{afro}}}}
			&\includegraphics[width=1.9cm]{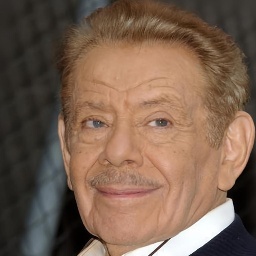}
			&\includegraphics[width=1.9cm]{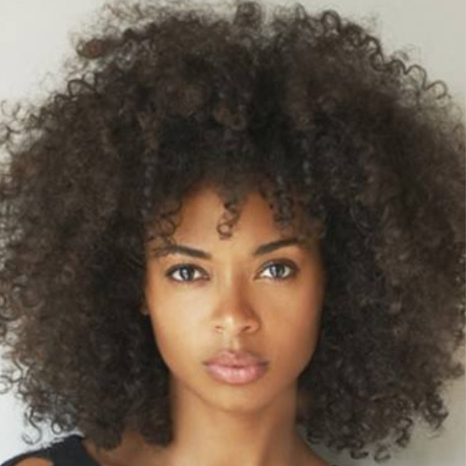}
			&\includegraphics[width=1.9cm]{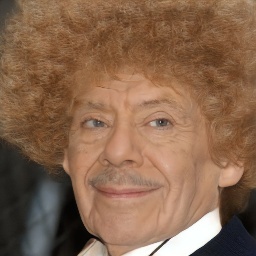}
			&\includegraphics[width=1.9cm]{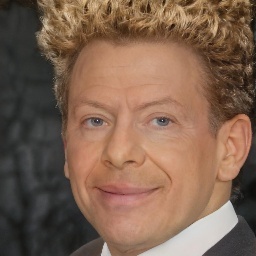}
			&\includegraphics[width=1.9cm]{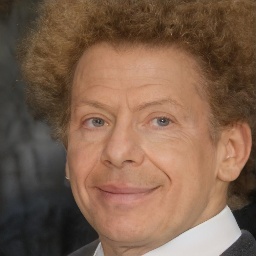}
			&\includegraphics[width=1.9cm]{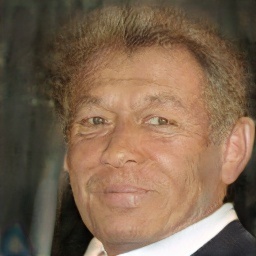}
			&\includegraphics[width=1.9cm]{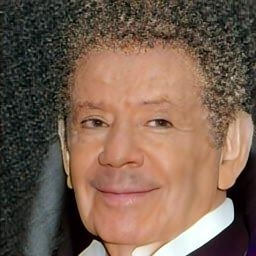}
			\\
			
			\raisebox{0.3cm}{\rotatebox[origin=c]{90}{\footnotesize{{bob cut}}}}
			&\includegraphics[width=1.9cm]{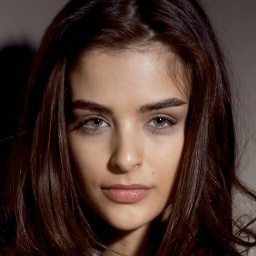}
			&\includegraphics[width=1.9cm]{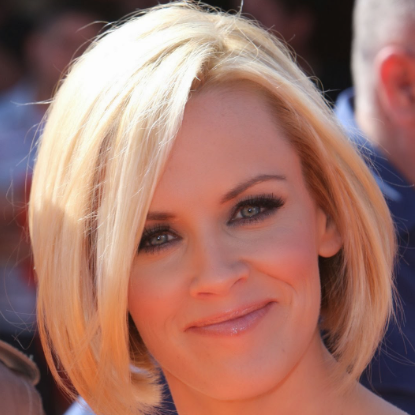}
			&\includegraphics[width=1.9cm]{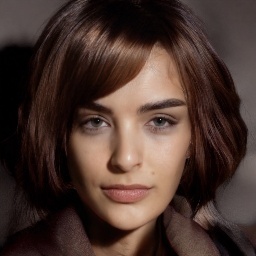}
			&\includegraphics[width=1.9cm]{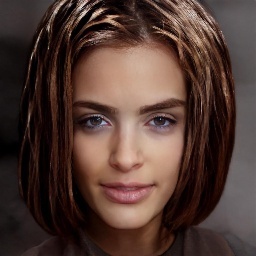}
			&\includegraphics[width=1.9cm]{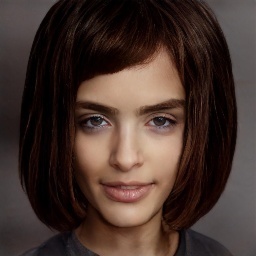}
			&\includegraphics[width=1.9cm]{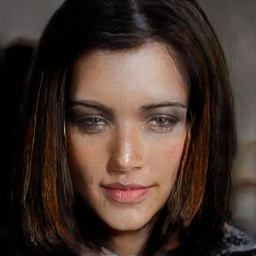}
			&\includegraphics[width=1.9cm]{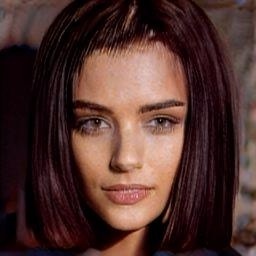}
			\\
			
			\raisebox{0.3cm}{\rotatebox[origin=c]{90}{\footnotesize{{bowl cut}}}}
			&\includegraphics[width=1.9cm]{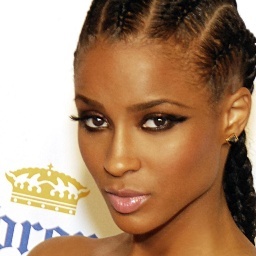}
			&\includegraphics[width=1.9cm]{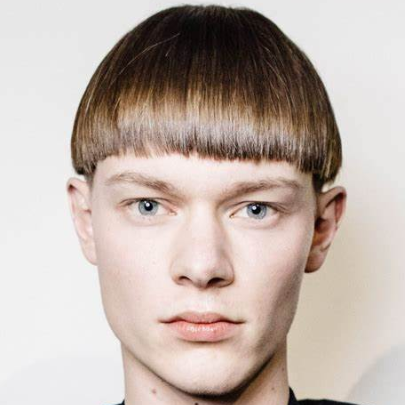}
			&\includegraphics[width=1.9cm]{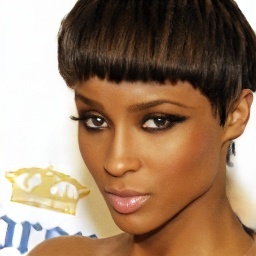}
			&\includegraphics[width=1.9cm]{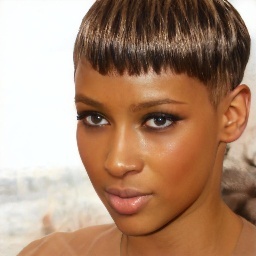}
			&\includegraphics[width=1.9cm]{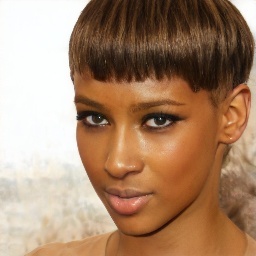}
			&\includegraphics[width=1.9cm]{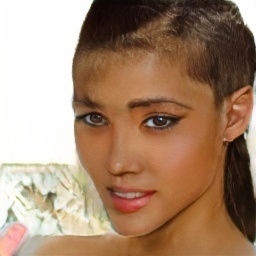}
			&\includegraphics[width=1.9cm]{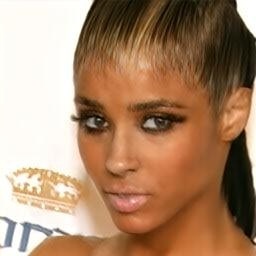}
			\\
			
			\raisebox{0.35cm}{\rotatebox[origin=c]{90}{\footnotesize{{mohawk}}}}
			&\includegraphics[width=1.9cm]{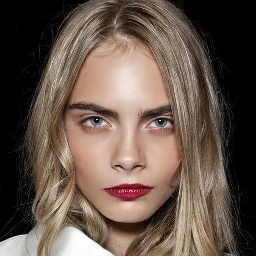}
			&\includegraphics[width=1.9cm]{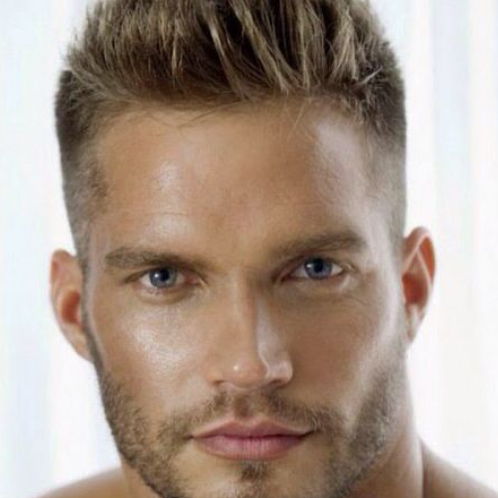}
			&\includegraphics[width=1.9cm]{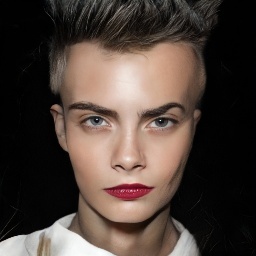}
			&\includegraphics[width=1.9cm]{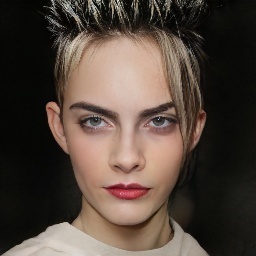}
			&\includegraphics[width=1.9cm]{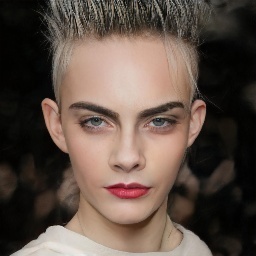}
			&\includegraphics[width=1.9cm]{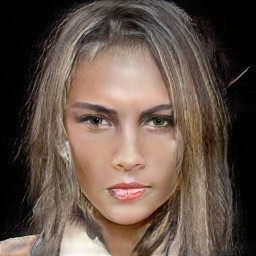}
			&\includegraphics[width=1.9cm]{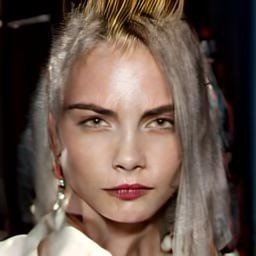}
			\\
			
			\raisebox{0.3cm}{\rotatebox[origin=c]{90}{\footnotesize{{purple}}}}
			&\includegraphics[width=1.9cm]{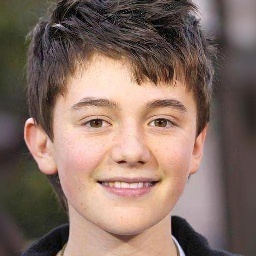}
			&\includegraphics[width=1.9cm]{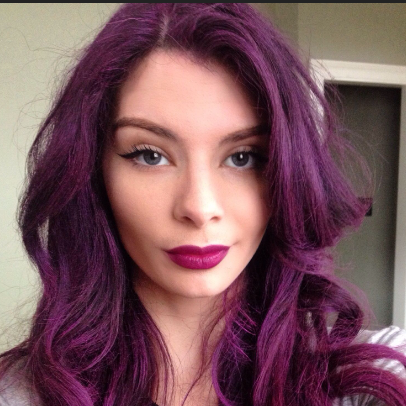}
			&\includegraphics[width=1.9cm]{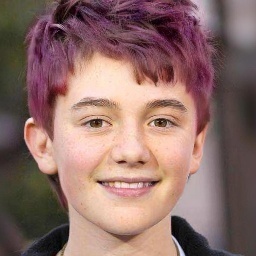}
			&\includegraphics[width=1.9cm]{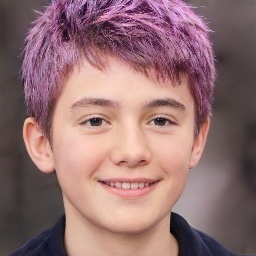}
			&\includegraphics[width=1.9cm]{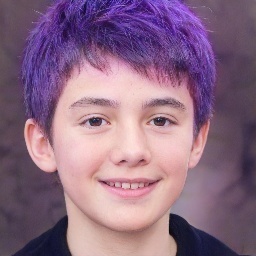}
			&\includegraphics[width=1.9cm]{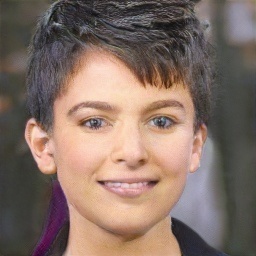}
			&\includegraphics[width=1.9cm]{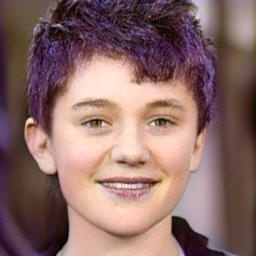}
			\\

			\raisebox{0.3cm}{\rotatebox[origin=c]{90}{\footnotesize{{green}}}}
			&\includegraphics[width=1.9cm]{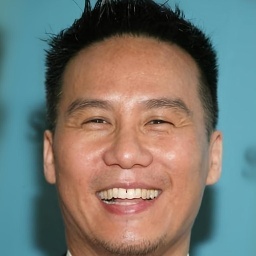}
			&\includegraphics[width=1.9cm]{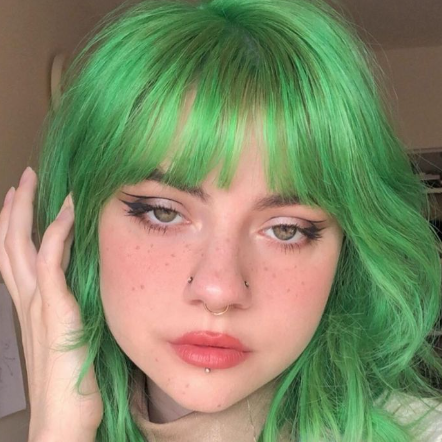}
			&\includegraphics[width=1.9cm]{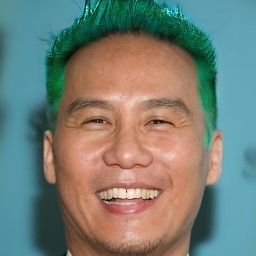}
			&\includegraphics[width=1.9cm]{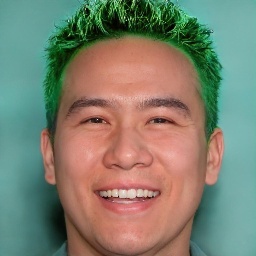}
			&\includegraphics[width=1.9cm]{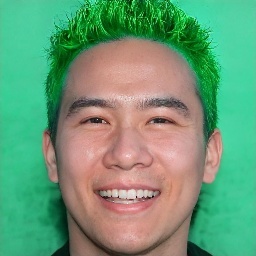}
			&\includegraphics[width=1.9cm]{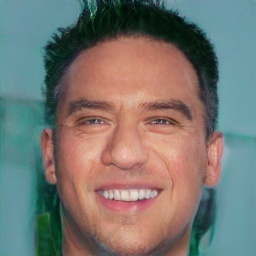}
			&\includegraphics[width=1.9cm]{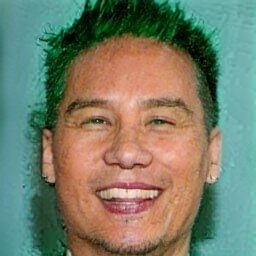}	
			\\
			
			\raisebox{0.2cm}{\rotatebox[origin=c]{90}{\footnotesize{{blond}}}}
			&\includegraphics[width=1.9cm]{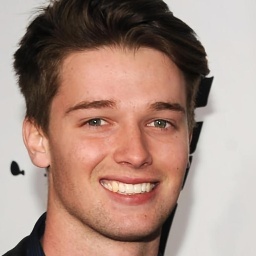}
			&\includegraphics[width=1.9cm]{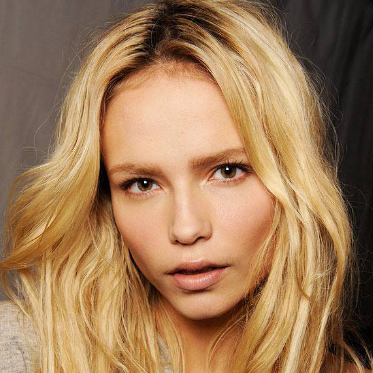}
			&\includegraphics[width=1.9cm]{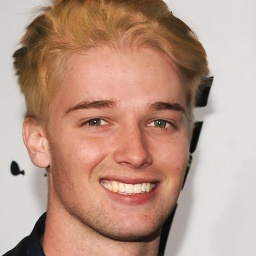}
			&\includegraphics[width=1.9cm]{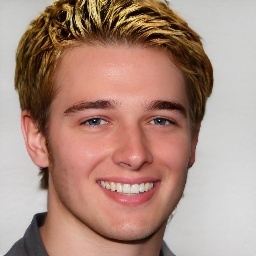}
			&\includegraphics[width=1.9cm]{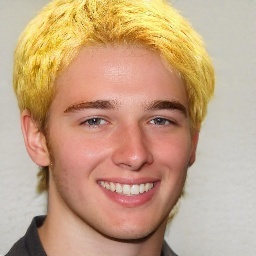}
			&\includegraphics[width=1.9cm]{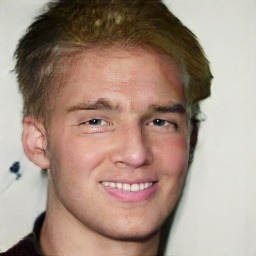}	
			&\includegraphics[width=1.9cm]{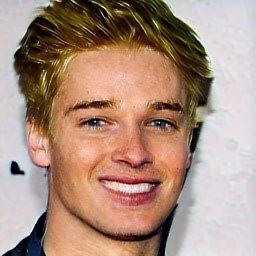}	
			\\
			
			\raisebox{0.55cm}{\rotatebox[origin=c]{90}{\footnotesize{{braid brown}}}}
			&\includegraphics[width=1.9cm]{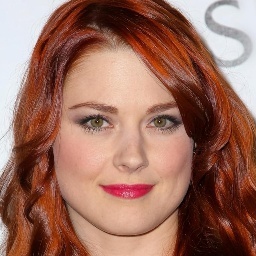}
			&\includegraphics[width=1.9cm]{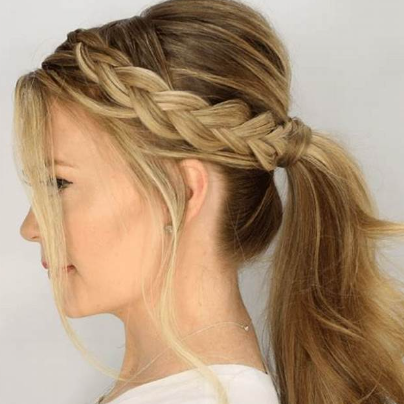}
			&\includegraphics[width=1.9cm]{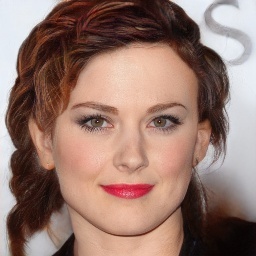}
			&\includegraphics[width=1.9cm]{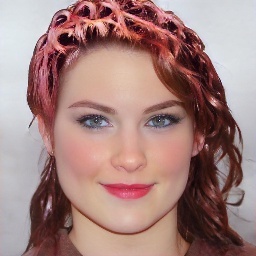}
			&\includegraphics[width=1.9cm]{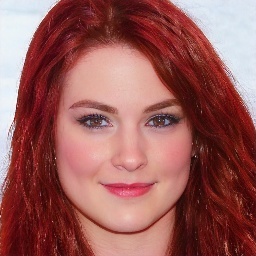}
			&\includegraphics[width=1.9cm]{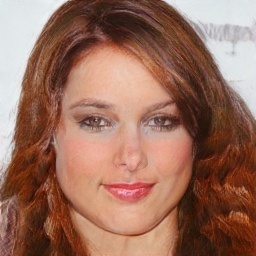}
			&\includegraphics[width=1.9cm]{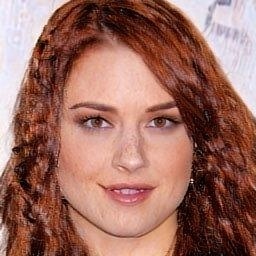}
			\\
			
			\raisebox{0.63cm}{\rotatebox[origin=c]{90}{\footnotesize{{crew yellow}}}}
			&\includegraphics[width=1.9cm]{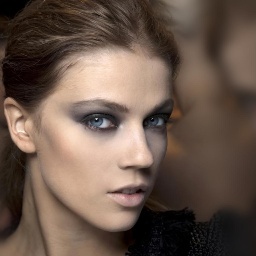}
			&\includegraphics[width=1.9cm]{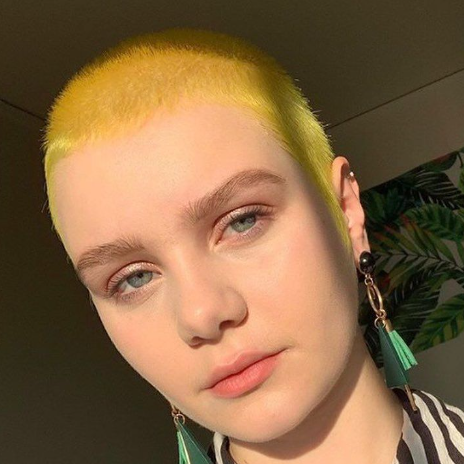}
			&\includegraphics[width=1.9cm]{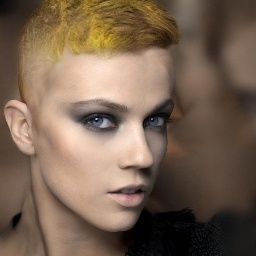}
			&\includegraphics[width=1.9cm]{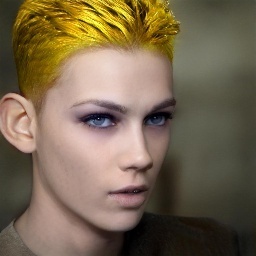}
			&\includegraphics[width=1.9cm]{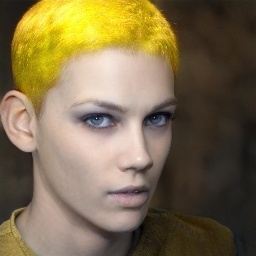}
			&\includegraphics[width=1.9cm]{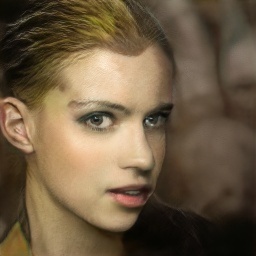}
			&\includegraphics[width=1.9cm]{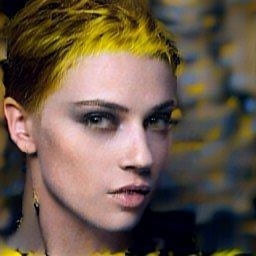}
			\\
			
			\raisebox{0.6cm}{\rotatebox[origin=c]{90}{\footnotesize{{perm gray}}}}
			&\includegraphics[width=1.9cm]{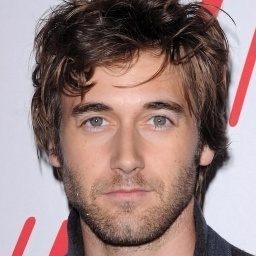}
			&\includegraphics[width=1.9cm]{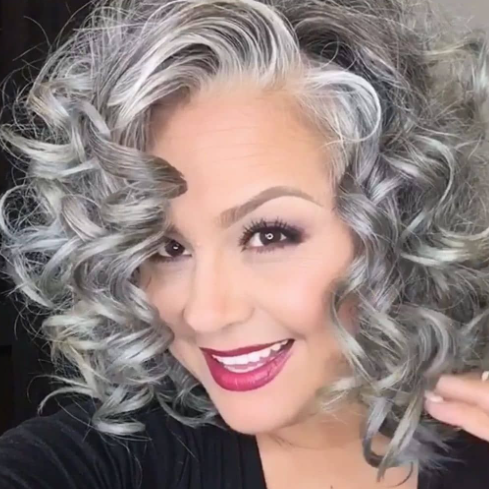}
			&\includegraphics[width=1.9cm]{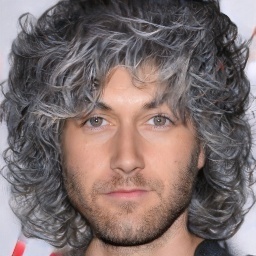}
			&\includegraphics[width=1.9cm]{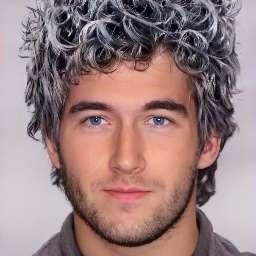}
			&\includegraphics[width=1.9cm]{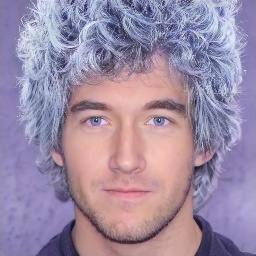}
			&\includegraphics[width=1.9cm]{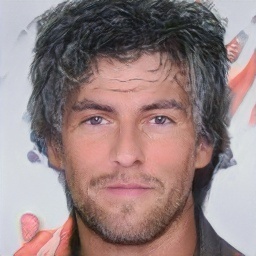}
			&\includegraphics[width=1.9cm]{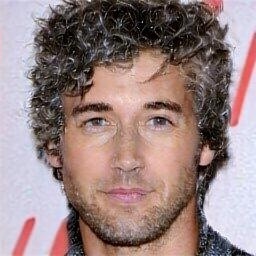}
			\\			
			
		\end{tabular}
	\end{center}
	\caption{Visual comparison with HairCLIP~\cite{wei2022hairclip},  StyleCLIP-Mapper~\cite{patashnik2021styleclip}, TediGAN~\cite{xia2021tedigan} and DiffusionCLIP~\cite{kim2022diffusionclip}. The simplified text descriptions (editing hairstyle, hair color, or both of them) are listed on the leftmost side. We additionally provide an example image for each description for better comparison. Our approach demonstrates better editing effects and irrelevant attribute preservation (e.g., identity, background, etc.).} 
	\label{fig:textcomparefig-1}
\end{figure*}

\begin{figure*}[tb]
	\begin{center}
		\setlength{\tabcolsep}{0.5pt}
		\begin{tabular}{m{0.3cm}<{\centering}m{2.15cm}<{\centering}m{2.15cm}<{\centering}m{2.15cm}<{\centering}m{2.15cm}<{\centering}m{2.15cm}<{\centering}m{2.15cm}<{\centering}m{2.15cm}<{\centering}}
			& \small{Input Image} & \small{Example} & \small{Ours} & \small{HairCLIP} & \small{StyleCLIP} & \small{TediGAN} & \small{DiffCLIP}
			\\
			
			\raisebox{0.15cm}{\rotatebox[origin=c]{90}{\footnotesize{{afro}}}}
			&\includegraphics[width=1.9cm]{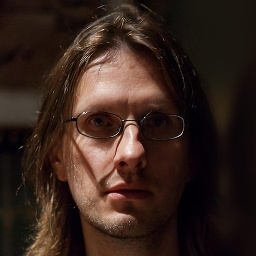}
			&\includegraphics[width=1.9cm]{supp-images/textcompare-image/afro_examplar.jpg}
			&\includegraphics[width=1.9cm]{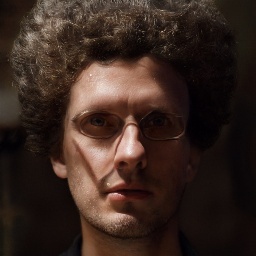}
			&\includegraphics[width=1.9cm]{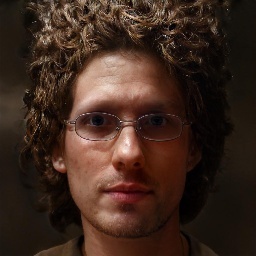}
			&\includegraphics[width=1.9cm]{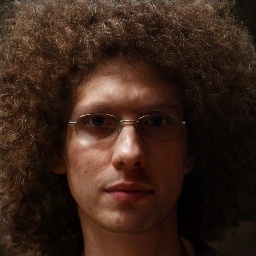}
			&\includegraphics[width=1.9cm]{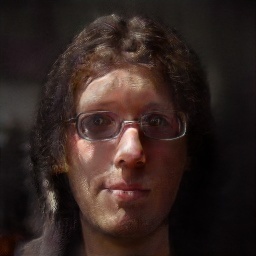}
			&\includegraphics[width=1.9cm]{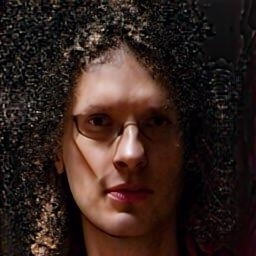}
			\\
			
			\raisebox{0.3cm}{\rotatebox[origin=c]{90}{\footnotesize{{bob cut}}}}
			&\includegraphics[width=1.9cm]{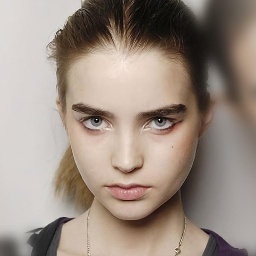}
			&\includegraphics[width=1.9cm]{supp-images/textcompare-image/bob_examplar.jpg}
			&\includegraphics[width=1.9cm]{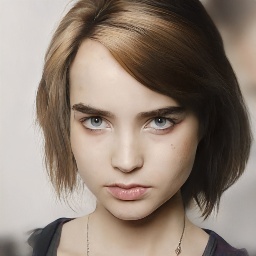}
			&\includegraphics[width=1.9cm]{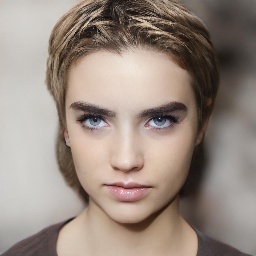}
			&\includegraphics[width=1.9cm]{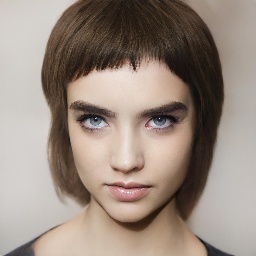}
			&\includegraphics[width=1.9cm]{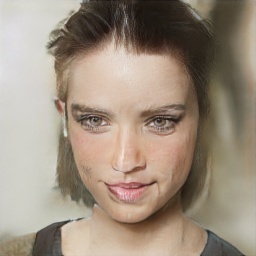}
			&\includegraphics[width=1.9cm]{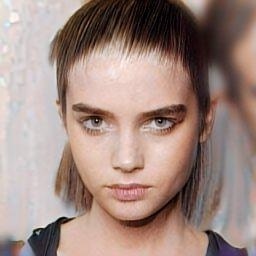}
			\\
			
			\raisebox{0.3cm}{\rotatebox[origin=c]{90}{\footnotesize{{bowl cut}}}}
			&\includegraphics[width=1.9cm]{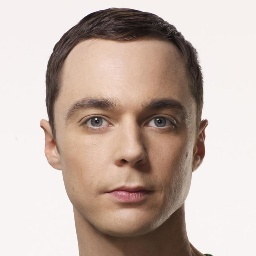}
			&\includegraphics[width=1.9cm]{supp-images/textcompare-image/bowl_examplar.jpg}
			&\includegraphics[width=1.9cm]{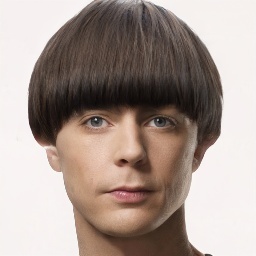}
			&\includegraphics[width=1.9cm]{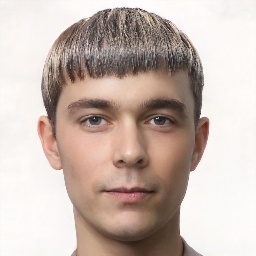}
			&\includegraphics[width=1.9cm]{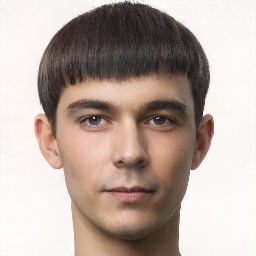}
			&\includegraphics[width=1.9cm]{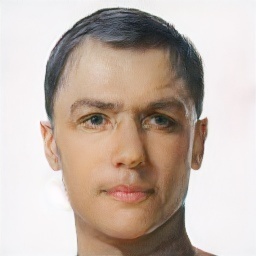}
			&\includegraphics[width=1.9cm]{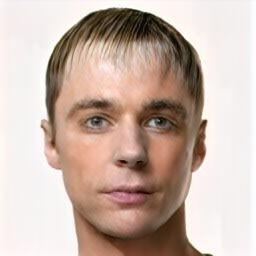}
			\\
			
			\raisebox{0.35cm}{\rotatebox[origin=c]{90}{\footnotesize{{mohawk}}}}
			&\includegraphics[width=1.9cm]{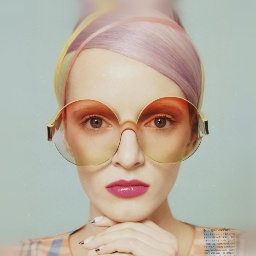}
			&\includegraphics[width=1.9cm]{supp-images/textcompare-image/mohawk_examplar.jpg}
			&\includegraphics[width=1.9cm]{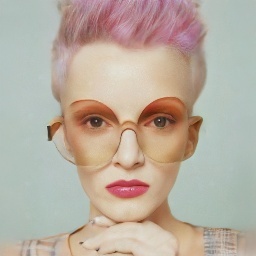}
			&\includegraphics[width=1.9cm]{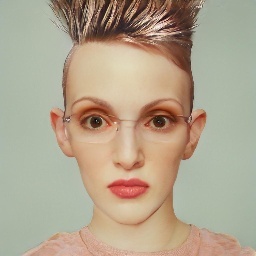}
			&\includegraphics[width=1.9cm]{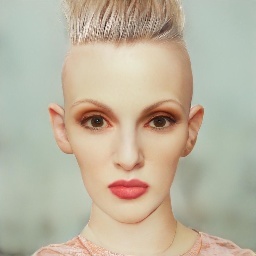}
			&\includegraphics[width=1.9cm]{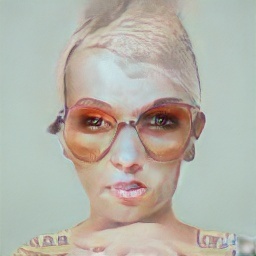}
			&\includegraphics[width=1.9cm]{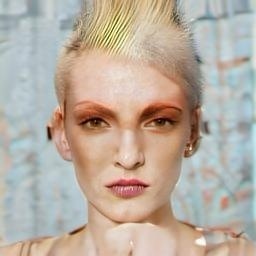}
			\\
			
			\raisebox{0.3cm}{\rotatebox[origin=c]{90}{\footnotesize{{purple}}}}
			&\includegraphics[width=1.9cm]{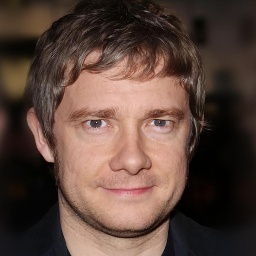}
			&\includegraphics[width=1.9cm]{supp-images/textcompare-image/purple_hair_examplar.jpg}
			&\includegraphics[width=1.9cm]{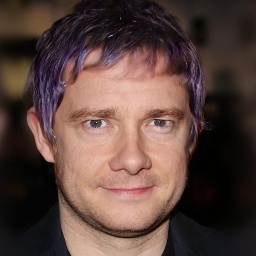}
			&\includegraphics[width=1.9cm]{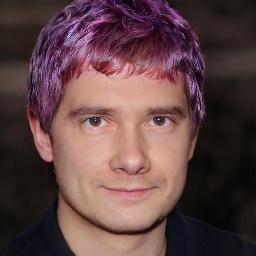}
			&\includegraphics[width=1.9cm]{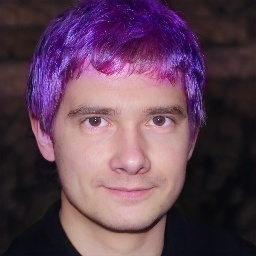}
			&\includegraphics[width=1.9cm]{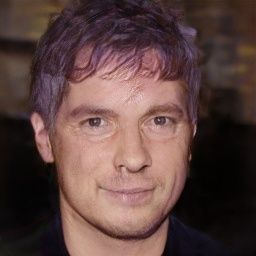}
			&\includegraphics[width=1.9cm]{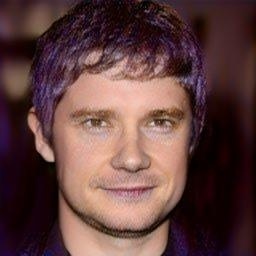}
			\\

			\raisebox{0.3cm}{\rotatebox[origin=c]{90}{\footnotesize{{green}}}}
			&\includegraphics[width=1.9cm]{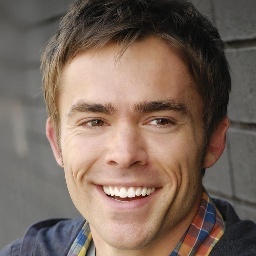}
			&\includegraphics[width=1.9cm]{supp-images/textcompare-image/green_hair_examplar.jpg}
			&\includegraphics[width=1.9cm]{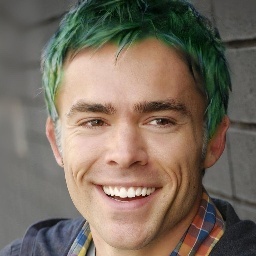}
			&\includegraphics[width=1.9cm]{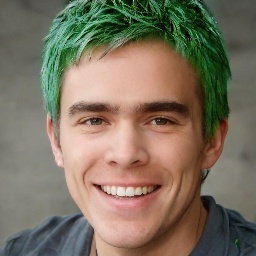}
			&\includegraphics[width=1.9cm]{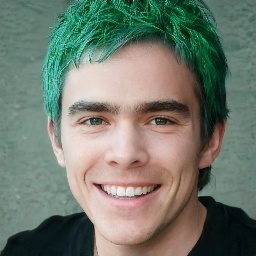}
			&\includegraphics[width=1.9cm]{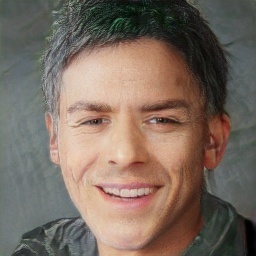}
			&\includegraphics[width=1.9cm]{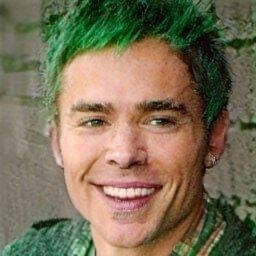}	
			\\
			
			\raisebox{0.2cm}{\rotatebox[origin=c]{90}{\footnotesize{{blond}}}}
			&\includegraphics[width=1.9cm]{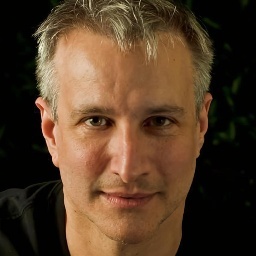}
			&\includegraphics[width=1.9cm]{supp-images/textcompare-image/blond_hair_examplar.jpg}
			&\includegraphics[width=1.9cm]{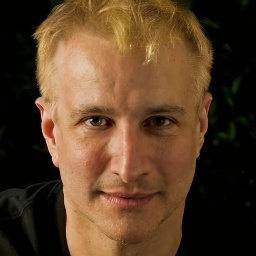}
			&\includegraphics[width=1.9cm]{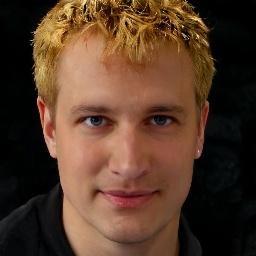}
			&\includegraphics[width=1.9cm]{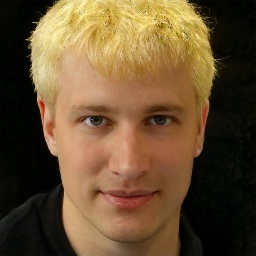}
			&\includegraphics[width=1.9cm]{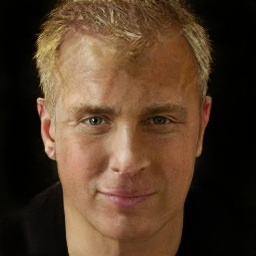}	
			&\includegraphics[width=1.9cm]{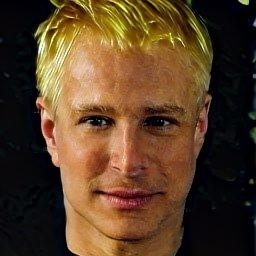}	
			\\
			
			\raisebox{0.55cm}{\rotatebox[origin=c]{90}{\footnotesize{{braid brown}}}}
			&\includegraphics[width=1.9cm]{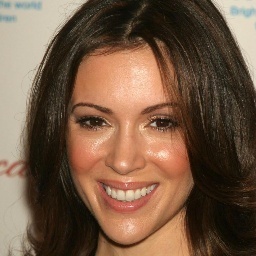}
			&\includegraphics[width=1.9cm]{supp-images/textcompare-image/braid_brown_examplar.jpg}
			&\includegraphics[width=1.9cm]{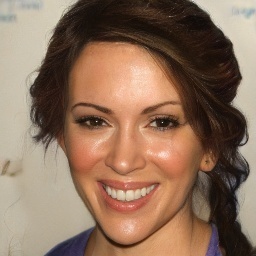}
			&\includegraphics[width=1.9cm]{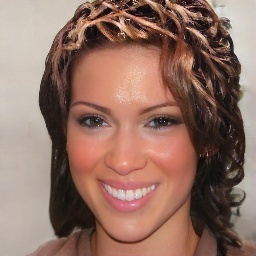}
			&\includegraphics[width=1.9cm]{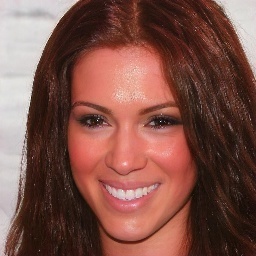}
			&\includegraphics[width=1.9cm]{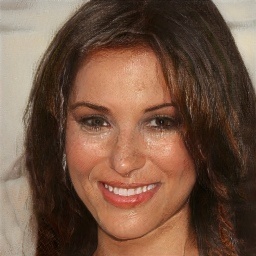}
			&\includegraphics[width=1.9cm]{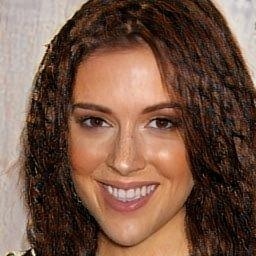}
			\\
			
			\raisebox{0.63cm}{\rotatebox[origin=c]{90}{\footnotesize{{crew yellow}}}}
			&\includegraphics[width=1.9cm]{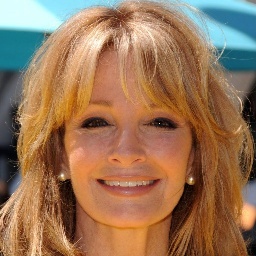}
			&\includegraphics[width=1.9cm]{supp-images/textcompare-image/crew_cut_yellow_examplar.jpg}
			&\includegraphics[width=1.9cm]{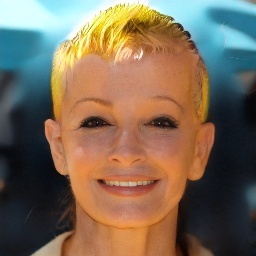}
			&\includegraphics[width=1.9cm]{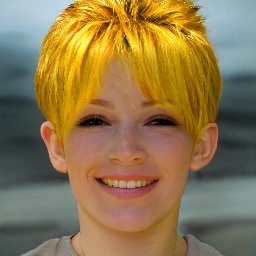}
			&\includegraphics[width=1.9cm]{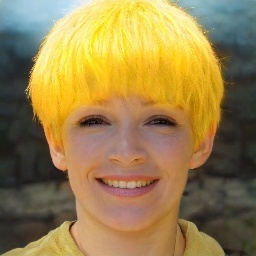}
			&\includegraphics[width=1.9cm]{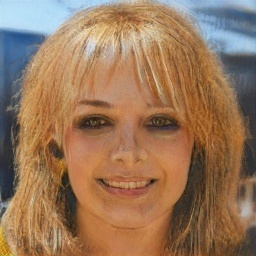}
			&\includegraphics[width=1.9cm]{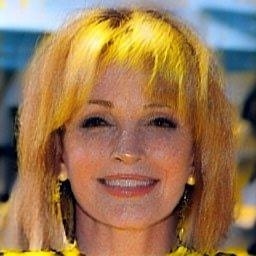}
			\\
			
			\raisebox{0.6cm}{\rotatebox[origin=c]{90}{\footnotesize{{perm gray}}}}
			&\includegraphics[width=1.9cm]{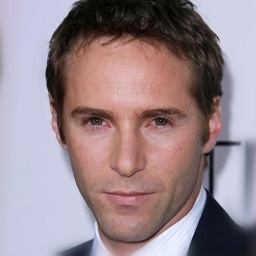}
			&\includegraphics[width=1.9cm]{supp-images/textcompare-image/perm_gray_examplar.jpg}
			&\includegraphics[width=1.9cm]{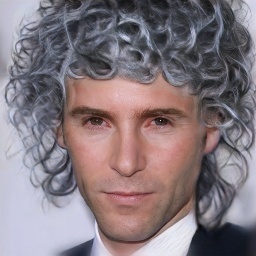}
			&\includegraphics[width=1.9cm]{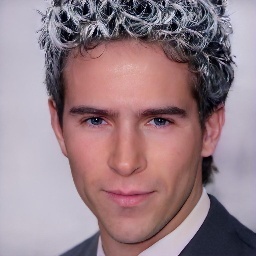}
			&\includegraphics[width=1.9cm]{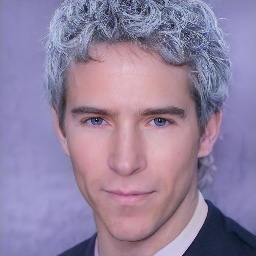}
			&\includegraphics[width=1.9cm]{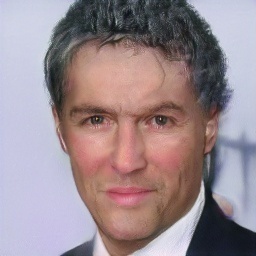}
			&\includegraphics[width=1.9cm]{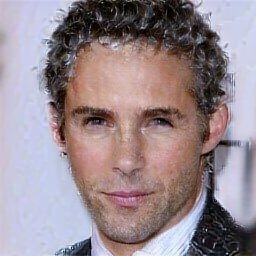}
			\\			
			
		\end{tabular}
	\end{center}
	\caption{Visual comparison with HairCLIP~\cite{wei2022hairclip},  StyleCLIP-Mapper~\cite{patashnik2021styleclip}, TediGAN~\cite{xia2021tedigan} and DiffusionCLIP~\cite{kim2022diffusionclip}. The simplified text descriptions (editing hairstyle, hair color, or both of them) are listed on the leftmost side. We additionally provide an example image for each description for better comparison. Our approach demonstrates better editing effects and irrelevant attribute preservation (e.g., identity, background, etc.).} 
	\label{fig:textcomparefig-2}
\end{figure*}

\begin{figure*}[t]
	\centering
	\includegraphics[width=\textwidth]{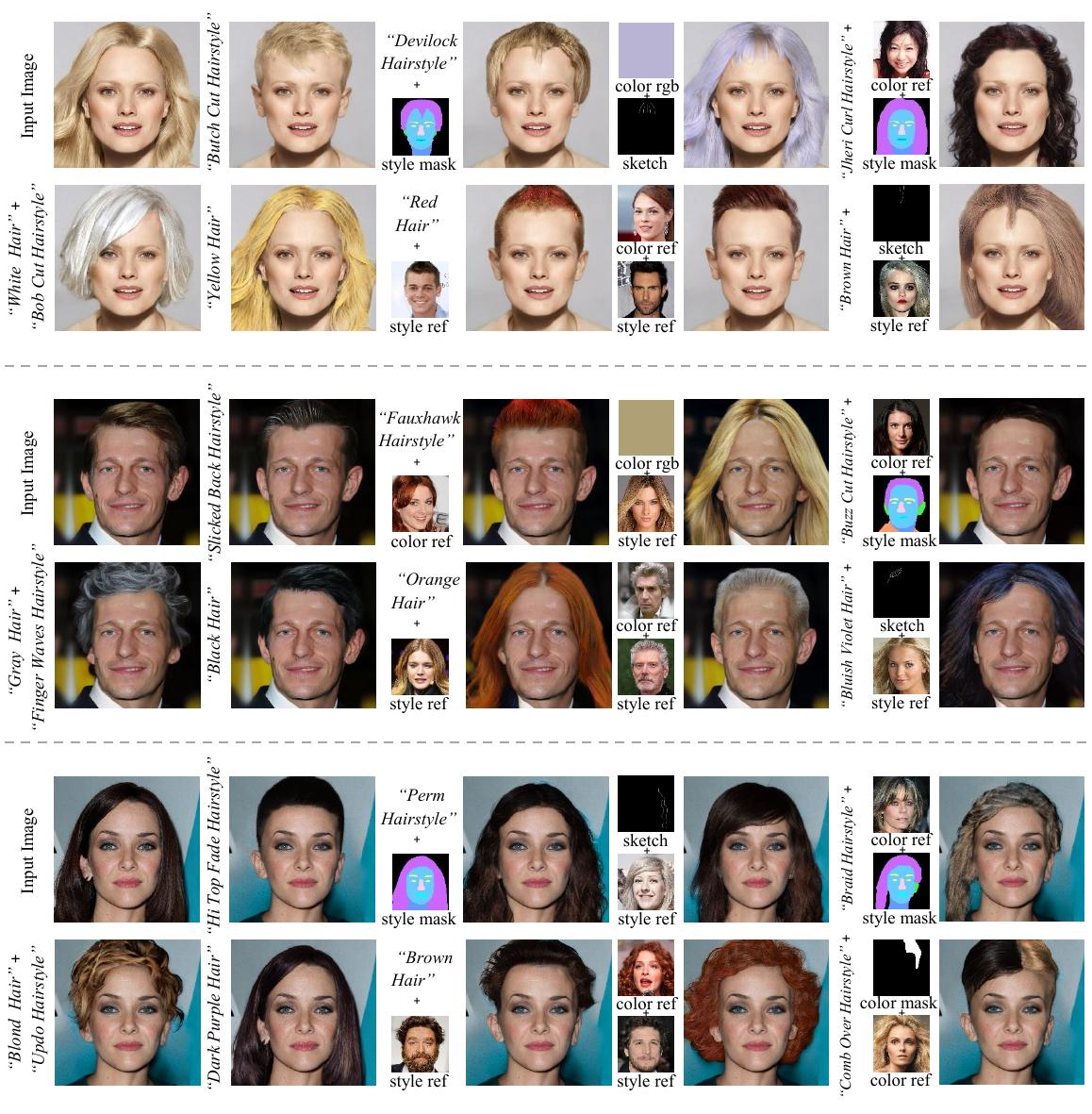} 
	\caption{HairCLIPv2 supports hairstyle and color editing individually or jointly with unprecedented user interaction mode support, including text, mask, sketch, reference image, etc.} 
	\label{fig:supp_teaser}
\end{figure*}

In Figures \ref{fig:textcomparefig-1}, \ref{fig:textcomparefig-2}, \ref{fig:transfercomparefig}, \ref{fig:sketchcomparefig}, \ref{fig:crossmodalcomparefig}, and \ref{fig:supp_teaser} we give more visual comparison results with other methods and our results for the comprehensive cross-modal conditional inputs.

\begin{figure*}[tb]
	\begin{center}
		\setlength{\tabcolsep}{0.5pt}
		\begin{tabular}{m{1.85cm}<{\centering}m{1.85cm}<{\centering}m{1.85cm}<{\centering}|m{1.85cm}<{\centering}m{1.85cm}<{\centering}m{1.85cm}<{\centering}m{1.85cm}<{\centering}m{1.85cm}<{\centering}m{1.85cm}<{\centering}}
			\scriptsize{Input Image} & \scriptsize{Hairstyle Ref} & \scriptsize{Color Ref} & \scriptsize{Ours} & \scriptsize{HairCLIP~\cite{wei2022hairclip}} & \scriptsize{LOHO~\cite{saha2021loho}} & \scriptsize{Barbershop~\cite{zhu2021barbershop}} & \scriptsize{SYH~\cite{kim2022style}} & \scriptsize{MichiGAN~\cite{tan2020michigan}}
			\\
			
			\includegraphics[width=1.8cm]{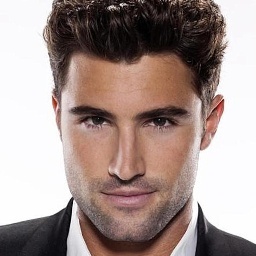}
			&\includegraphics[width=1.8cm]{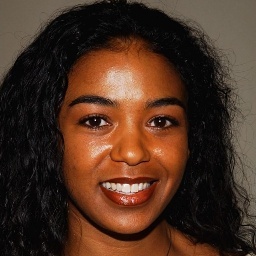}
			&\includegraphics[width=1.8cm]{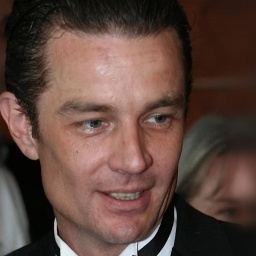}
			&\includegraphics[width=1.8cm]{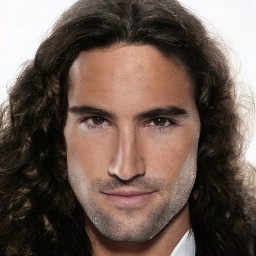}
			&\includegraphics[width=1.8cm]{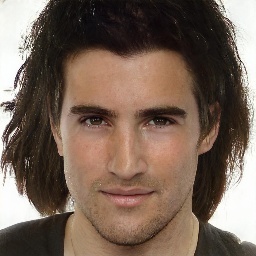}
			&\includegraphics[width=1.8cm]{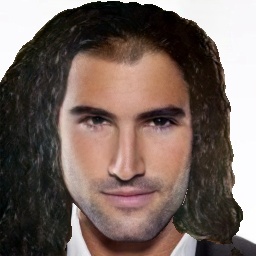}
			&\includegraphics[width=1.8cm]{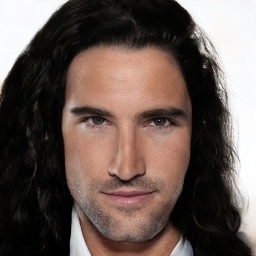}
			&\includegraphics[width=1.8cm]{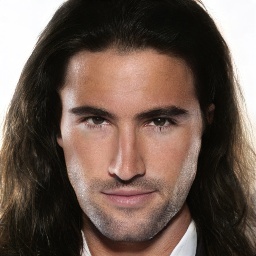}
			&\includegraphics[width=1.8cm]{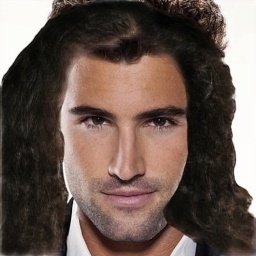}			
			\\
			
			\includegraphics[width=1.8cm]{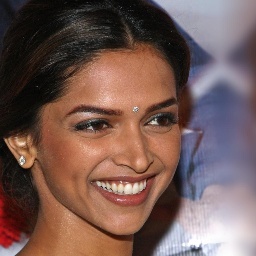}
			&\includegraphics[width=1.8cm]{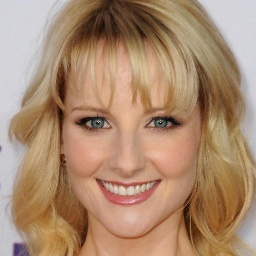}
			&\includegraphics[width=1.8cm]{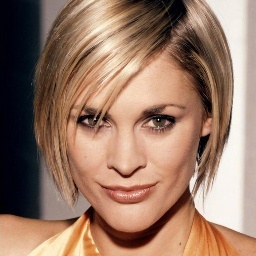}
			&\includegraphics[width=1.8cm]{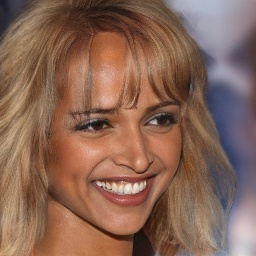}
			&\includegraphics[width=1.8cm]{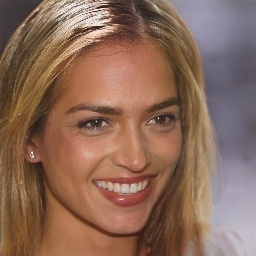}
			&\includegraphics[width=1.8cm]{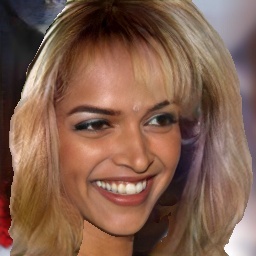}
			&\includegraphics[width=1.8cm]{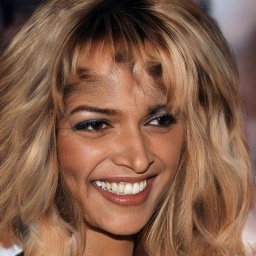}
			&\includegraphics[width=1.8cm]{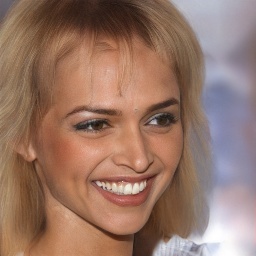}
			&\includegraphics[width=1.8cm]{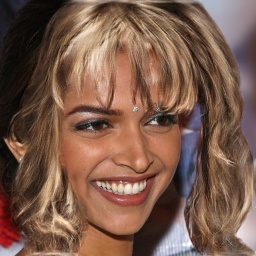}			
			\\
			
			\includegraphics[width=1.8cm]{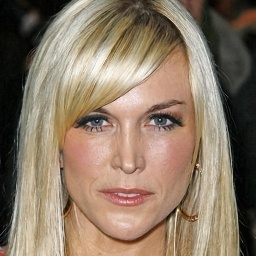}
			&\includegraphics[width=1.8cm]{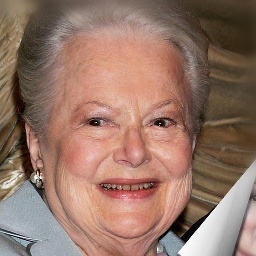}
			&\includegraphics[width=1.8cm]{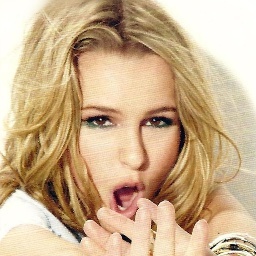}
			&\includegraphics[width=1.8cm]{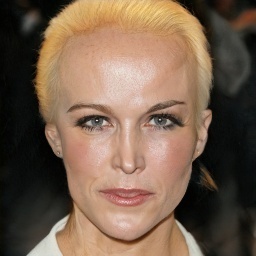}
			&\includegraphics[width=1.8cm]{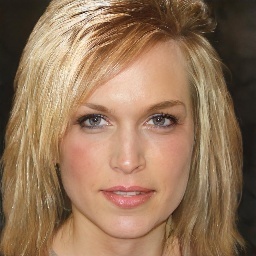}
			&\includegraphics[width=1.8cm]{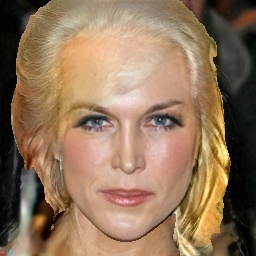}
			&\includegraphics[width=1.8cm]{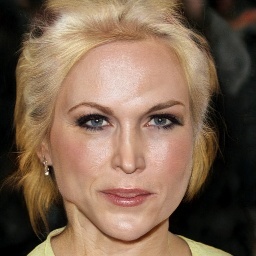}
			&\includegraphics[width=1.8cm]{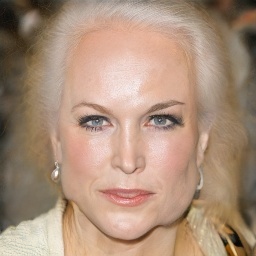}
			&\includegraphics[width=1.8cm]{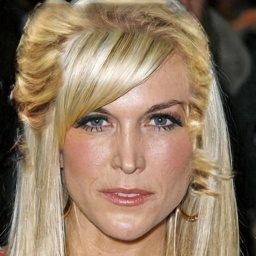}			
			\\
			
			\includegraphics[width=1.8cm]{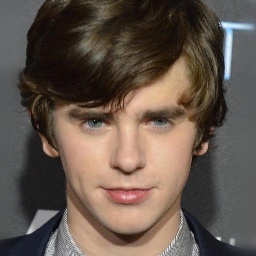}
			&\includegraphics[width=1.8cm]{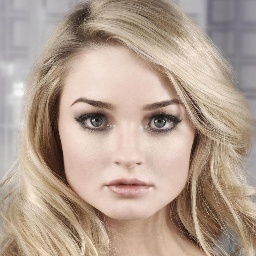}
			&\includegraphics[width=1.8cm]{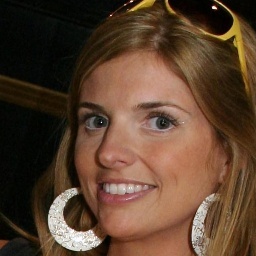}
			&\includegraphics[width=1.8cm]{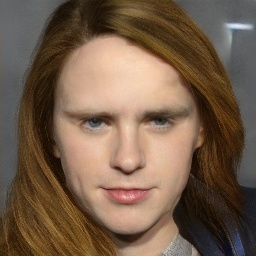}
			&\includegraphics[width=1.8cm]{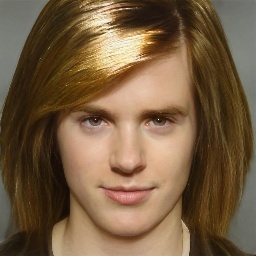}
			&\includegraphics[width=1.8cm]{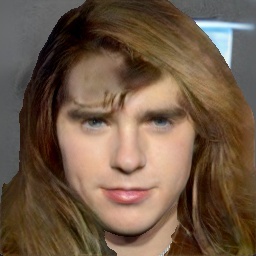}
			&\includegraphics[width=1.8cm]{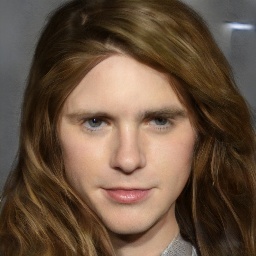}
			&\includegraphics[width=1.8cm]{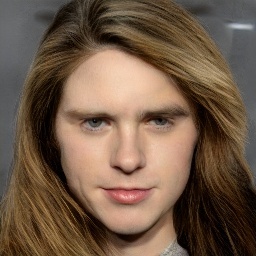}
			&\includegraphics[width=1.8cm]{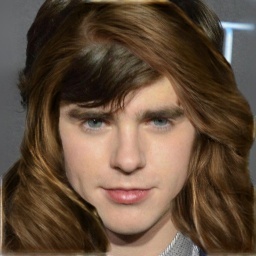}			
			\\
			
			\includegraphics[width=1.8cm]{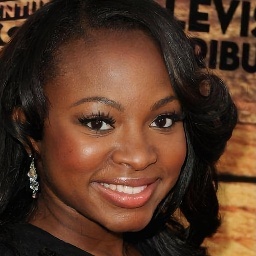}
			&\includegraphics[width=1.8cm]{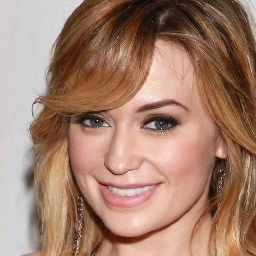}
			&\includegraphics[width=1.8cm]{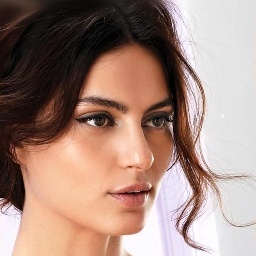}
			&\includegraphics[width=1.8cm]{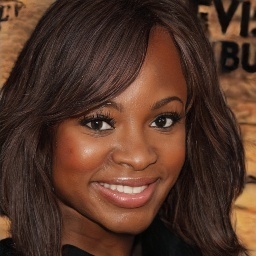}
			&\includegraphics[width=1.8cm]{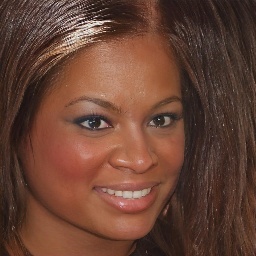}
			&\includegraphics[width=1.8cm]{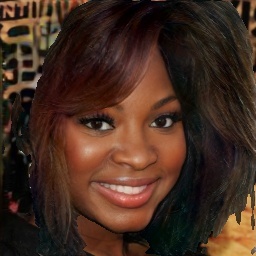}
			&\includegraphics[width=1.8cm]{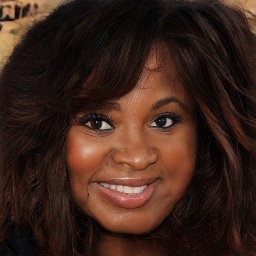}
			&\includegraphics[width=1.8cm]{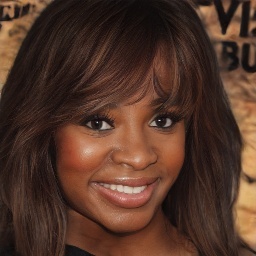}
			&\includegraphics[width=1.8cm]{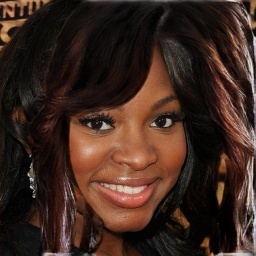}			
			\\

			\includegraphics[width=1.8cm]{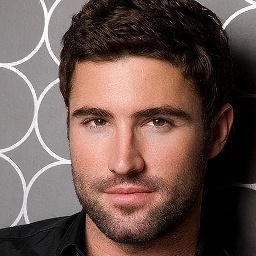}
			&\includegraphics[width=1.8cm]{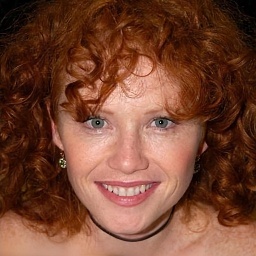}
			&\includegraphics[width=1.8cm]{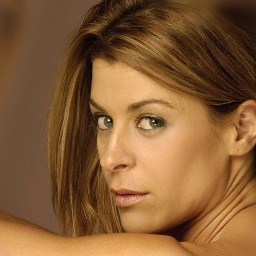}
			&\includegraphics[width=1.8cm]{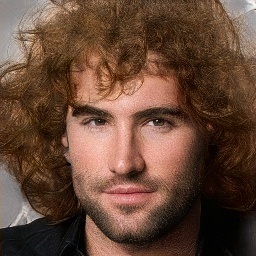}
			&\includegraphics[width=1.8cm]{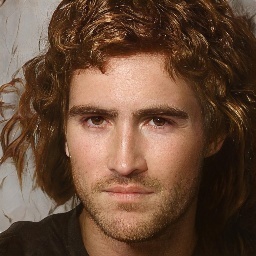}
			&\includegraphics[width=1.8cm]{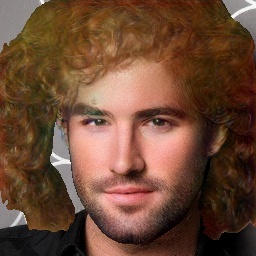}
			&\includegraphics[width=1.8cm]{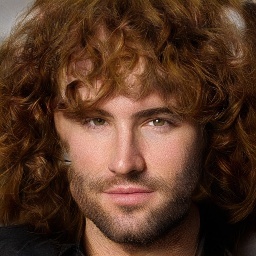}
			&\includegraphics[width=1.8cm]{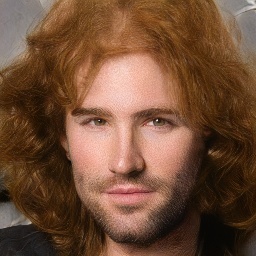}
			&\includegraphics[width=1.8cm]{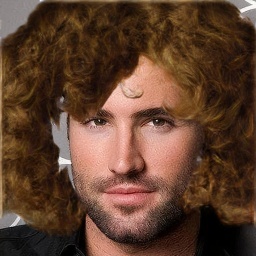}			
			\\
			
			\includegraphics[width=1.8cm]{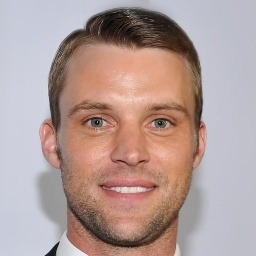}
			&\includegraphics[width=1.8cm]{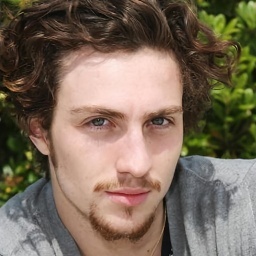}
			&\includegraphics[width=1.8cm]{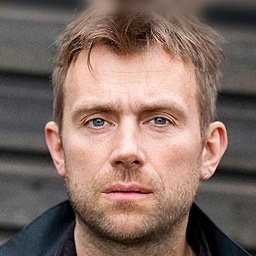}
			&\includegraphics[width=1.8cm]{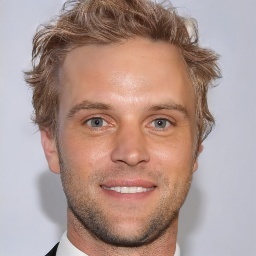}
			&\includegraphics[width=1.8cm]{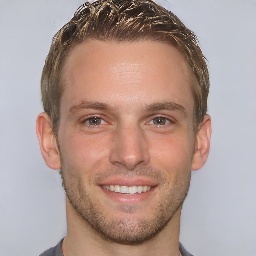}
			&\includegraphics[width=1.8cm]{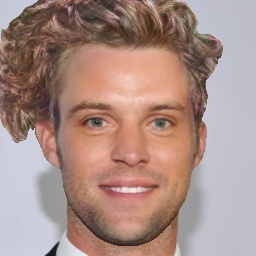}
			&\includegraphics[width=1.8cm]{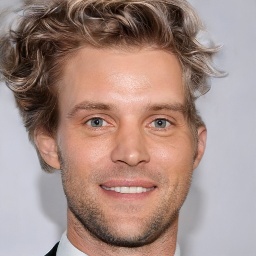}
			&\includegraphics[width=1.8cm]{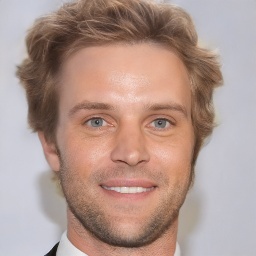}
			&\includegraphics[width=1.8cm]{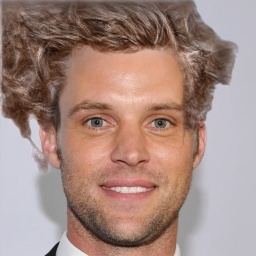}			
			\\
			
			\includegraphics[width=1.8cm]{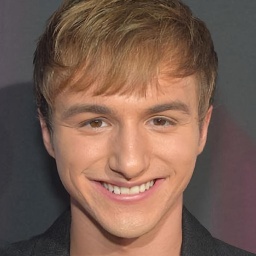}
			&\includegraphics[width=1.8cm]{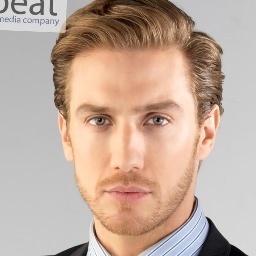}
			&\includegraphics[width=1.8cm]{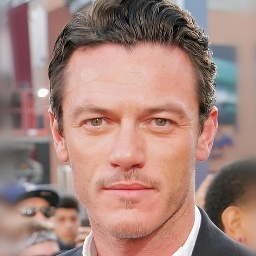}
			&\includegraphics[width=1.8cm]{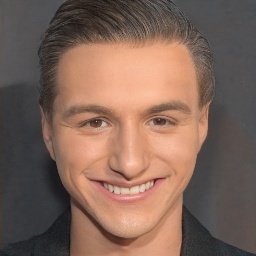}
			&\includegraphics[width=1.8cm]{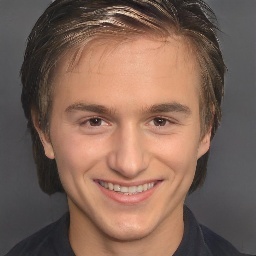}
			&\includegraphics[width=1.8cm]{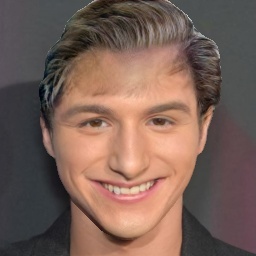}
			&\includegraphics[width=1.8cm]{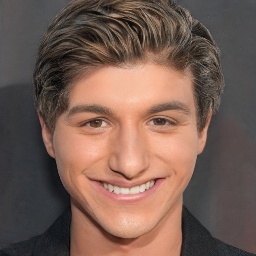}
			&\includegraphics[width=1.8cm]{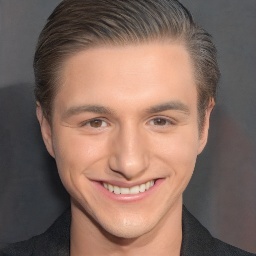}
			&\includegraphics[width=1.8cm]{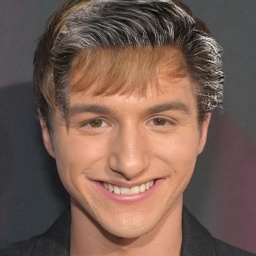}			
			\\
			
			\includegraphics[width=1.8cm]{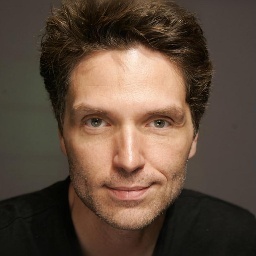}
			&\includegraphics[width=1.8cm]{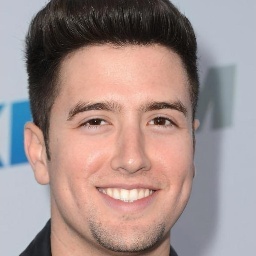}
			&\includegraphics[width=1.8cm]{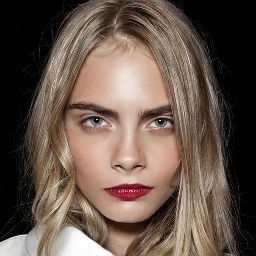}
			&\includegraphics[width=1.8cm]{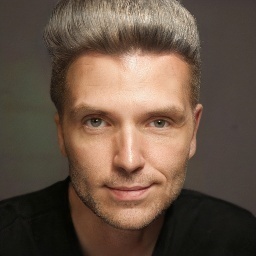}
			&\includegraphics[width=1.8cm]{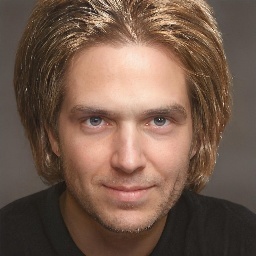}
			&\includegraphics[width=1.8cm]{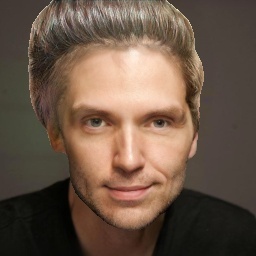}
			&\includegraphics[width=1.8cm]{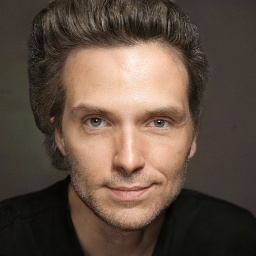}
			&\includegraphics[width=1.8cm]{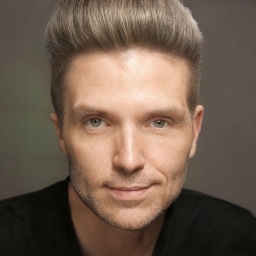}
			&\includegraphics[width=1.8cm]{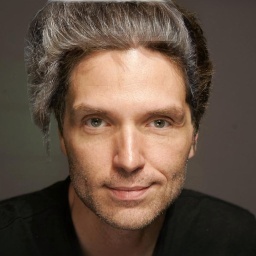}			
			\\			
			
		\end{tabular}
	\end{center}
	\caption{Visual comparison with HairCLIP~\cite{wei2022hairclip},  LOHO~\cite{saha2021loho}, Barbershop~\cite{zhu2021barbershop}, SYH~\cite{kim2022style} and MichiGAN~\cite{tan2020michigan} on hair transfer. Only our method and SYH can accomplish unaligned hair transfer while keeping irrelevant attributes unmodified.} 
	\label{fig:transfercomparefig}
	\vspace{-0.5em}
\end{figure*}

\begin{figure*}[htp]
	\begin{center}
		\setlength{\tabcolsep}{2pt}
		\begin{tabular}{m{2.8cm}<{\centering}m{2.8cm}<{\centering}m{2.8cm}<{\centering}m{2.8cm}<{\centering}m{2.8cm}<{\centering}}
			Input Image & Input Sketch & Ours & MichiGAN & SketchSalon
			\\
			
			\includegraphics[width=2.7cm]{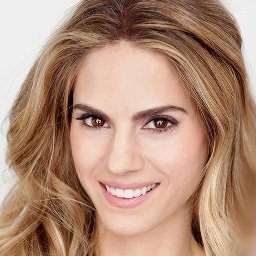}
			&\includegraphics[width=2.7cm]{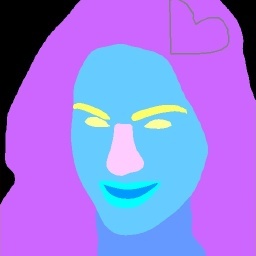}
			&\includegraphics[width=2.7cm]{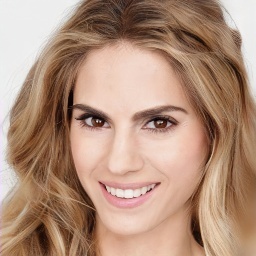}
			&\includegraphics[width=2.7cm]{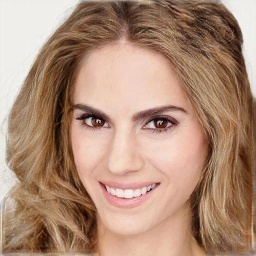}
			&\includegraphics[width=2.7cm]{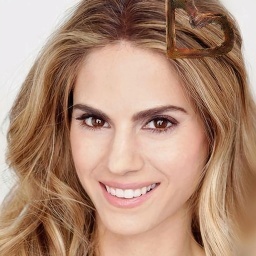}
			\\
			
			\includegraphics[width=2.7cm]{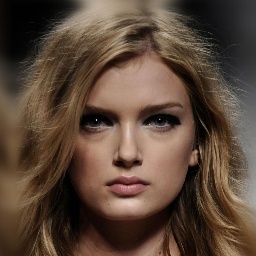}
			&\includegraphics[width=2.7cm]{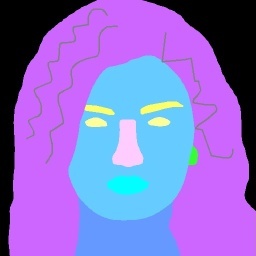}
			&\includegraphics[width=2.7cm]{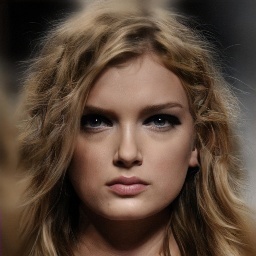}
			&\includegraphics[width=2.7cm]{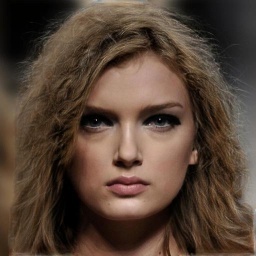}
			&\includegraphics[width=2.7cm]{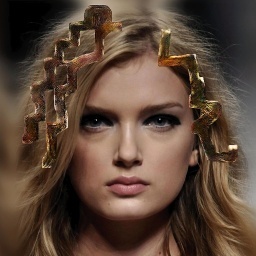}
			\\
			
			\includegraphics[width=2.7cm]{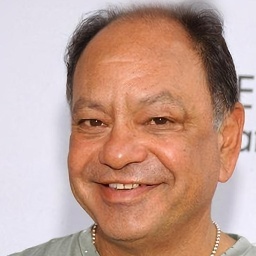}
			&\includegraphics[width=2.7cm]{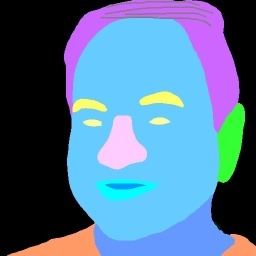}
			&\includegraphics[width=2.7cm]{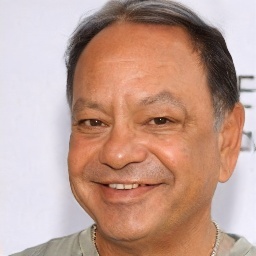}
			&\includegraphics[width=2.7cm]{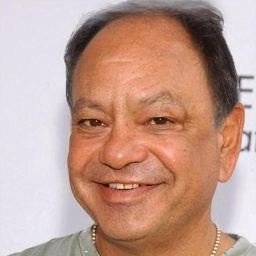}
			&\includegraphics[width=2.7cm]{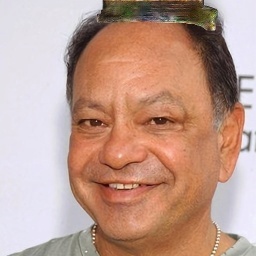}
			\\
			
			
			\includegraphics[width=2.7cm]{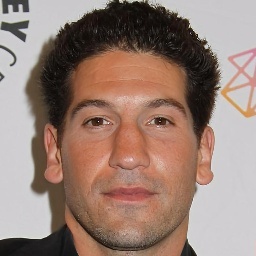}
			&\includegraphics[width=2.7cm]{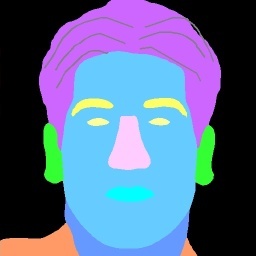}
			&\includegraphics[width=2.7cm]{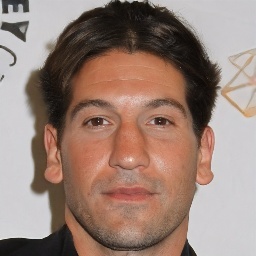}
			&\includegraphics[width=2.7cm]{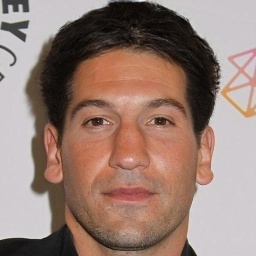}
			&\includegraphics[width=2.7cm]{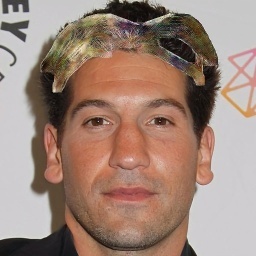}
			\\
			
			\includegraphics[width=2.7cm]{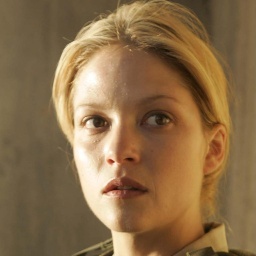}
			&\includegraphics[width=2.7cm]{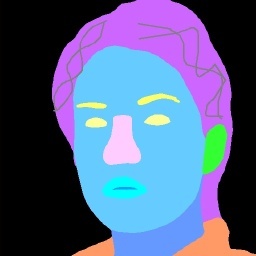}
			&\includegraphics[width=2.7cm]{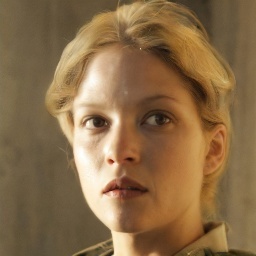}
			&\includegraphics[width=2.7cm]{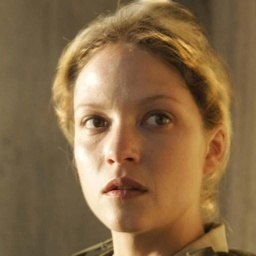}
			&\includegraphics[width=2.7cm]{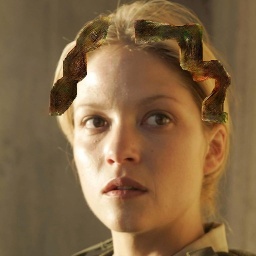}
			\\
			
			\includegraphics[width=2.7cm]{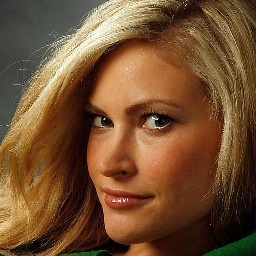}
			&\includegraphics[width=2.7cm]{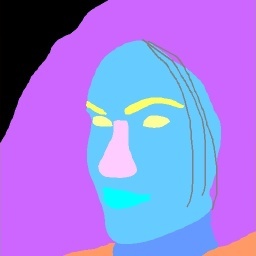}
			&\includegraphics[width=2.7cm]{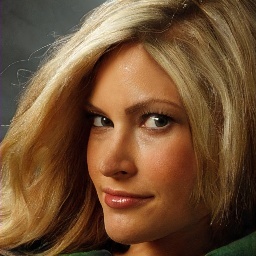}
			&\includegraphics[width=2.7cm]{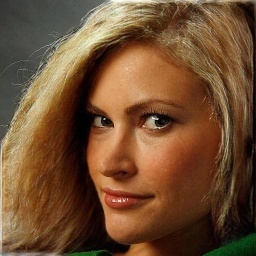}
			&\includegraphics[width=2.7cm]{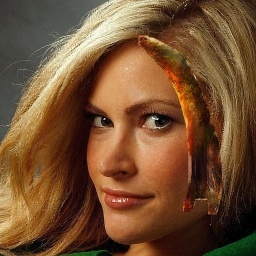}
			\\
			
		\end{tabular}
	\end{center}
	\caption{Qualitative comparison with MichiGAN~\cite{tan2020michigan} and SketchSalon~\cite{xiao2021sketchhairsalon} on sketch-based local hair editing. We provide sketches drawn in the facial parsing map for better visualization.} 
	\label{fig:sketchcomparefig}
\end{figure*}

\begin{figure*}[t]
	\centering
	\includegraphics[width=\textwidth]{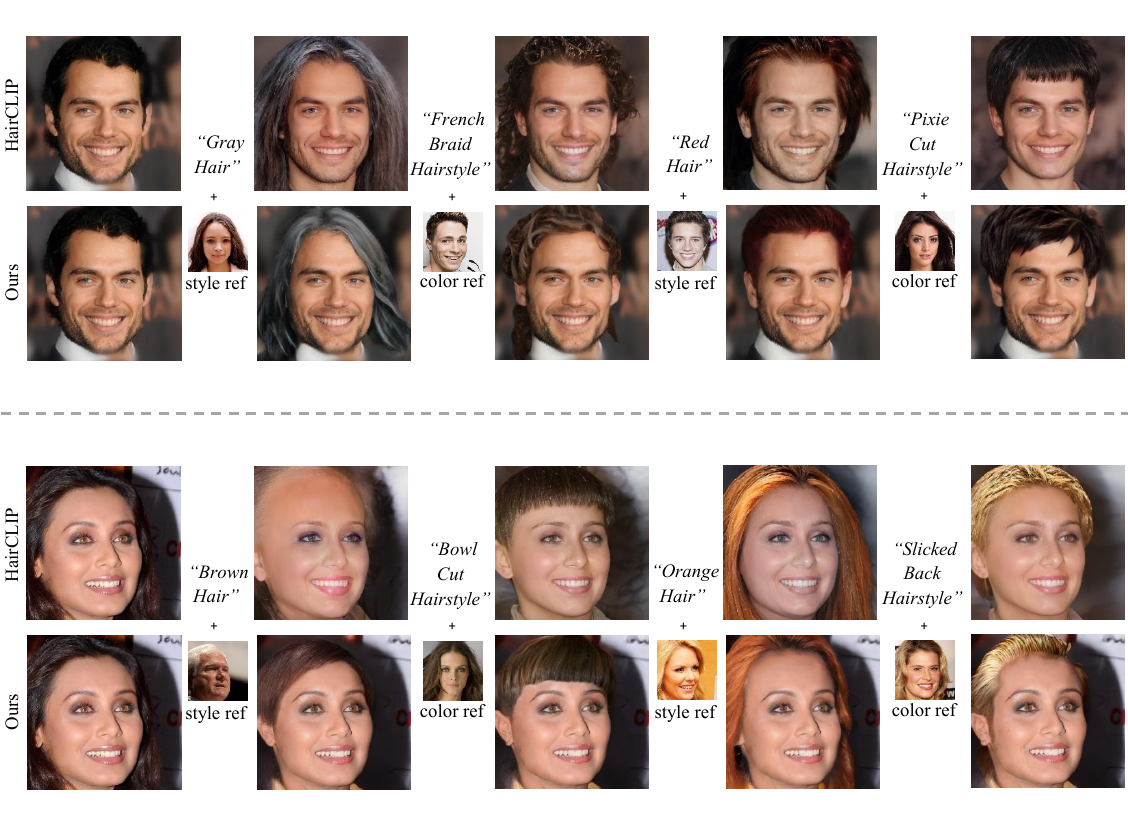} 
	\caption{Qualitative comparison with HairCLIP on cross-modal conditional input setting. Our approach exhibits better editing effects and excellent preservation of irrelevant attributes. The first column are the input supp-images.} 
	\label{fig:crossmodalcomparefig}
\end{figure*}

\end{document}